\documentclass{article}
\PassOptionsToPackage{table}{xcolor}
\usepackage{iclr2027_conference,times}
\usepackage[hyphens]{url}
\usepackage{graphicx}
\usepackage{xcolor}
\usepackage{amsmath,amssymb}
\usepackage{booktabs,array,makecell}
\usepackage{algorithm,algorithmic}
\usepackage{placeins}
\usepackage{wrapfig}
\usepackage{float}
\usepackage{needspace}
\usepackage{hyperref}
\hypersetup{hidelinks}
\iclrfinalcopy
\title{Is In-Domain Training Enough for Fine-Grained Industrial Anomaly Understanding?}
\author{
\hspace*{-0.35em}\parbox{\textwidth}{\raggedright
\textbf{Xingwu Zhang$^{1}$}\textsuperscript{*}\hspace{0.15em},\hspace{0.35em}
\textbf{Duanyang Du$^{2}$}\textsuperscript{*}\hspace{0.15em},\hspace{0.35em}
\textbf{Huiling Zhu$^{3}$}\textsuperscript{*}\hspace{0.15em},\hspace{0.35em}
\textbf{Jiayue Dai$^{2}$},\hspace{0.35em}
\textbf{Yixiao Liu$^{1}$},\hspace{0.35em}
\\[1mm]\textbf{Guozhi Liu$^{4}$},\hspace{0.35em}
\textbf{Zhihan Zhang$^{3}$},\hspace{0.35em}
\textbf{Zijun Long$^{1}$}\textsuperscript{\ensuremath{\dagger}}\\[2mm]
\normalfont
$^{1}$Hunan University\hspace{1em}
$^{2}$University of Aberdeen\hspace{1em}
$^{3}$South China Normal University\\[1mm]
$^{4}$South China University of Technology
}
}
\newcommand{\oursDK}{85.2}
\newcommand{\bestSelectedMLLM}{74.8}

\newcommand{\method}{SiGMA}

\begin{document}
\maketitle
\begingroup
\renewcommand{\thefootnote}{\fnsymbol{footnote}}
\footnotetext[1]{These authors contributed equally to this work.}
\footnotetext[2]{Corresponding author.}
\endgroup
\fancyhf{}
\cfoot{\thepage}
\renewcommand{\headrulewidth}{0pt}
\begin{abstract}

A single multimodal large language model (MLLM) struggles to excel simultaneously at detection, localization, description, and reasoning in multimodal industrial anomaly understanding (MM-IAU). We show that in-domain training does not close this gap. On MMAD, a widely adopted MM-IAU benchmark, trained specialists reach at most 75.5\% accuracy in defect localization, against 92.3\% for human experts, and even detect anomalies less accurately than their untrained base model.
Meanwhile, different MLLMs offer complementary strengths but share this weakness in fine-grained perception, so combining them alone cannot remove it. We therefore propose SiGMA, a spatially grounded multi-agent framework that divides labor between heterogeneous MLLM agents and a dedicated visual defect expert. A multimodal searcher supplies industrial knowledge and normal references, the defect expert turns query–reference comparison into calibrated anomaly evidence, and a label-free reliability controller weighs each source by task-wise competence and query-level evidence quality. SiGMA reaches 85.2\% average accuracy on MMAD, 4.0\% above the strongest trained specialist and Gemini-2.5-Pro and within 1.5\% of human experts. Even with three agents of at most 9B parameters, it reaches 84.4\%, and new MLLMs join without retraining.


\end{abstract}

\section{Introduction}
\label{sec:intro}
Multimodal industrial anomaly understanding (MM-IAU) requires detecting whether a product is defective, localizing the defect, describing its appearance, and reasoning about its potential impact~\citep{jiang2025mmad}. Together, these tasks demand a breadth of perceptual and reasoning capabilities that a single multimodal large language model (MLLM) struggles to master~\citep{jiang2025mmad,li2025triad}. Defects such as small scratches or missing components often occupy only a small fraction of the image~\citep{bergmann2019mvtec,zou2022spot}, so detecting and localizing them requires fine-grained comparison with a normal reference, whereas describing and analyzing them relies on semantic knowledge and reasoning. The prevailing remedy is to specialize a single MLLM through in-domain supervised fine-tuning or reinforcement learning~\citep{chao2025anomalyr1,zhao2025omniad,guan2025emit,kang2026judo,wang2026maugpt}. This raises a basic question of whether in-domain training is enough for fine-grained industrial anomaly understanding.

\suppressfloats[t]
\begin{figure*}[t]
  \centering
  \includegraphics[width=\textwidth]{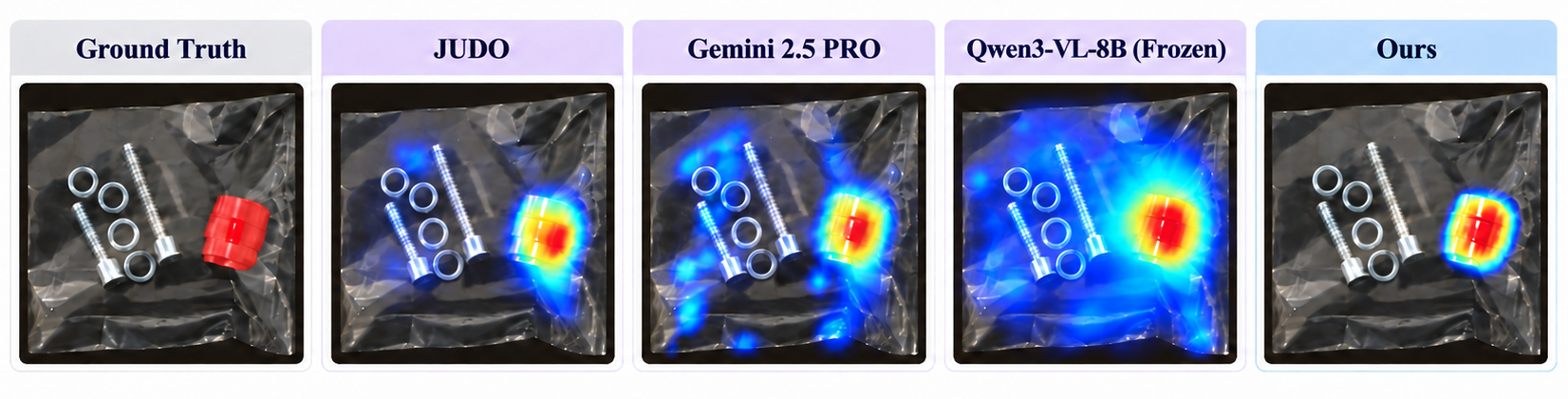}
  \vspace{-8mm}
  \caption{Fine-grained defect visualization on MMAD shows that \method{} concentrates its response on the defect while the baselines show diffuse background activations.}
  \label{fig:intro_qualitative_comparison}
  \vspace{-3mm}
\end{figure*}

Our analysis of 39,670 questions from MMAD~\citep{jiang2025mmad}, a widely adopted MM-IAU benchmark covering seven tasks, suggests that it is not (\autoref{tab:compare}). Specialization improves particular capabilities but leaves models uneven across tasks. Although OmniAD~\citep{zhao2025omniad} and JUDO~\citep{kang2026judo} are both trained from Qwen2.5-VL-7B~\citep{bai2025qwen25vl}, JUDO outperforms OmniAD by 17.4\% in defect description, whereas OmniAD leads in defect classification, localization, and anomaly detection. More importantly, training leaves a pronounced shortfall in fine-grained perception. The strongest trained method for defect localization reaches only 75.5\% accuracy, compared with 92.3\% for human experts~\citep{jiang2025mmad}, and both OmniAD and JUDO detect anomalies less accurately than their untrained base model (68.8\% and 65.0\% vs.\ 70.5\%). In-domain training thus leaves a substantial gap in the capability most central to industrial inspection.

Different MLLMs nevertheless offer complementary strengths, which multi-model methods can combine~\citep{jiang2023llmblender,wang2025mixture,ong2025routellm}. Among the ten compact MLLMs in \autoref{tab:compare}, Qwen3-VL-4B~\citep{bai2025qwen3vl} performs best on anomaly detection and defect localization, whereas GLM-4.1V-9B-Thinking~\citep{hong2025glm41v} leads defect description, defect analysis, and object classification while ranking only eighth of ten in anomaly detection. This complementarity, however, does not extend to fine-grained perception, where the weakness is shared across models. Defect localization averages only 55.94\% across the ten models, and even selecting the best model for each task leaves detection and localization at 74.1\% and 63.6\%. Nor does scale remove this weakness. Within the Qwen3-VL family, the 8B model localizes defects less accurately than the 4B model (60.8\% vs.\ 63.6\%), and even Gemini-2.5-Pro~\citep{comanici2025gemini25} reaches only 67.2\% in defect localization. Because combining predictions can only recover errors on which models disagree~\citep{krogh1994neural}, composing MLLMs alone cannot close this gap.

We therefore pursue a division of labor that brings heterogeneous MLLM agents and a dedicated visual defect expert into the same framework. The agents contribute complementary judgments across inspection tasks, while the defect expert supplies localized defect evidence through explicit comparison with normal references~\citep{zhang2026rad}, a source that does not rely on MLLM perception. 
This evidence must be used with care, since naively overlaying anomaly maps on MLLM inputs lowers anomaly-detection accuracy by up to 12.62\%, partly because detectors respond spuriously on normal images~\citep{jiang2025mmad}. The expert's evidence therefore needs calibration, and its influence should depend on the task and on the evidence quality of each query.


To this end, we introduce SiGMA, a spatially grounded multi-agent framework following a search--reason--ground--aggregate workflow. A multimodal searcher retrieves industrial knowledge and a normal reference as shared context, and heterogeneous MLLM agents answer independently. In parallel, a defect expert calibrates query--reference discrepancies against normal statistics and converts them into answers through expert reader agents. A label-free reliability controller then weighs each source by task-wise competence and query-level evidence quality. \autoref{fig:intro_qualitative_comparison} illustrates the resulting fine-grained defect localization. All model parameters remain frozen, so new MLLMs can join the team without retraining.

Our contributions are as follows. (i) We show that in-domain training is not enough for fine-grained industrial anomaly understanding, since trained MLLMs remain uneven across tasks and share a perceptual weakness, which motivates a division of labor between complementary MLLMs and a visual defect expert. (ii) We introduce SiGMA, which combines heterogeneous MLLM predictions with calibrated reference-based defect evidence through a label-free reliability controller that accounts for task-wise source competence and query-specific evidence quality. (iii) On MMAD, SiGMA reaches 85.2\% average accuracy, 4.0\% above both JUDO and Gemini-2.5-Pro, with its largest gains over trained methods in anomaly detection ($13.8\%\uparrow$) and defect localization ($4.1\%\uparrow$).


\section{Related Work}
\label{sec:related}

\paragraph{Industrial anomaly understanding.}
MLLM-based methods for MM-IAU~\citep{jiang2025mmad} mostly specialize a single model through supervised fine-tuning or reinforcement learning~\citep{chao2025anomalyr1,zhao2025omniad,guan2025emit,kang2026judo}, learned comparison with normal templates~\citep{jiang2026adcopilot}, internal expert modules~\citep{wang2026maugpt}, or trained tool use~\citep{miao2025agentiad}. Others keep one MLLM frozen and guide it with retrieved knowledge and normal images~\citep{jiang2025mmad,chen2025echo}, a design our searcher adopts, or with test-time latent optimization~\citep{chen2026reasoniad}. In both cases one model carries every capability, so the system's perception rests on that model. SiGMA instead divides these capabilities among heterogeneous frozen MLLMs and a reference-based visual defect expert.


\paragraph{Multi-MLLM composition.}
Combining models helps when their errors differ~\citep{krogh1994neural}. Existing methods rank and fuse responses~\citep{jiang2023llmblender}, route queries~\citep{ong2025routellm}, or synthesize outputs with an aggregator model~\citep{wang2025mixture}. Ranking and routing learn their selectors from labeled or preference data, which industrial settings rarely provide, while aggregator-based synthesis adds model calls and relies on a single aggregator to judge which responses to trust. More fundamentally, these methods cannot fix a weakness shared by the whole pool. Label-free reliability estimation~\citep{dawid1979observer,whitehill2009glad} removes the need for labels but typically assigns each source one competence across all items and uses no evidence beyond the votes. Our SiGMA extends this estimation with task-wise competence and admits an external visual expert, weighted by query-level evidence quality.
\vspace{-3mm}


\paragraph{Visual evidence for MLLMs.}
Other methods feed vision-expert evidence to a single MLLM, either by training it to consume anomaly maps or expert tokens~\citep{gu2024anomalygpt,li2023myriad,li2025triad,xu2025anomalyov} or without training through map overlays~\citep{jiang2025mmad} and confidence-gated attention guidance~\citep{peng2026eagle}. Trained variants tie the evidence to one model, overlays can mislead the MLLM with spurious responses on normal images, and attention guidance requires access to the MLLM's internal attention. Our SiGMA instead calibrates reference-based evidence~\citep{zhang2026rad} against normal residual statistics, lets several MLLM readers interpret it, and weighs their answers by label-free reliability and evidence quality.

\section{Method} 

 \begin{figure*}[!t]
      \centering
      \includegraphics[width=\textwidth]{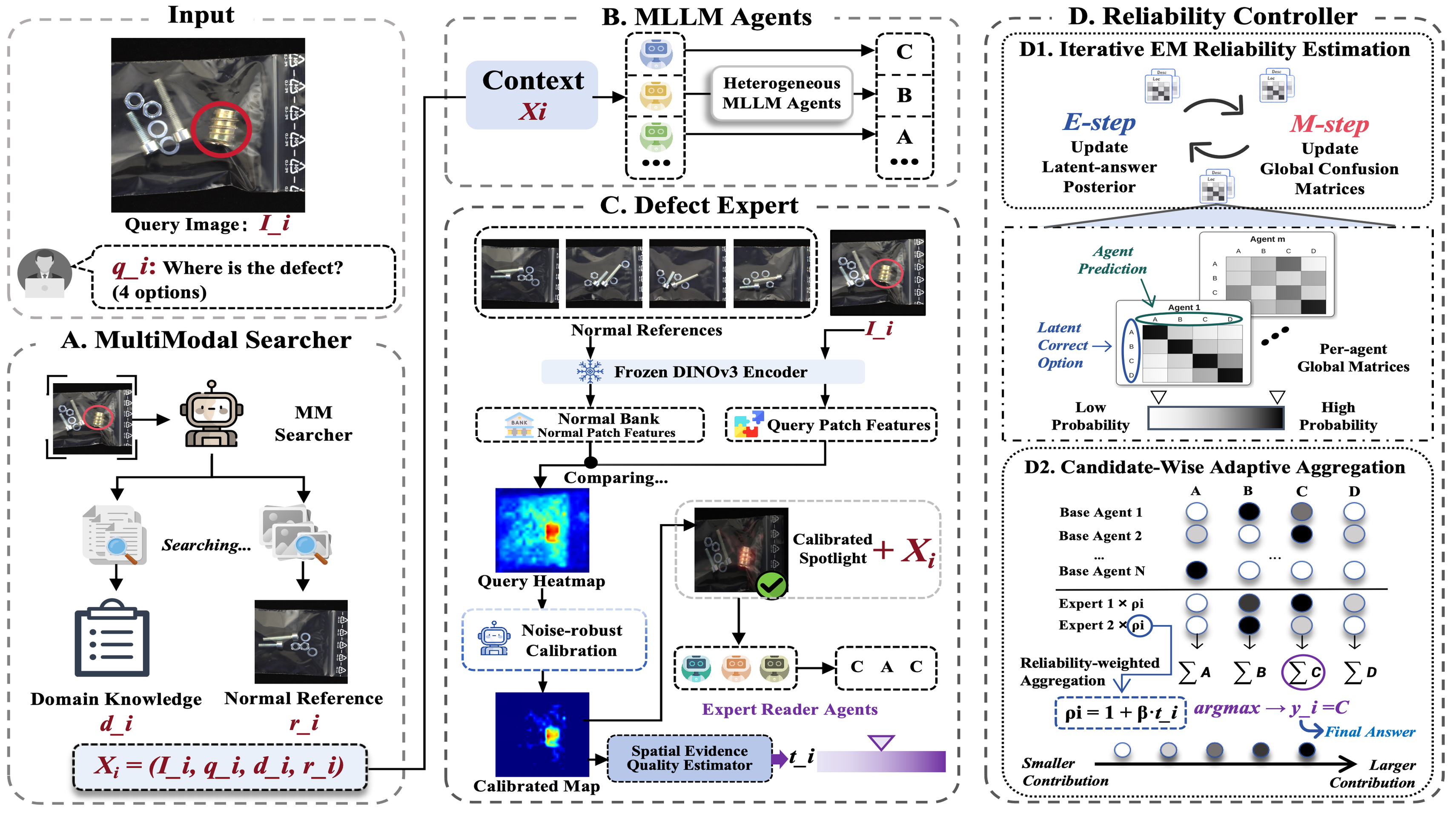}
     \vspace{-6mm}
      \caption{
      Overview of the search--reason--ground--aggregate workflow in
      \method{}.    }
      \label{fig:pipeline}
      \vspace{-3mm}
  \end{figure*}



\subsection{Problem Formulation and Motivation}
Each MM-IAU instance consists of a query image, an inspection question, and a set of candidate answers. Section~\ref{sec:intro} identified three properties of this task. Capabilities are distributed across MLLMs, fine-grained perception is a weakness they share, and the value of expert evidence varies across tasks and queries. SiGMA therefore divides labor among three roles within a search--reason--ground--aggregate workflow (\autoref{fig:pipeline}
).

\subsection{Multimodal Searcher}
Judging whether a product is defective requires knowing what it should look like, which general-purpose MLLMs cannot infer from the query image alone. Following prior work that guides MLLMs with retrieved domain knowledge and normal images~\citep{jiang2025mmad,chen2025echo}, we construct query-specific context by multimodal retrieval. For the \(i\)-th MM-IAU instance, let \(I_i\) denote the query image and \(q_i\) the complete multiple-choice prompt, including the question and candidate-answer contents. The searcher encodes \(I_i\) with a frozen CLIP model~\citep{radford2021learning}. Its text branch retrieves industrial knowledge \(d_i\), including product semantics, normal structures, and defect-related descriptions, and its visual branch selects the most similar normal reference \(r_i\) from a category-specific gallery. The knowledge \(d_i\) grounds the heterogeneous agents semantically, whereas \(r_i\) anchors normal appearance for both the agents and the defect expert. The sources of the textual knowledge and normal references are specified in Appendix~\ref{app:reference_budget}.

\subsection{Heterogeneous MLLM Agents} 
The selected off-the-shelf MLLMs exhibit distinct task-wise strengths and solve partially different subsets of MM-IAU instances (Section~\ref{sec:intro}). 
Rather than collapsing this diversity into a single specialized model, we retain them as heterogeneous frozen agents that generate complementary candidate predictions. Let \(\mathcal{A}=\{\mathrm{A},\mathrm{B},\mathrm{C},\mathrm{D}\}\) denote the shared option-label set.
For a pool of \(N\) frozen MLLM agents, agent \(m\) independently produces \(o_{im}=f_m(I_i,q_i,\mathcal{A},d_i,r_i)\), for \(m=1,\ldots,N\),
where \(f_m\) denotes the frozen inference process of agent \(m\), and \(o_{im} \in \mathcal{A}\) is its prediction for instance \(i\).


\subsection{Fine-Grained Defect Expert} 
Composing heterogeneous agents cannot recover perceptual errors that all of them share. Small scratches, missing components, and subtle structural deformations may occupy only a limited portion of the image and can be overlooked during holistic visual reasoning by every agent. We therefore introduce a fine-grained defect expert that localizes query--reference discrepancies without relying on MLLM perception, calibrates the resulting evidence to suppress unreliable responses, and passes the calibrated map to expert readers agents, that we call readers for short. The readers need not find a subtle defect themselves, since they only interpret explicitly highlighted evidence and map it to an answer option.


\paragraph{Spatially grounded anomaly localization.} To expose localized defect cues that may not be directly captured by general-purpose MLLMs, we adopt a RAD-style training-free detector~\citep{zhang2026rad}, whose reference-based patch matching naturally aligns with our goal of grounding anomaly reasoning in spatial deviations from normal appearance. Specifically, a frozen DINOv3 encoder~\citep{simeoni2025dinov3} extracts patch-level features from the query image \(I_i\) and the retrieved normal reference \(r_i\). Local discrepancies between their patch features are used to construct a dense anomaly map \(\mathbf{H}_i = \{h_i(p)\}_{p \in \Omega_i}\), where \(\Omega_i\) denotes the spatial domain of the map and \(h_i(p)\) is the anomaly response at location \(p\).  Unlike a global anomaly score, \(\mathbf{H}_i\) retains where the query deviates from normal appearance, which defect localization requires. The calibrated map defined below is rendered as an overlay on the query image for the readers.

\paragraph{Noise-robust calibration.}

Per-image normalization rescales each raw map by its own response range, so weak query--reference variations can be amplified into salient responses even on normal images. On MMAD, visualizing detector maps for MLLMs improves defect localization but lowers anomaly-detection accuracy by up to 12.62\%, partly because of such spurious responses~\citep{jiang2025mmad}. We therefore calibrate the map against normal residual statistics.

To establish an absolute anomaly scale, the multimodal searcher first retrieves a set \(\mathcal{G}_i\) of \(K\) category-matched normal references most similar to \(I_i\). We then apply the same query--reference detector to all distinct references in \(\mathcal{G}_i\) and collect their residual responses into the normal residual pool \(\mathcal{R}_i^{\mathrm{N}}\). Using this pool, let \(p_{\mathrm{low}}\) denote a high calibration quantile and let \(\kappa > 0\) denote an expansion coefficient. The lower and upper calibration bounds are defined as
\begin{equation}
\scalebox{1.00}{$
\begin{aligned}
\theta_{i,\mathrm{low}}
&=
\operatorname{Quantile}_{p_{\mathrm{low}}}
\left(
\mathcal{R}_i^{\mathrm{N}}
\right), \quad
\theta_{i,\mathrm{high}}
=
\theta_{i,\mathrm{low}}
+
\kappa
\max
\left\{
\theta_{i,\mathrm{low}}
-
\operatorname{Median}
\left(
\mathcal{R}_i^{\mathrm{N}}
\right),
0
\right\}.
\end{aligned}
$}
\end{equation}
Using these category-calibrated bounds, we normalize the anomaly response as
\begin{equation}
\widetilde{h}_i(p)
=
\operatorname{clip}
\left(
\frac{h_i(p)-\theta_{i,\mathrm{low}}}
{\max\{\theta_{i,\mathrm{high}}-\theta_{i,\mathrm{low}},\epsilon\}},
0, 1
\right),
\qquad
p \in \Omega_i,
\end{equation}
where \(\epsilon=10^{-8}\) prevents division by zero when the bounds coincide, yielding the calibrated anomaly map \(\widetilde{\mathbf{H}}_i = \left\{\widetilde{h}_i(p)\right\}_{p \in \Omega_i}\).

This anchors the anomaly scale to normal query--reference variation rather than to per-query extrema. Responses below \(\theta_{i,\mathrm{low}}\) are suppressed, whereas stronger deviations are retained with comparable intensity across queries. However, pixel-wise calibration alone cannot distinguish a spatially coherent defect from isolated high-response noise. We therefore define the high-response region
$\Gamma_i = \{p \in \Omega_i : h_i(p) > \theta_{i,\mathrm{low}}\}$
and measure its spatial coherence using the area fraction of its largest connected component, \(b_i=|\operatorname{LCC}(\Gamma_i)|/|\Omega_i|\),
where \(\operatorname{LCC}(\cdot)\) returns the largest connected component and \(|\cdot|\) denotes spatial area. Applying the same statistic to the normal--normal residual maps used to construct \(\mathcal{R}_i^{\mathrm{N}}\) yields a set of normal area fractions \(\mathcal{B}_i^{\mathrm{N}}\). Given a high calibration quantile \(p_{\mathrm{area}}\) and a nonnegative margin \(\delta\), we define the spatial-coherence threshold as \(\theta_{i,\mathrm{area}}=\operatorname{Quantile}_{p_{\mathrm{area}}}(\mathcal{B}_i^{\mathrm{N}})+\delta\).
The spatial-coherence condition controls whether expert readers are admitted for anomaly detection. For localization, readers receive the calibrated overlay without this admission gate, so they can interpret the spatial pattern together with the original image. The controller below specifies the task-dependent use of these reader predictions.

\paragraph{Expert reader agents.}

The calibrated anomaly map exposes fine-grained defect cues that are difficult to capture through holistic MLLM reasoning alone. We use frozen expert reader agents to inject this localized evidence into the answer-generation process, thereby augmenting general-purpose MLLMs with explicit defect perception and spatial grounding. Let \(\mathcal{E}_i\) denote the set of expert readers that produce valid predictions for instance \(i\), and let \(e \in \mathcal{E}_i\) index an individual reader. Given the query image and its calibrated overlay, reader \(e\) produces \(o_{ie}=g_e(I_i,\widetilde{\mathbf{H}}_i,q_i,\mathcal{A})\in\mathcal{A}\),
where \(g_e\) denotes the frozen inference process of reader \(e\). By jointly observing the original query image and its calibrated spatial overlay, the readers can relate localized deviations to the semantic requirements of the question, specifically deciding whether a defect is present and identifying its location. This evidence-to-answer translation preserves the spatial grounding introduced by the defect expert while placing its output in the same prediction space as the base-agent answers. The reliability controller can therefore evaluate spatially grounded reader predictions alongside the complementary MLLM predictions and adaptively determine how strongly localized anomaly evidence should influence the final decision.

\subsection{Reliability Controller and Adaptive Aggregation}
\label{sec:controller}

The preceding modules provide complementary answer predictions, but their contributions depend on both the question type and the spatial evidence available for a query. The reliability controller accounts for these two factors by estimating global and task-wise source competence and modulating reader influence using spatial evidence. It operates on the predictions of frozen models; only the controller's aggregation statistics are estimated. The controller estimates these statistics from the unlabeled predictions of a batch of test queries, which in our experiments is the full MMAD evaluation set, so it operates transductively (Appendix~\ref{app:reference_budget}).

\paragraph{Query-level spatial evidence.}
Let \(\tau(i)\) denote the question type. Expert readers are used for defect localization and anomaly detection. Localization readers are queried whenever a calibrated map is available, including maps with no positive response. Detection readers are queried only when \(b_i>0\) and \(b_i\geq\theta_{i,\mathrm{area}}\). In either case, only valid option predictions enter \(\mathcal{E}_i\); unavailable or invalid reader outputs are omitted. For all other question types, \(\mathcal{E}_i=\varnothing\), and the controller aggregates the base agents alone.
For instances with at least one valid reader prediction, we define \(u_i=\max_{p\in\Omega_i}\widetilde h_i(p)\) for defect localization and \(u_i=b_i\) for anomaly detection, measuring response strength and spatial extent, respectively. To place them on a common relative scale, let \(\mathcal{V}_{\tau}=\{j:\tau(j)=\tau,\ \mathcal{E}_j\neq\varnothing\}\) and define
\begin{equation}
\label{eq:evidence_rank}
t_i=
\begin{cases}
\displaystyle\frac{1}{|\mathcal{V}_{\tau(i)}|}
\sum_{j\in\mathcal{V}_{\tau(i)}}\mathbf{1}[u_j\leq u_i],
&\mathcal{E}_i\neq\varnothing,\\[5pt]
0,&\mathcal{E}_i=\varnothing.
\end{cases}
\end{equation}
The first branch is well defined because its reference set contains \(i\). Equal statistics receive the same upper empirical rank. We then set
$\rho_i=1+\beta t_i,\beta\geq0$.
Thus \(\rho_i\in[1,1+\beta]\) amplifies admitted reader predictions according to relative evidence strength. The rank is a weighting statistic, not an estimate of the probability that an expert is correct. It is computed once and held fixed during reliability estimation.

\paragraph{Label-free source reliability.}
Let \(\mathcal{M}=\{1,\ldots,N\}\) and \(\mathcal{E}_i\) use disjoint source identifiers, and write \(\mathcal{P}_i=\mathcal{M}\cup\mathcal{E}_i\). All base agents provide valid predictions; reader abstentions contribute no observation. For each source \(v\), a row-stochastic matrix \(\mathbf C_v\) describes its option-confusion pattern: \([\mathbf C_v]_{\ell,z}\) associates a latent correct option \(\ell\) with an observed prediction \(z\). These indices refer to answer positions, not shared semantic classes across questions. Following the Dawid--Skene construction~\citep{dawid1979observer}, we combine source emissions multiplicatively, with a fixed uniform prior over the available answers \(\mathcal A_i\subseteq\mathcal A\).

To incorporate spatial evidence consistently into both reliability estimation and answer inference, define \(a_{iv}=1\) for base agents and \(a_{iv}=\rho_i\) for admitted readers. For an estimation set \(\mathcal D\), we maximize the regularized aggregation objective
\begin{equation}
\label{eq:controller_objective}
\begin{aligned}
\mathcal J_{\mathcal D}(\mathbf C)
={}&\sum_{i\in\mathcal D}\log\!\left[
\frac{1}{|\mathcal A_i|}\sum_{\ell\in\mathcal A_i}
\exp\!\left(\sum_{v\in\mathcal P_i}a_{iv}\log[\mathbf C_v]_{\ell,o_{iv}}\right)
\right] +\lambda\sum_{v}\sum_{\ell,z\in\mathcal A}\log[\mathbf C_v]_{\ell,z},
\qquad\lambda>0.
\end{aligned}
\end{equation}
This is an evidence-weighted surrogate for Dawid--Skene aggregation. With unit weights it recovers the corresponding regularized latent-label likelihood; with nonunit weights, the powered emissions define aggregation scores rather than normalized observation probabilities. The regularizer keeps every confusion-matrix entry positive.

Writing \(s_i(\ell)=\sum_{v\in\mathcal P_i}a_{iv}\log[\mathbf C_v]_{\ell,o_{iv}}\), the log-sum-exp identity expresses Eq.~\eqref{eq:controller_objective} as \(\mathcal J_{\mathcal D}(\mathbf C)=\max_{\pi}\mathcal F_{\mathcal D}(\pi,\mathbf C)\), where
\begin{equation}
\label{eq:controller_free_energy}
\begin{aligned}
\mathcal F_{\mathcal D}(\pi,\mathbf C)
={}&\sum_{i\in\mathcal D}\left[
\sum_{\ell\in\mathcal A_i}\pi_i(\ell)s_i(\ell)
+H(\pi_i)-\log|\mathcal A_i|\right]+\lambda\sum_v\sum_{\ell,z\in\mathcal A}\log[\mathbf C_v]_{\ell,z}.
\end{aligned}
\end{equation}
Here each \(\pi_i\) lies on the probability simplex over \(\mathcal A_i\), and \(H(\pi_i)=-\sum_{\ell\in\mathcal A_i}\pi_i(\ell)\log\pi_i(\ell)\) is its entropy. This formulation yields closed-form coordinate updates for the answer distributions and source matrices.

Let \(n_i(\ell)\) count valid votes for option \(\ell\). We initialize the latent-answer distribution by smoothed voting,
\begin{equation}
\label{eq:init}
\pi_i^{(0)}(\ell)=\frac{n_i(\ell)+\lambda}
{\sum_{z\in\mathcal A_i}(n_i(z)+\lambda)},\qquad \ell\in\mathcal A_i,
\end{equation}
and set \(\pi_i(\ell)=0\) for \(\ell\notin\mathcal A_i\). We then alternate the following updates.

\emph{Reliability update.}
Let \(\mathcal I_v=\{j\in\mathcal D:v\in\mathcal P_j\}\). For fixed \(\pi\), the coefficient of \(\log[\mathbf C_v]_{\ell,z}\) in Eq.~\eqref{eq:controller_free_energy} is \(\lambda+\sum_{j\in\mathcal I_v}a_{jv}\pi_j(\ell)\mathbf1[o_{jv}=z]\). Maximizing each row under its unit-sum constraint therefore gives
\begin{equation}
\label{eq:mstep}
[\mathbf C_v]_{\ell,z}
=\frac{\lambda+\sum_{j\in\mathcal I_v}a_{jv}\pi_j(\ell)\mathbf1[o_{jv}=z]}
{|\mathcal A|\lambda+\sum_{j\in\mathcal I_v}a_{jv}\pi_j(\ell)}.
\end{equation}
The same evidence weight used in answer inference therefore also weights that observation when estimating source competence. A row with no expected observations reduces to the uniform distribution.

\emph{Answer update.}
For fixed source matrices, maximizing the expected score plus entropy in Eq.~\eqref{eq:controller_free_energy} under the answer-simplex constraint gives
\begin{equation}
\label{eq:global_aggregation}
\begin{aligned}
s_i(\ell)&=\sum_{m\in\mathcal M}\log[\mathbf C_m]_{\ell,o_{im}}
+\rho_i\sum_{e\in\mathcal E_i}\log[\mathbf C_e]_{\ell,o_{ie}}, \quad \pi_i(\ell)=\frac{\exp s_i(\ell)}{\sum_{z\in\mathcal A_i}\exp s_i(z)},
\quad\ell\in\mathcal A_i.
\end{aligned}
\end{equation}
Each step thus maximizes one block of the same objective. With fixed evidence weights, \(\mathcal F_{\mathcal D}\) is nondecreasing and bounded above by zero, since \(H(\pi_i)\leq\log|\mathcal A_i|\) and every log matrix entry is nonpositive. Its values therefore converge, without a guarantee of global optimality (Appendix~\ref{app:controller_derivation}). The stopping criteria and controller hyperparameters are specified in Appendix~\ref{app:setup}.

\paragraph{Task-wise competence and final aggregation.}
We apply the same estimation procedure to the full batch to obtain \(\pi_i^{\mathrm{global}}\), and separately to each question type to obtain \(\pi_i^{\tau(i)}\). The global estimate pools observations across tasks, whereas the task-wise estimate captures differences in source competence. For types with fewer than \(n_{\min}\) instances, we use only the global estimate. Otherwise, we combine the two distributions using the normalized logarithmic pool
\begin{equation}
\label{eq:mix}
\pi_i^{\mathrm{final}}(\ell)=
\frac{[\pi_i^{\mathrm{global}}(\ell)]^{1-w}
[\pi_i^{\tau(i)}(\ell)]^{w}}
{\sum_{z\in\mathcal A_i}[\pi_i^{\mathrm{global}}(z)]^{1-w}
[\pi_i^{\tau(i)}(z)]^{w}},\qquad w\in[0,1].
\end{equation}
This pool interpolates the two log scores; it does not treat their overlapping observations as independent evidence. The final answer is \(\hat y_i=\arg\max_{\ell\in\mathcal A_i}\pi_i^{\mathrm{final}}(\ell)\), with ties resolved by option order. Together, task-wise estimation and query-dependent evidence weighting adapt the aggregation to complementary semantic and spatial predictions while leaving the underlying models frozen.

\section{Experiments and Analysis}
Our experiments test the argument of Section~\ref{sec:intro} in turn. Section~\ref{sec:exp-training} asks whether in-domain training is enough by comparing SiGMA with trained specialists and large-scale MLLMs. Section~\ref{sec:exp-roles} isolates what composition and the defect expert each contribute and tests whether composing MLLMs alone can close the perception gap. Section~\ref{sec:exp-evidence} examines how expert evidence should be calibrated and weighted, and Section~\ref{sec:exp-agents} reports the trade-off between accuracy and cost.

\begin{table*}[!t]
\centering\small
\resizebox{\textwidth}{!}{%
\begin{tabular}{l c cccc cc c}
\toprule
& \multicolumn{1}{c}{Anomaly}
& \multicolumn{4}{c}{Defect}
& \multicolumn{2}{c}{Object}
& \\
\cmidrule(lr){2-2}
\cmidrule(lr){3-6}
\cmidrule(lr){7-8}
Method
& Detection
& Classification
& Localization
& Description
& Analysis
& Classification
& Analysis
& Average \\
\midrule
Human (Expert)
& 95.2 & 75.0 & 92.3 & 83.3 & 94.2 & 86.1 & 80.4 & 86.7 \\
\midrule

\multicolumn{9}{l}{\textbf{\emph{Large-scale MLLMs}}} \\
InternVL2-76B~\citep{chen2024internvl}
& 68.3 & 54.2 & 56.7 & 66.3 & 80.5 & 86.4 & 82.9 & 70.8 \\

GPT-4o~\citep{hurst2024gpt4o}
& 68.6 & 65.8 & 55.6 & 73.2 & 83.4 & 95.0 & 82.8 & 74.9 \\

GPT-5-mini~\citep{singh2025gpt5}
& 64.1 & 67.4 & 69.1 & 79.0 & \underline{86.7} & 94.0 & 83.4 & 77.7 \\

Gemini-2.5-Flash~\citep{comanici2025gemini25}
& \textbf{83.4} & 69.9 & 63.3 & 76.4 & 81.6 & 94.0 & 82.0 & 78.7 \\

Gemini-2.5-Pro~\citep{comanici2025gemini25}
& \underline{83.1} & 73.9 & 67.2 & 80.0 & 86.3 & 94.9 & 83.1
& \underline{81.2} \\

\midrule
\multicolumn{9}{l}{\textbf{\emph{Individual frozen MLLMs (Selected)}}} \\
Phi-3.5-vision~\citep{abdin2024phi3}
& 55.8 & 38.0 & 44.6 & 58.8 & 77.6 & 77.3 & 75.2 & 61.0 \\

LLaVA-OneVision-7B~\citep{li2024llavaonevision}
& 56.1 & 55.6 & 45.8 & 65.8 & 75.5 & 91.3 & 81.5 & 67.4 \\

Qwen3-VL-4B-Thinking~\citep{bai2025qwen3vl}
& 66.7 & 53.4 & 54.5 & 64.5 & 72.5 & 88.5 & 78.9 & 68.4 \\

InternVL3-8B~\citep{zhu2025internvl3}
& 66.3 & 56.4 & 51.2 & 67.7 & 80.3 & 87.8 & 83.7 & 70.5 \\

MiMo-VL-7B-RL~\citep{yue2025mimovl}
& 68.7 & 56.0 & 57.3 & 70.1 & 77.1 & 88.5 & 77.8 & 70.8 \\

Qwen2.5-VL-7B~\citep{bai2025qwen25vl}
& 70.5 & 56.5 & 60.1 & 64.3 & 79.1 & 92.3 & 84.1 & 72.4 \\

GLM-4.1V-9B-Thinking~\citep{hong2025glm41v}
& 61.6 & 60.6 & 62.2 & 74.0 & 82.4 & 93.7 & 83.1 & 73.9 \\

Qwen3-VL-8B~\citep{bai2025qwen3vl}
& 73.5 & 62.4 & 60.8 & 68.8 & 79.5 & 92.7 & 83.9 & 74.5 \\

Qwen3.5-9B~\citep{qwen2026qwen35}
& 68.7 & 62.4 & 59.3 & 73.8 & 81.7 & 91.5 & 85.5 & 74.7 \\

Qwen3-VL-4B~\citep{bai2025qwen3vl}
& 74.1 & 59.7 & 63.6 & 70.1 & 80.0 & 92.4 & 83.4
& \bestSelectedMLLM{} \\

\midrule
\multicolumn{9}{l}{\textbf{\emph{Training-based methods}}} \\
AnomalyGPT~\citep{gu2024anomalygpt}
& 65.6 & 27.5 & 28.0 & 36.9 & 32.1 & 29.8 & 35.8 & 36.5 \\

AnomalyR1~\citep{chao2025anomalyr1}
& 60.6 & 63.6 & 70.1 & 80.5 & 85.3 & 92.5 & 86.2 & 77.0 \\

OmniAD~\citep{zhao2025omniad}
& 68.8 & \underline{78.8} & \underline{75.5} & 67.2 & 86.4
& 96.0 & 86.4 & 79.9 \\

JUDO~\citep{kang2026judo}
& 65.0 & 74.7 & 73.0 & \textbf{84.6} & \textbf{89.4}
& 94.0 & \underline{87.6} & \underline{81.2} \\
\midrule

\multicolumn{9}{l}{\textbf{\emph{Optimization-based methods}}} \\
Reason-IAD~\citep{chen2026reasoniad}
& 75.2 & 75.1 & 62.8 & 76.4 & 85.6
& \underline{96.3} & 84.6 & 79.4 \\
\midrule

\rowcolor{gray!12}
\textbf{\method{} (Ours, Training-free)}
& 82.6
& \textbf{79.2}
& \textbf{79.6}
& \underline{83.2}
& 86.4
& \textbf{97.3}
& \textbf{88.1}
& \textbf{\oursDK{}} \\
\rowcolor{gray!12}
\textit{Gap to best trained}
& 13.8$\uparrow$ & 0.4$\uparrow$ & 4.1$\uparrow$ & 1.4$\downarrow$ & 3.0$\downarrow$ & 1.3$\uparrow$ & 0.5$\uparrow$ & 4.0$\uparrow$\\
\bottomrule
\end{tabular}
\vspace{-2mm}}
\caption{MMAD performance comparison (accuracy, \%). In the one-shot setting, each MLLM receives one normal reference image alongside the query. Bold and underlined values mark the best and second-best results.}
\label{tab:compare}
\vspace{-2mm}
\end{table*}

\subsection{Experimental Setup}
\label{sec:experimental_setup}
We evaluate on MMAD~\citep{jiang2025mmad} (39,670 multiple-choice questions across seven types), reporting accuracy per type and the unweighted mean; stated gains and losses are absolute differences, denoted by \%. Appendix~\ref{app:setup} enumerates the ten candidate MLLMs and identifies the eight base agents and three expert readers in the main configuration. It also specifies the constituent datasets, baselines, one-shot protocol, and hyperparameters. The team and hyperparameters were selected on MMAD answers, so we also report the full ten-agent pool without subset selection. For a fixed configuration, the controller uses no labels and is fit transductively on the unlabeled evaluation predictions (Appendix~\ref{app:reference_budget}). Each MLLM receives one retrieved normal reference, while the expert uses a category-specific bank of at most 256 images.

\subsection{Is In-Domain Training Enough?}
\label{sec:exp-training}

\paragraph{Gains concentrate where training falls short.}
\autoref{tab:compare} compares SiGMA with trained specialists. SiGMA improves most on the two perception-centric tasks, raising anomaly detection by \textbf{13.8\%$\uparrow$} and defect localization by \textbf{4.1\%$\uparrow$} over the best training-based methods. The contrast is sharpest in anomaly detection, where all four training-based methods (\textbf{60.6--68.8\%}) fall below the frozen Qwen2.5-VL-7B (\textbf{70.5\%}), the base model of OmniAD and JUDO, whereas SiGMA reaches \textbf{82.6\%}. On the other five tasks SiGMA ranges from \textbf{3.0\%$\downarrow$} below to \textbf{1.3\%$\uparrow$} above the best training-based method, and on average it exceeds the strongest one, JUDO, by \textbf{4.0\%$\uparrow$}. SiGMA trails JUDO only in defect description (\textbf{1.4\%$\downarrow$}) and defect analysis (\textbf{3.0\%$\downarrow$}), the two tasks that rely most on domain-specific expression and reasoning. These results suggest that in-domain training mainly improves domain-specific language, while perceptual limitations persist.

\paragraph{Scale does not close the gap either.}
Among large-scale MLLMs, the best defect-localization accuracy is \textbf{69.1\%} (GPT-5-mini), \textbf{10.5\%$\downarrow$} below SiGMA. In anomaly detection, SiGMA comes within \textbf{0.8\%} of Gemini-2.5-Flash and \textbf{0.5\%} of Gemini-2.5-Pro while using open models of at most 9B parameters, and all three remain at least \textbf{11.8\%$\downarrow$} below human experts (\textbf{95.2\%}). SiGMA exceeds Gemini-2.5-Pro by \textbf{4.0\%$\uparrow$} on average and achieves the best result among all methods on four of the seven tasks. Relative to human experts, it is \textbf{1.5\%$\downarrow$} lower on average and higher on defect classification, object classification, and object analysis. The largest remaining gaps are anomaly detection (\textbf{12.6\%$\downarrow$}) and defect localization (\textbf{12.7\%$\downarrow$}), the capabilities that Section~\ref{sec:intro} identified as the shared weakness of MLLMs.

\paragraph{Selection, retrieval and references do not explain the gains.}
We test each alternative explanation in turn. First, the full ten-agent pool (\hyperref[fig:pool]{Figure~\ref*{fig:pool}(a)}), which involves no subset selection, reaches \textbf{84.9\%}, which is \textbf{3.7\%$\uparrow$} above JUDO and Gemini-2.5-Pro. Second, retrieval alone does not explain the gains, since adding the multimodal searcher to the best single MLLM yields \textbf{78.9\%} (\autoref{tab:ablation}), below both JUDO and Gemini-2.5-Pro (\textbf{81.2\%}). Third, limiting the defect expert to a single normal image ($K{=}1$), so that every component uses one normal reference as the one-shot baselines do, reaches \textbf{76.6\%} in defect localization, still above the best training-based method (\textbf{75.5\%}).

\subsection{What Does Each Role Contribute?}
\label{sec:exp-roles}

\autoref{tab:ablation} adds components cumulatively to the best single MLLM and ablates expert evidence and base-agent diversity. Relative to the best single MLLM, SiGMA gains \textbf{10.4\%$\uparrow$} on average and improves every task. The per-task breakdown shows how these gains divide between the two roles.

\Needspace{14\baselineskip}\begin{wraptable}[12]{r}{0.48\textwidth}
  \centering
  \scriptsize
  \setlength{\tabcolsep}{2pt}
  \begin{tabular}{@{}>{\raggedright\arraybackslash}p{0.55\linewidth}rrrr@{}}
    \toprule
    Configuration & AD & DL & Others & Avg. \\
    \midrule
    (a) Best single (Qwen3-VL-4B)
      & 74.1 & 63.6 & 77.1 & 74.8 \\
    (b) $+$ Multimodal searcher
      & 74.3 & 64.8 & 82.6 & 78.9 \\
    (c) $+$ Heterogeneous agents (MV)
      & 74.7 & 65.4 & 85.8 & 81.3 \\
    (d) $+$ Defect expert (MV)
      & 80.5 & 73.6 & 85.8 & 83.3 \\
    (e) $+$ Reliability controller
      & \textbf{82.6} & \textbf{79.6} & \textbf{86.8} & \textbf{\oursDK{}} \\
    \midrule
    (f) \method{} without defect expert
      & 75.1 & 65.8 & 86.7 & 82.0 \\
    (g) \method{} with one base agent
      & 82.0 & 79.5 & 82.6 & 82.1 \\
    \bottomrule
  \end{tabular}
  \vspace{-1mm}
  \caption{Component ablation result of \method{} (accuracy, \%). AD, anomaly detection; DL, defect localization; Others, mean of the other five tasks on MMAD; MV, majority vote.}
  \label{tab:ablation}
\end{wraptable}
\paragraph{Composition lifts semantic tasks but not perception.}
Composing heterogeneous agents (row c versus row b) improves the five tasks other than anomaly detection and defect localization by \textbf{3.2\%$\uparrow$} on average, but improves these two tasks by only \textbf{0.4\%$\uparrow$} and \textbf{0.6\%$\uparrow$}. Shared errors limit what aggregation can recover~\citep{krogh1994neural}.
 Removing the defect expert (row f) largely erases the perception gains; Appendix~\ref{app:ablation_full} details the effects of removing each role.

\paragraph{The defect expert targets the shared weakness.}
Adding the defect expert (row d versus row c) raises average accuracy by \textbf{2.0\%$\uparrow$}, entirely through anomaly detection and defect localization; Appendix~\ref{app:ablation_full} gives the per-task gains. Conversely, keeping the expert but reducing the base team to one agent (row g) preserves most of the perception gains but lowers the other five tasks by \textbf{4.2\%$\downarrow$} on average. Neither role substitutes for the other, consistent with the division of labor motivated in Section~\ref{sec:intro}. The reliability controller (row e) adds a further \textbf{1.9\%$\uparrow$}, which Section~\ref{sec:exp-evidence} analyzes.

\subsection{Does Expert Evidence Need Calibration?}
\label{sec:exp-evidence}

\vspace{-4mm}
\Needspace{15\baselineskip}\begin{wrapfigure}[12]{r}{0.48\textwidth}
  \vspace{-5mm}
  \centering
  \includegraphics[width=\linewidth,trim=0 4bp 0 18bp,clip]{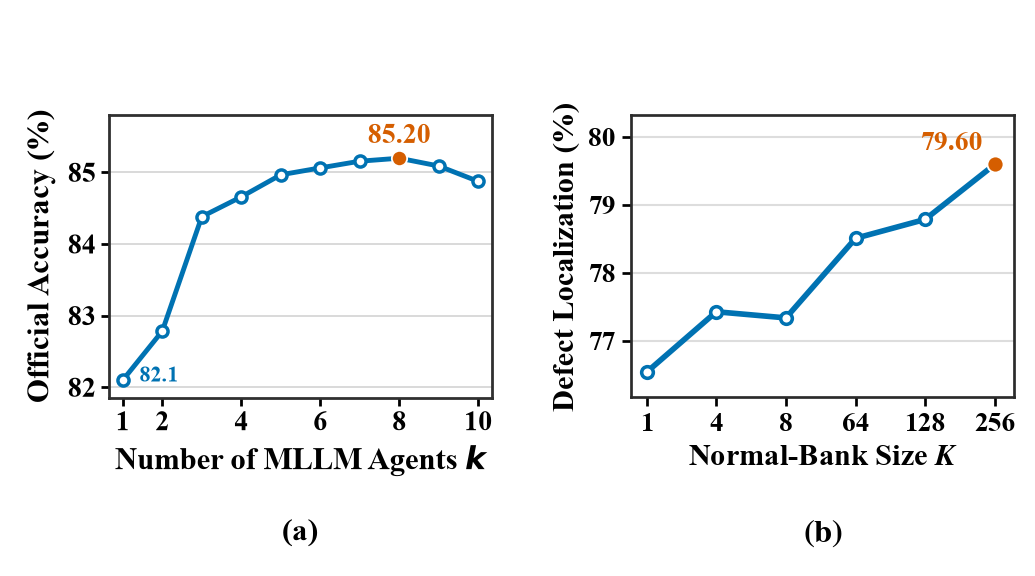}
  \setlength{\abovecaptionskip}{2pt}
  \vspace{-6mm}
  \caption{Scaling behavior of \method{}. \textbf{(a)} Best-subset accuracy by agent count $k$ (Appendix~\ref{app:reference_budget}). \textbf{(b)} Defect localization by normal-bank size $K$.}
  \label{fig:pool}
\end{wrapfigure}

Section~\ref{sec:intro} argued that expert evidence helps only if it is calibrated and weighted by task and query. \autoref{tab:evidence} in Appendix~\ref{app:evidence_ablation} tests both requirements.
\vspace{-0.3\baselineskip}
\paragraph{Calibration.}
With the base agents, controller, and reader route fixed, noise-robust calibration raises anomaly-detection accuracy from \textbf{73.5\%} to \textbf{80.9\%} and reduces the false-positive rate from \textbf{24.2\%} to \textbf{12.8\%} (\hyperref[tab:evidence]{Table~\ref*{tab:evidence}(a)}). Adding the coherence gate further raises accuracy to \textbf{82.6\%} and lowers the false-positive rate to \textbf{9.4\%}. Qualitative examples appear in Appendix~\ref{app:calibration_focus}.

\paragraph{Normal-bank size.}
\hyperref[fig:pool]{Figure~\ref*{fig:pool}(b)} varies the expert's normal-bank size $K$ while each MLLM receives one reference (Appendix~\ref{app:reference_budget}). Defect localization rises from \textbf{76.6\%} at $K{=}1$, where RAD shares the MLLMs' reference and skips noise-robust calibration, to \textbf{79.6\%} at $K{=}256$ with calibration. This comparison changes both bank size and calibration availability.

\paragraph{Task- and query-dependent weighting.}
\hyperref[tab:evidence]{Table~\ref*{tab:evidence}(b)} varies the aggregation rule with all prediction sources fixed. Replacing majority voting with label-free Dawid--Skene estimation adds \textbf{0.7\%$\uparrow$} on average, task-wise competence adds \textbf{0.8\%$\uparrow$}, and query-level evidence weighting adds \textbf{0.4\%$\uparrow$}, mainly in anomaly detection (\textbf{1.7\%$\uparrow$}).

\subsection{How Many Agents Are Enough?}
\label{sec:exp-agents}
\hyperref[fig:pool]{Figure~\ref*{fig:pool}(a)} reports the best subset found for each team size $k$. A three-agent team reaches \textbf{84.4\%}, \textbf{0.8\%$\downarrow$} below the eight-agent peak and \textbf{3.2\%$\uparrow$} above JUDO and Gemini-2.5-Pro. Accuracy declines slightly beyond eight agents, to \textbf{85.1\%} at $k{=}9$ and \textbf{84.9\%} at $k{=}10$. Appendix~\ref{app:reference_budget} specifies the corresponding model additions and unrounded results. Thus, enlarging the base-agent pool does not necessarily improve accuracy.
 Each configuration retains the three-reader route selected in Appendix~\ref{app:reader_count}, so the point $k{=}1$ can involve up to four MLLM calls when expert readers are admitted. Appendix~\ref{app:efficiency} records the available historical cost information. New MLLMs can join the pool without retraining, and the controller re-estimates their task-wise competence from unlabeled predictions.




\WFclear




\section{Conclusion and Limitations}
This work asked whether in-domain training is enough for fine-grained industrial anomaly understanding, and our results on MMAD suggest that it is not. Trained specialists trail human experts most in anomaly detection and defect localization, a perceptual weakness that different MLLMs share and that composing them does not remove. SiGMA therefore pairs heterogeneous frozen MLLM agents, which improve the semantic tasks, with a defect expert whose calibrated evidence improves detection and localization, and weighs each source by task and query without labels. It reaches 85.2\% average accuracy, 4.0\% above the strongest trained specialist and Gemini-2.5-Pro, and 84.4\% with three agents of at most 9B parameters.

Several limitations remain. Although our experiments use only MMAD, the benchmark spans four source datasets and seven tasks, offering some evidence of generalization across industrial settings. SiGMA requires no model training, and its source-reliability parameters can be estimated automatically from unlabeled predictions on new data. Our controller estimates source reliability from the unlabeled evaluation batch, making the full result transductive. Even without batch-level statistics, majority voting over the same sources reaches 83.3\% (\autoref{tab:ablation}, row d), exceeding every baseline in \autoref{tab:compare}. Aggregation cannot correct errors that every source shares, which SiGMA mitigates with a defect expert independent of MLLM perception, and without it defect localization drops by 13.8\% (\autoref{tab:ablation}). SiGMA still trails human experts by 12.6\% in anomaly detection and 12.7\% in defect localization, though by less than any trained method. Finally, several MLLMs add inference cost, but a three-agent team retains 84.4\% at 1.5\,s per question, against 1.2\,s for JUDO, and new MLLMs can join the team without retraining the existing models.

\FloatBarrier
\bibliographystyle{iclr2027_conference}
\bibliography{aaai2027}
\clearpage
\appendix
\raggedbottom
\section*{Appendix}

\section{Experimental Setup Details}
\label{app:setup}
 
\paragraph{Benchmark.}
MMAD~\citep{jiang2025mmad} covers seven question types, namely anomaly detection, four defect-level tasks (classification, localization, description, and analysis), and two object-level tasks (classification and analysis). Its images are drawn from MVTec-AD~\citep{bergmann2019mvtec}, VisA~\citep{zou2022spot}, MVTec-LOCO~\citep{bergmann2022beyond}, and GoodsAD~\citep{zhang2024pkugoodsad}. We evaluate 39,670 questions and report accuracy per question type together with the unweighted mean over the seven types.
 
\paragraph{MLLM pool and team.}
The candidate pool contains ten open-source MLLMs from eight model series, namely three Qwen3-VL variants (8B, 4B, and 4B-Thinking)~\citep{bai2025qwen3vl}, Qwen3.5-9B~\citep{qwen2026qwen35}, GLM-4.1V-9B-Thinking~\citep{hong2025glm41v}, Qwen2.5-VL-7B~\citep{bai2025qwen25vl}, MiMo-VL-7B-RL~\citep{yue2025mimovl}, InternVL3-8B~\citep{zhu2025internvl3}, LLaVA-OneVision-7B~\citep{li2024llavaonevision}, and Phi-3.5-vision~\citep{abdin2024phi3}. The main configuration uses eight base agents: GLM-4.1V-9B-Thinking, LLaVA-OneVision-7B, MiMo-VL-7B-RL-2508, Phi-3.5-vision-instruct, Qwen2.5-VL-7B-Instruct, Qwen3-VL-4B-Instruct, Qwen3-VL-8B-Instruct, and Qwen3.5-9B, selected as described in Appendix~\ref{app:reference_budget}. Its three expert readers are Qwen3-VL-8B, GLM-4.1V-9B-Thinking, and InternVL3-8B, chosen by the team-size study in Appendix~\ref{app:reader_count}. Readers add inference calls; InternVL3-8B is used as a reader but is not part of the eight-base-agent team. All MLLMs are served with vLLM~\citep{kwon2023efficient} in bfloat16, and the defect expert uses a frozen DINOv3~\citep{simeoni2025dinov3} backbone.
 
\paragraph{Baselines.}
The large-scale MLLMs are InternVL2-76B~\citep{chen2024internvl}, GPT-4o~\citep{hurst2024gpt4o}, GPT-5-mini~\citep{singh2025gpt5}, and Gemini-2.5-Flash and Gemini-2.5-Pro~\citep{comanici2025gemini25}. The training-based methods are AnomalyGPT~\citep{gu2024anomalygpt}, AnomalyR1~\citep{chao2025anomalyr1}, OmniAD~\citep{zhao2025omniad}, and JUDO~\citep{kang2026judo}, and the optimization-based method is Reason-IAD~\citep{chen2026reasoniad}. The ten pool models are also evaluated individually under the MMAD one-shot protocol, in which each model receives one normal reference alongside the query image.
 
\paragraph{Hyperparameters.}
\autoref{tab:hyperparams} lists all hyperparameters. They were selected on MMAD answers together with the agent team (Appendix~\ref{app:reference_budget}) and are fixed for all reported questions.
 
\begin{table}[h]
\centering
\small
\begin{tabular}{@{}llll@{}}
\toprule
Component & Symbol & Value & Role \\
\midrule
Base agents & -- & 1 & Retrieved normal references per agent input \\
Defect expert & -- & $\le 256$ & Size of the category-specific normal bank \\
\midrule
Noise-robust calibration
  & $p_{\mathrm{low}}$ & 0.99 & Quantile of normal residuals defining $\theta_{i,\mathrm{low}}$ \\
  & $\kappa$ & 3 & Expansion coefficient defining $\theta_{i,\mathrm{high}}$ \\
  & $p_{\mathrm{area}}$ & 0.99 & Quantile of normal area fractions defining $\theta_{i,\mathrm{area}}$ \\
  & $\delta$ & 0.01 & Margin of the spatial-coherence threshold \\
  & $\epsilon$ & $10^{-8}$ & Numerical floor in map normalization \\
\midrule
Reliability controller
  & $\lambda$ & 0.35 & Smoothing of vote initialization and confusion matrices \\
  & $\beta$ & 8 & Evidence-weight scale in $\rho_i = 1 + \beta t_i$ \\
  & $w$ & 0.35 & Weight of the task-wise posterior in the logarithmic pool \\
  & $n_{\min}$ & 400 & Minimum instances for a task-wise fit \\
  & -- & 40 & Maximum number of coordinate updates \\
  & -- & $10^{-6}$ & Tolerance on the mean posterior $\ell_1$ change \\
\bottomrule
\end{tabular}
\caption{Hyperparameters of \method{}, fixed for all reported results.}
\label{tab:hyperparams}
\end{table}

\section{Reference Data and Evaluation Protocol}
\label{app:reference_budget}

\paragraph{Textual domain knowledge.}The text branch of the multimodal searcher uses the domain knowledge provided with MMAD~\citep{jiang2025mmad} as its knowledge database. This resource supplies the product semantics, normal-structure descriptions, and defect-related descriptions used to construct the retrieved context $d_i$. It is separate from the normal-image reference gallery described below and does not include evaluation answers.\paragraph{Normal-reference gallery.}
For each product category, we construct the reference gallery from normal images in the training partition of its source dataset. MMAD draws its one-shot normal templates from these same underlying image pools~\citep{jiang2025mmad}. Our gallery is assembled from the category-level pool rather than restricted to MMAD's query-specific template lists, and contains at most 256 images per category. Evaluation query images are excluded from the gallery. Each MLLM receives one retrieved normal reference. At $K=1$, that image also constitutes the spatial expert's entire normal bank, and Noise-Robust Calibration is skipped. For $K>1$, the expert uses the larger bank for calibration while each MLLM still receives one reference.

\paragraph{Agent-subset analysis.}
To study the accuracy--cost trade-off, we evaluated candidate subsets of the ten base MLLMs on MMAD and report the highest-accuracy subset for each team size in \hyperref[fig:pool]{Figure~\ref*{fig:pool}(a)}. The MMAD evaluation answers were used to compare these subsets, including the eight-agent team yielding \textbf{85.2\%} in \autoref{tab:compare}. This benchmark-level choice of team is separate from the reliability controller: once a team is fixed, its source statistics and per-question predictions are computed without answer labels. The agent-count curve describes the evaluated combinations rather than an automatic team-selection procedure for unseen data. The archived SiGMA averages for $k{=}8,9,10$ are \textbf{85.2\%}, \textbf{85.1\%}, and \textbf{84.9\%}, respectively. Relative to the eight-agent team, $k{=}9$ adds InternVL3-8B and $k{=}10$ also includes Qwen3-VL-4B-Thinking.

\paragraph{Hyperparameter selection.}
We selected the reported hyperparameters by comparing candidate settings using the MMAD evaluation answers, following the same benchmark-level selection protocol as the agent subsets above. This includes the noise-robust calibration and spatial-coherence gate settings, together with the reliability controller settings listed in Appendix~\ref{app:setup}. Once selected, these values are fixed for all reported questions; answer labels are not used when estimating source reliability or aggregating predictions for a fixed configuration. Thus, the reported accuracy reflects configuration choices made on MMAD and should not be interpreted as performance under label-free hyperparameter selection on an unseen benchmark.

\paragraph{Unlabeled batch estimation.}
The empirical CDF used to normalize spatial-evidence statistics and the global and task-wise source-reliability estimates are computed from the unlabeled predictions and valid spatial statistics of the MMAD evaluation questions. We use no separate query calibration split. Ground-truth answers enter the team and hyperparameter comparisons above and the final accuracy calculation, but not the controller estimates for a fixed configuration. The reported controller therefore operates on the evaluation set as a transductive batch, rather than on isolated queries.

\paragraph{Reusing statistics for sequential inference.}
The controller can also be applied to incoming queries using statistics estimated from a preceding unlabeled batch, provided that the prediction sources and question types are the same. Let $\mathcal D_{\mathrm{hist}}$ denote that historical batch. After fitting the controller on $\mathcal D_{\mathrm{hist}}$, we retain its global and available task-wise source matrices, together with the empirical CDF $\widehat F_{\tau}^{\mathrm{hist}}$ of valid spatial-evidence statistics for each question type $\tau$. For a new query $i$, the searcher, base agents, and applicable expert readers run as usual. When an expert prediction is admitted and a historical CDF exists for $\tau(i)$, we set $t_i=\widehat F_{\tau(i)}^{\mathrm{hist}}(u_i)$ and $\rho_i=1+\beta t_i$; otherwise we use $t_i=0$ (and hence $\rho_i=1$ if an expert is admitted). We then compute the answer distributions with the stored source matrices using Eq.~\eqref{eq:global_aggregation}, apply the global--task-wise pool in Eq.~\eqref{eq:mix} when a task-wise fit is available, and do not update the matrices or CDF with the new query. This makes the controller's aggregation of a new query independent of other incoming queries after the historical statistics have been fitted. It does not change the normal-reference gallery or its query-specific spatial calibration. We have not evaluated this frozen-statistics protocol; the reported MMAD accuracies use the transductive batch protocol described above, and should not be interpreted as measured sequential-inference results.

\section{Reliability Controller: Derivation and Properties}
\label{app:controller_derivation}

\paragraph{Variational identity and answer update.}
For a fixed estimation set \(\mathcal D\), the observed votes, source availability, and weights \(a_{iv}\) remain fixed throughout optimization. Let \(\Delta(\mathcal A_i)\) denote the probability simplex over the available answers and define \(\bar\pi_i(\ell;\mathbf C)=\exp s_i(\ell)/\sum_{z\in\mathcal A_i}\exp s_i(z)\). Substituting this definition into the KL divergence yields the exact gap between the objective and the variational bound in Eqs.~\eqref{eq:controller_objective}--\eqref{eq:controller_free_energy}:
\begin{equation}
\mathcal J_{\mathcal D}(\mathbf C)-\mathcal F_{\mathcal D}(\pi,\mathbf C)
=\sum_{i\in\mathcal D}\operatorname{KL}\!\left(\pi_i\,\|\,\bar\pi_i(\cdot;\mathbf C)\right)\geq0.
\end{equation}
Equality holds exactly when each \(\pi_i=\bar\pi_i\). Thus \(\mathcal J_{\mathcal D}(\mathbf C)=\max_{\pi_i\in\Delta(\mathcal A_i)}\mathcal F_{\mathcal D}(\pi,\mathbf C)\), and the answer update in Eq.~\eqref{eq:global_aggregation} is the unique maximizer for fixed \(\mathbf C\). The uniform prior contributes the same constant to every available answer and cancels in the softmax normalization.

\paragraph{Reliability update.}
For fixed \(\pi\), define
\(N_{v\ell z}=\sum_{j\in\mathcal I_v}a_{jv}\pi_j(\ell)\mathbf1[o_{jv}=z]\).
Each confusion-matrix row solves
\begin{equation}
\max_{c_z>0,\ \sum_z c_z=1}\ \sum_{z\in\mathcal A}(N_{v\ell z}+\lambda)\log c_z.
\end{equation}
The Lagrange equations give \(c_z=(N_{v\ell z}+\lambda)/\sum_{z'}(N_{v\ell z'}+\lambda)\), which is Eq.~\eqref{eq:mstep}. Here \(\lambda\sum_z\log c_z\) corresponds to a symmetric Dirichlet regularizer with concentration \(1+\lambda\). In particular, expert weights must appear in both the numerator and denominator of the expected-count update. Using weighted answer scores with unweighted counts would not constitute coordinate maximization of Eq.~\eqref{eq:controller_free_energy}.

\paragraph{Optimization properties and termination.}
Positive smoothing makes every updated matrix entry strictly positive, so all log scores are finite. Each block update maximizes \(\mathcal F_{\mathcal D}\) with the other block fixed. Moreover, \(H(\pi_i)\leq\log|\mathcal A_i|\), all source weights are nonnegative, and every log matrix entry is nonpositive; hence \(\mathcal F_{\mathcal D}\leq0\). Its values therefore converge monotonically in exact arithmetic. After each answer update the variational bound is tight, so the corresponding values of \(\mathcal J_{\mathcal D}\) are also nondecreasing. These statements concern objective values, not uniqueness of the solution or convergence to a global maximum. We stop after at most 40 updates, or earlier when the mean change \(|\mathcal D|^{-1}\sum_{i\in\mathcal D}\|\pi_i^{(r+1)}-\pi_i^{(r)}\|_1\) is below \(10^{-6}\).

\paragraph{Interpretation.}
When \(a_{iv}=1\), the updates reduce to regularized Dawid--Skene estimation with a fixed uniform answer prior, known candidate sets, and missing expert observations omitted. For nonunit weights, our objective uses powers of source emissions as evidence-dependent scores. It is not the likelihood obtained by normalizing each powered emission over possible observations; that alternative introduces parameter-dependent normalization terms and would require a different reliability update. Accordingly, \(\pi_i\) is a normalized aggregation distribution, and we do not interpret it as a calibrated probability of correctness. The factorized score is a working approximation: sources can share errors, particularly when they use related models or common visual evidence. Smoothed-vote initialization anchors the latent options to observed answer positions, but does not guarantee recovery of the correct answers from unlabeled agreement.

\paragraph{Global--task-wise pooling.}
For a task with at least \(n_{\min}=400\) instances, Eq.~\eqref{eq:mix} is the unique minimizer
\begin{equation}
\underset{q\in\Delta(\mathcal A_i)}{\arg\min}\;
(1-w)\operatorname{KL}(q\|\pi_i^{\mathrm{global}})
+w\operatorname{KL}(q\|\pi_i^{\tau(i)}).
\end{equation}
Expanding the two divergences and applying a simplex Lagrange multiplier yields the normalized geometric mean. This characterizes the pool as a compromise between two positive distributions estimated at different levels of granularity. For smaller tasks no task-specific fit is used. Evidence ranks are computed once over the relevant valid instances of each question type and reused in both fits.

\paragraph{Evidence-rank boundaries.}
No empirical CDF is evaluated for a task with no admitted expert predictions. If all valid statistics for a task are identical, Eq.~\eqref{eq:evidence_rank} assigns them all rank one; it then provides no query-level discrimination. In particular, clipping can produce ties in the localization peak statistic. The empirical rank encodes relative response strength and does not certify evidence correctness.

\section{Expert-Reader Team Size}
\label{app:reader_count}

We assess how many expert readers are needed to translate calibrated spatial
evidence into answer predictions. Starting from a ten-reader candidate pool,
we compare fixed nested teams of different sizes on the full 39{,}670-question
MMAD evaluation. The eight base MLLM agents, retrieved domain knowledge,
reliability controller, and treatment of missing reader outputs are held
fixed. We evaluate two one-shot normal-reference settings:
\emph{Similar reference} uses a category-matched normal image selected by
the multimodal searcher according to query--reference visual similarity,
whereas \emph{Random reference} uses a randomly selected category-matched
normal image. Thus, the table columns distinguish reference-selection
strategies, and the rows vary the number of expert readers.
The selected three-reader team comprises Qwen3-VL-8B,
GLM-4.1V-9B-Thinking, and InternVL3-8B.

\begin{table}[H]
\centering
\small
\setlength{\tabcolsep}{18pt}
\renewcommand{\arraystretch}{1.16}
\begin{tabular}{rcc}
\toprule
Readers & \makecell[c]{Similar reference\\accuracy (\%)}
        & \makecell[c]{Random reference\\accuracy (\%)} \\
\midrule
1 & \textbf{84.7} & \textbf{84.7} \\
\rowcolor{gray!12}
\textbf{3} & \textbf{85.2} & \textbf{85.2} \\
4 & \textbf{85.1} & \textbf{85.0} \\
5 & \textbf{85.2} & \textbf{85.1} \\
10 & \textbf{85.2} & \textbf{85.0} \\
\bottomrule
\end{tabular}
\caption{Expert-reader team-size ablation under a fixed nested expansion.
Accuracy is measured on the full MMAD Benchmark.}
\label{tab:reader_count}
\end{table}

With similarity-based references, increasing the team from one to three
readers raises accuracy from \textbf{84.7\%} to \textbf{85.2\%}, a
\textbf{0.5\%$\uparrow$} gain at the reported precision. The random-reference
setting shows the same rounded change, from \textbf{84.7\%} to
\textbf{85.2\%}. Performance then saturates: with similarity-based
references, the three-, five-, and ten-reader teams all report
\textbf{85.2\%} after rounding, while three readers achieve the highest
unrounded accuracy in both reference settings. Additional readers do not
produce a consistent gain. We use three readers as the best tested choice
under this nested expansion; this does not establish optimality among all
possible three-reader subsets.

\section{Per-Task Component Ablation}
\label{app:ablation_full}

\autoref{tab:ablation_full} reports all seven question types for the configurations summarized in \autoref{tab:ablation}. Rows (b) to (g) use the same retrieved context, and every row that includes the defect expert uses the same three readers and calibration settings (Appendix~\ref{app:setup}). Rows (c) and (d) aggregate predictions by unweighted majority vote, whereas rows (e) to (g) use the reliability controller.

\paragraph{Cumulative additions.}
Retrieval and composition act mainly on the five tasks other than anomaly detection and defect localization. Together they raise these tasks by \textbf{8.7\%$\uparrow$} on average, led by defect classification (\textbf{17.9\%$\uparrow$}) and defect description (\textbf{12.3\%$\uparrow$}), but raise anomaly detection and defect localization by only \textbf{0.6\%$\uparrow$} and \textbf{1.8\%$\uparrow$}. The defect expert shows the opposite pattern. It raises anomaly detection by \textbf{5.8\%$\uparrow$} and defect localization by \textbf{8.2\%$\uparrow$} and, because the readers act only on these two tasks, leaves the other five unchanged. The reliability controller then improves every task. Its largest gains are in defect localization (\textbf{6.0\%$\uparrow$}) and anomaly detection (\textbf{2.1\%$\uparrow$}), the two tasks on which it also weights reader predictions by evidence quality, and its gains elsewhere range from \textbf{0.8\%$\uparrow$} to \textbf{1.6\%$\uparrow$}.

\paragraph{Removing one role.}
Removing the defect expert from the full system (row f) lowers anomaly detection by \textbf{7.5\%$\downarrow$} and defect localization by \textbf{13.8\%$\downarrow$}, while the other five tasks change by at most \textbf{0.3\%}. These small changes arise because the global reliability estimate pools observations across tasks. Reducing the base team to one agent (row g) has the reverse effect. Anomaly detection and defect localization fall by only \textbf{0.6\%$\downarrow$} and \textbf{0.1\%$\downarrow$}, whereas the other five tasks fall by \textbf{2.5\%$\downarrow$} to \textbf{7.2\%$\downarrow$}, most in defect classification. On these five tasks row (g) equals row (b), because the single base agent is then the only prediction source. The base agents and the defect expert therefore cover complementary tasks, and neither can replace the other.

\begin{table}[t]
\centering
\scriptsize
\setlength{\tabcolsep}{2.5pt}
\renewcommand{\arraystretch}{1.14}
\begin{tabular*}{\linewidth}{@{\extracolsep{\fill}}l>{\raggedright\arraybackslash}m{0.30\linewidth}*{8}{r}@{}}
\toprule
 & & Anomaly & \multicolumn{4}{c}{Defect} & \multicolumn{2}{c}{Object} & \\
\cmidrule(lr){3-3}\cmidrule(lr){4-7}\cmidrule(lr){8-9}
 & Configuration & Det. & Cls. & Loc. & Desc. & Anal. & Cls. & Anal. & Avg. \\
\midrule
(a) & Best single MLLM (Qwen3-VL-4B) & 74.1 & 59.7 & 63.6 & 70.1 & 80.0 & 92.4 & 83.4 & 74.8 \\
(b) & \quad + Multimodal searcher & 74.3 & 72.0 & 64.8 & 78.8 & 82.1 & 94.8 & 85.5 & 78.9 \\
(c) & \quad + Heterogeneous agents (majority vote) & 74.7 & 77.6 & 65.4 & 82.4 & 85.6 & 96.3 & 87.1 & 81.3 \\
(d) & \quad + Defect expert (majority vote) & 80.5 & 77.6 & 73.6 & 82.4 & 85.6 & 96.3 & 87.1 & 83.3 \\
(e) & \quad + Reliability controller (SiGMA) & 82.6 & 79.2 & 79.6 & 83.2 & 86.4 & 97.3 & 88.1 & 85.2 \\
\midrule
(f) & SiGMA without the defect expert & 75.1 & 78.9 & 65.8 & 83.0 & 86.2 & 97.2 & 88.0 & 82.0 \\
(g) & SiGMA with one base agent & 82.0 & 72.0 & 79.5 & 78.8 & 82.1 & 94.8 & 85.5 & 82.1 \\
\bottomrule
\end{tabular*}
\caption{Per-task ablation on MMAD (accuracy, \%). Rows (a) to (e) add components cumulatively, while rows (f) and (g) ablate expert evidence and base-agent diversity, respectively. Expert readers act only on anomaly detection and defect localization, so row (d) equals row (c) on the other five tasks. Row (g) uses Qwen3-VL-4B and retains the three expert readers.}
\label{tab:ablation_full}
\end{table}

\FloatBarrier
\Needspace{22\baselineskip}
\section{Using Expert Evidence}
\label{app:evidence_ablation} \autoref{tab:evidence} provides the detailed comparisons analyzed in Section~\ref{sec:exp-evidence}: part (a) varies evidence routing, map calibration, and the coherence gate; part (b) varies aggregation with the prediction sources fixed.

\begin{table}[H]
\centering
\scriptsize
\setlength{\tabcolsep}{3pt}
\begin{minipage}[t]{0.50\linewidth}
\centering
\begin{tabular}{lccc}
\toprule
Use of expert evidence & AD & DL & FPR$\downarrow$ \\
\midrule
No defect expert & 75.1 & 65.8 & 17.8 \\
Raw overlay on agent inputs & 69.8 & 75.0 & 30.6 \\
Raw map through readers & 73.5 & 76.8 & 24.2 \\
Calibrated map through readers & 80.9 & 79.6 & 12.8 \\
\quad + Coherence gate (SiGMA) & 82.6 & 79.6 & 9.4 \\
\bottomrule
\end{tabular}\\[3pt]
(a) Supplying expert evidence
\end{minipage}\hfill
\begin{minipage}[t]{0.48\linewidth}
\centering
\begin{tabular}{lccc}
\toprule
Aggregation rule & AD & DL & Avg. \\
\midrule
Majority vote & 80.5 & 73.6 & 83.3 \\
Dawid--Skene, global & 80.7 & 76.3 & 84.0 \\
\quad + Task-wise competence & 80.9 & 79.1 & 84.8 \\
\quad + Evidence weight (SiGMA) & 82.6 & 79.6 & 85.2 \\
\bottomrule
\end{tabular}\\[3pt]
(b) Aggregating predictions
\end{minipage}
\caption{Using expert evidence (accuracy, \%). (a) Ways of supplying anomaly maps, with the base agents and controller fixed. FPR is the fraction of normal images judged defective in anomaly-detection questions. (b) Aggregation rules, with all prediction sources fixed. AD and DL denote anomaly detection and defect localization.}
\label{tab:evidence}
\end{table}
\FloatBarrier

\paragraph{Routing raw evidence.}
\hyperref[tab:evidence]{Table~\ref*{tab:evidence}(a)} shows why a localization signal cannot simply be overlaid on the base-agent inputs. Relative to no expert, the raw overlay raises defect-localization accuracy from 65.8\% to 75.0\%, but lowers anomaly-detection accuracy from 75.1\% to 69.8\% and raises the false-positive rate on normal images from 17.8\% to 30.6\%. Passing the raw map through expert readers instead improves detection to 73.5\% and reduces false positives to 24.2\%. Routing limits the harm, yet raw evidence still performs worse than no expert on these two detection measures.

\paragraph{Calibration and gating.}
Calibrating the reader map raises detection by 7.4\% and localization by 2.8\% relative to the raw-reader row, while reducing the false-positive rate by 11.4\%. The spatial-coherence gate then raises detection from 80.9\% to 82.6\% and reduces false positives from 12.8\% to 9.4\%; localization remains at 79.6\%. This pattern matches the task-dependent design in Section~\ref{sec:controller}: the gate controls reader admission for anomaly detection, while localization can use the calibrated overlay without that test. Relative to no expert, the complete route gains 7.5\% in detection and 13.8\% in localization, and reduces the false-positive rate by 8.4\%.

\paragraph{Combining fixed prediction sources.}
\hyperref[tab:evidence]{Table~\ref*{tab:evidence}(b)} fixes the prediction sources and varies only aggregation. Global Dawid--Skene estimation raises average accuracy from 83.3\% under majority vote to 84.0\%, and task-wise competence raises it to 84.8\%. Query-level evidence weighting brings the average to 85.2\%, with a 1.7\% detection gain over task-wise competence. The two table parts therefore test distinct choices; their intermediate rows should be compared within each part. These gains are measured on MMAD with a team and hyperparameters selected using MMAD answers (Appendix~\ref{app:reference_budget}), rather than on a held-out benchmark.

\section{Training, Inference, and Serving Cost}
\label{app:efficiency}

\autoref{tab:efficiency_cost} compares reported accuracy, in-domain training
requirements, and inference cost across retrieval-guided, optimization-based,
and trained agentic approaches.

\begin{table}[H]
\centering
\small
\setlength{\tabcolsep}{3pt}
\renewcommand{\arraystretch}{1.16}
\begin{tabular}{@{}>{\raggedright\arraybackslash}m{0.16\linewidth}
                  >{\centering\arraybackslash}m{0.10\linewidth}
                  >{\centering\arraybackslash}m{0.17\linewidth}
                  >{\raggedright\arraybackslash}m{0.35\linewidth}
                  >{\centering\arraybackslash}m{\dimexpr0.22\linewidth-8\tabcolsep\relax}@{}}
\toprule
Method & \makecell{Accuracy\\(\%)} & \makecell{In-domain\\training/\\optimizing}
       & Inference and token cost & \makecell{Latency\\(s/sample)} \\
\midrule
Echo & 77.3 & $\times$
     & Retrieval and expert-guided MLLM reasoning & 1.51 \\
\addlinespace[4pt]
Reason-IAD & 79.4 & $\checkmark$
           & 10 latent-reasoning iterations & --- \\
\addlinespace[4pt]
JUDO & 81.2 & $\checkmark$
     & Single-pass inference with a specialized MLLM & 1.23 \\
\addlinespace[4pt]
AgentIAD & 82.9 & $\checkmark$
         & Multi-round reasoning with adaptive tool use & 2.90 \\
\addlinespace[4pt]
\rowcolor{gray!12}
\textbf{\method{}} & \textbf{85.2} & $\times$
 & Parallel agent inference and CPU-based reliability aggregation;
   25.04--228.44 tokens/question & 1.88 \\
\bottomrule
\end{tabular}
\caption{Accuracy and computational cost. $\checkmark$/$\times$ indicate
whether in-domain model-parameter training is required.
Baseline accuracies follow the original studies~\citep{chen2025echo,chen2026reasoniad,kang2026judo,miao2025agentiad}:
Echo reports five tasks on MVTec-AD and VisA, and AgentIAD uses a
6{,}400-sample MMAD evaluation split. Latencies for Echo, JUDO, and
\method{}, and token usage for \method{}, are measured in our evaluation.
AgentIAD's latency is reported on 1{,}600 Brain Tumor MRI samples at batch
size one. Evaluation settings differ across rows. A dash denotes an
unavailable measurement.}
\label{tab:efficiency_cost}
\end{table}

\method{} avoids in-domain parameter updates, while its inference cost scales
with the number of participating models. The full configuration uses eight
base agents and up to three expert readers, with a normal bank of at most
256 images. Controller statistics are estimated from the unlabeled evaluation
batch (Appendix~\ref{app:reference_budget}). These operations, along with
reference retrieval and spatial calibration, account for its test-time cost.

\begin{table}[H]
\centering
\small
\setlength{\tabcolsep}{4pt}
\begin{tabular}{@{}lcccc@{}}
\toprule
Configuration & MLLM calls & Latency (s) & Peak mem. (GiB) & Avg. acc. (\%) \\
\midrule
Single 7B MLLM & 1 & 1.2 & 15.2 & 81.2 \\
SiGMA, three base agents & 3 + up to 3 & 1.5 & 85 & 84.4 \\
SiGMA, eight base agents & 8 + up to 3 & 1.88 & 105 & 85.2 \\
\bottomrule
\end{tabular}
\caption{Inference cost per question on 4$\times$ NVIDIA L20 GPUs. Peak memory is reported across the four GPUs, not per device. Reader calls occur only for admitted anomaly-detection and defect-localization questions. Latency includes retrieval and the defect expert.}
\label{tab:cost}
\end{table}

\paragraph{Four-GPU serving and cost controls.}
On four L20 GPUs, eight base agents reach 85.2\% accuracy at 1.88~s per
question and 105~GiB peak memory; three agents reach 84.4\% at 1.5~s and
85~GiB (\autoref{tab:cost}). The smaller team saves 0.38~s and 20~GiB
for a 0.8-point accuracy decrease. Both retain up to three readers for
applicable detection and localization questions; the spatial-coherence gate
further limits detection calls. Parallel base-agent inference and conditional
readers are included in these measurements. Caching normal-reference features
and quantizing MLLM weights remain unevaluated cost-reduction options.

\section{Qualitative Effects of Noise-Robust Calibration}
\label{app:calibration_focus} 
\begin{figure}[H]
\centering
\small
\setlength{\tabcolsep}{0pt}
\renewcommand{\arraystretch}{1.12}
\begin{tabular*}{\linewidth}{@{\extracolsep{\fill}}ccccc@{}}
\toprule
Query & Normal reference & Raw map & Calibrated map & Overlay \\
\midrule
\includegraphics[width=0.18\linewidth,height=0.70in,keepaspectratio]{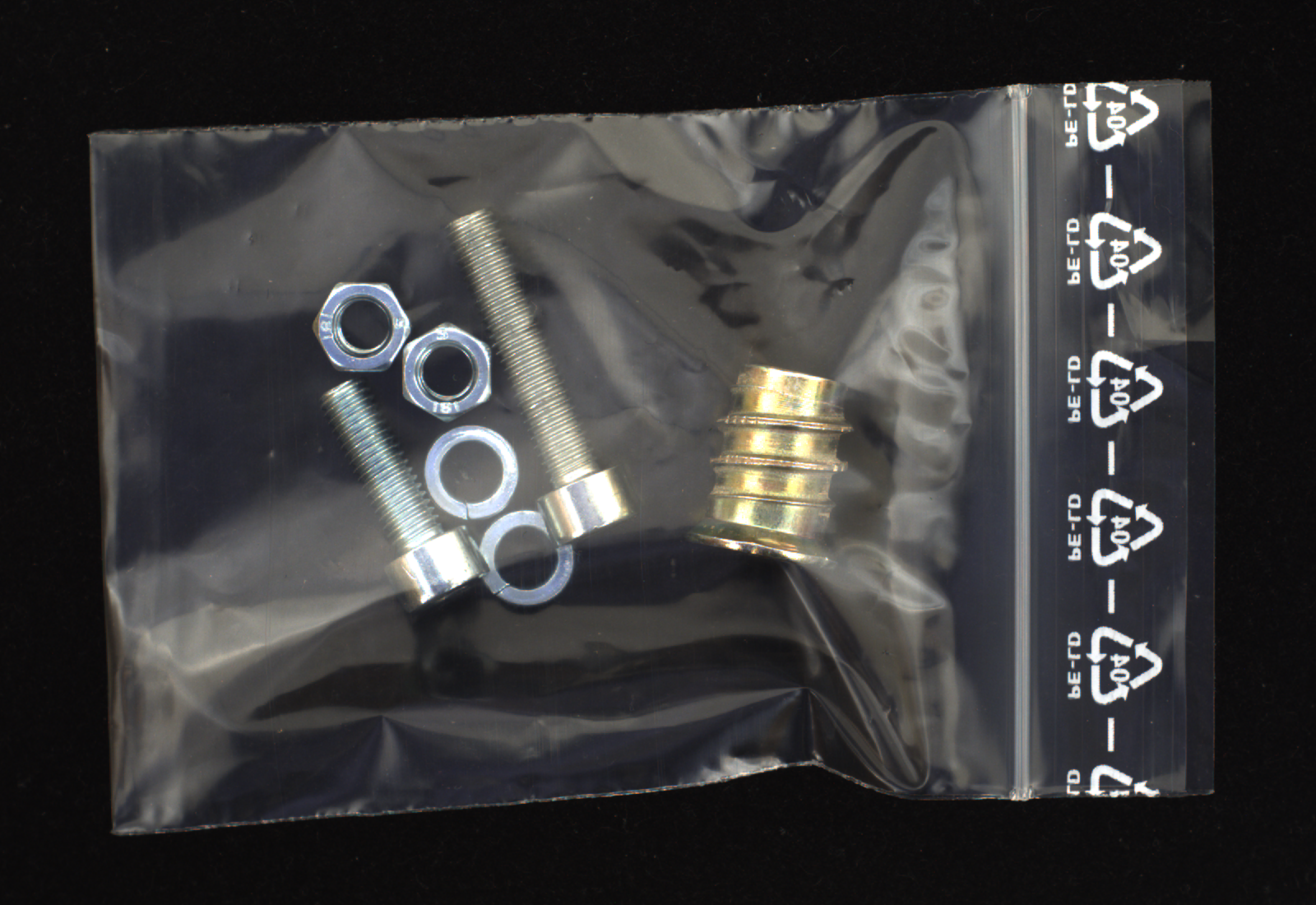} & \includegraphics[width=0.18\linewidth,height=0.70in,keepaspectratio]{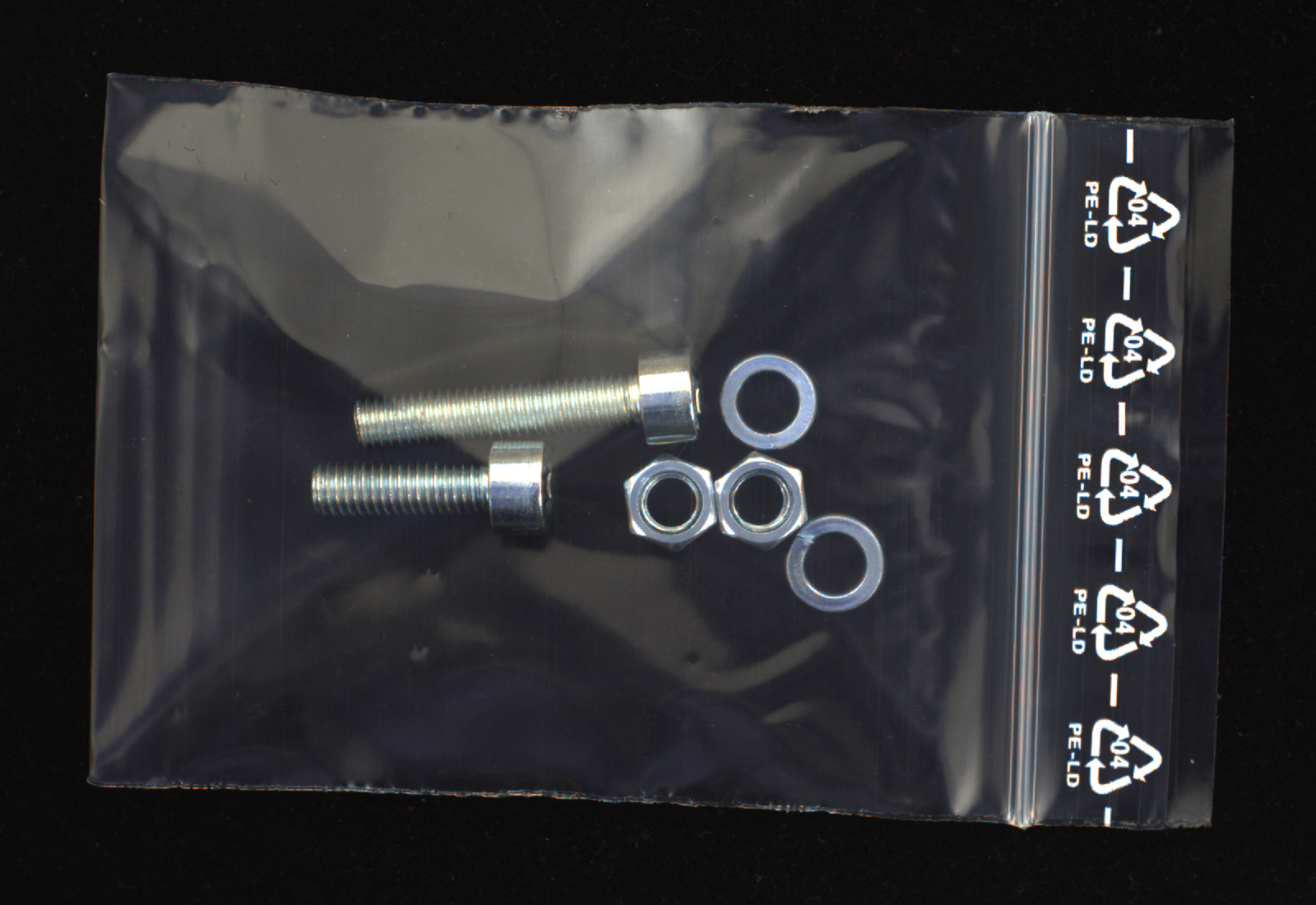} & \includegraphics[width=0.18\linewidth,height=0.70in,keepaspectratio]{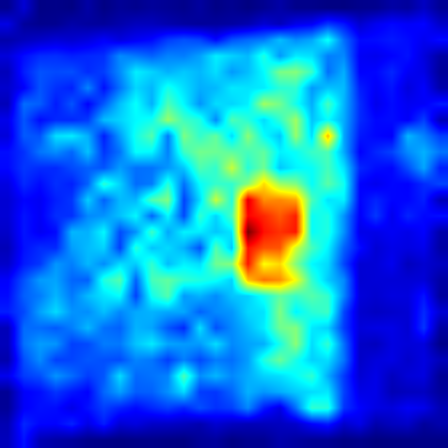} & \includegraphics[width=0.18\linewidth,height=0.70in,keepaspectratio]{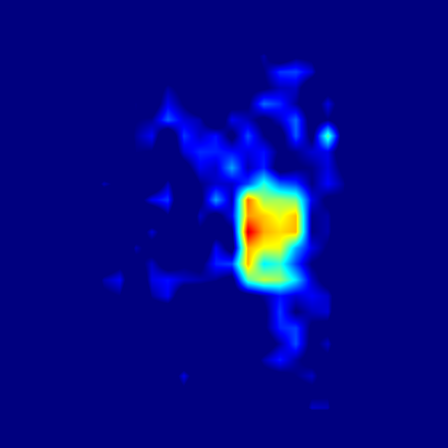} & \includegraphics[width=0.18\linewidth,height=0.70in,keepaspectratio]{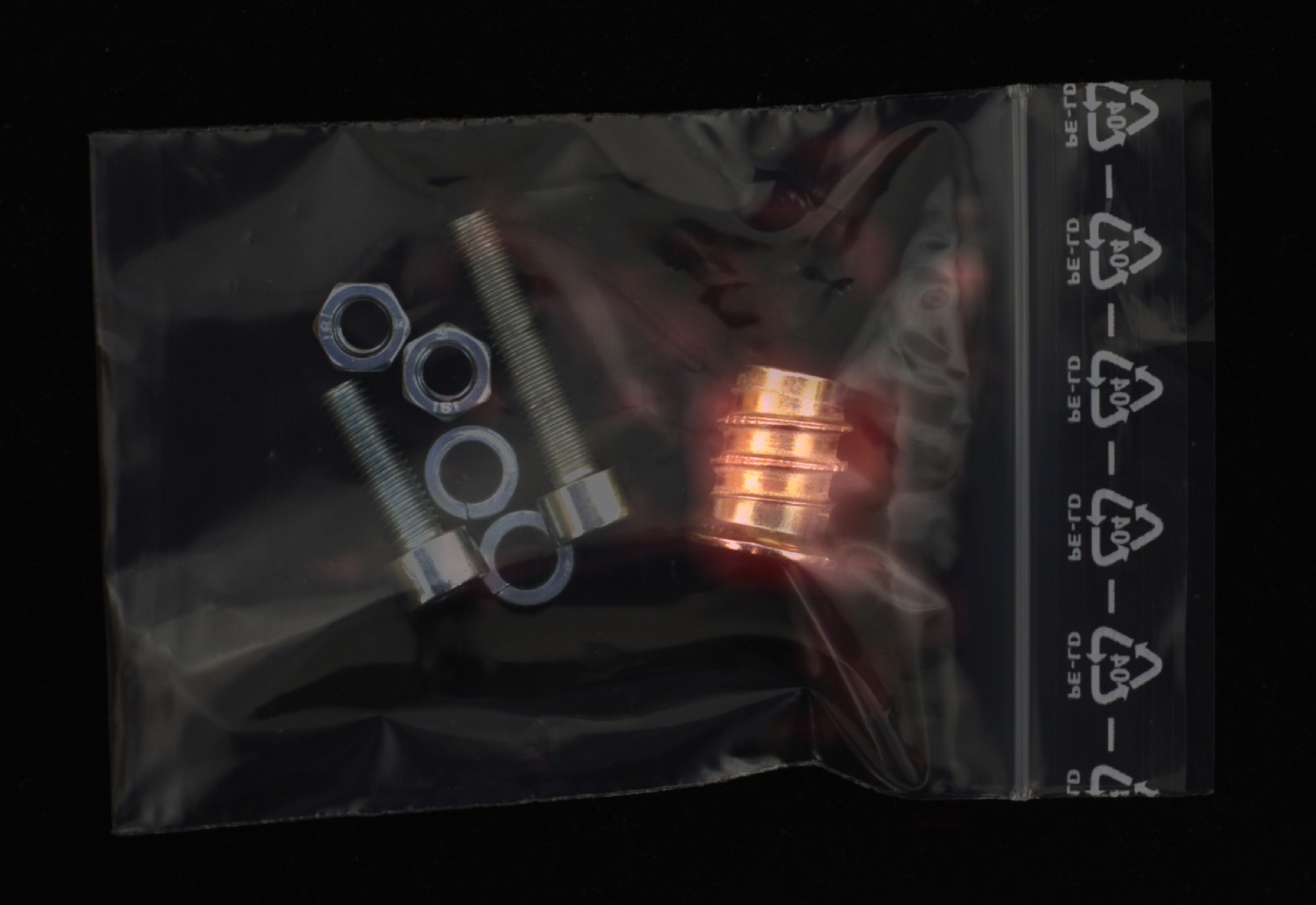} \\
\includegraphics[width=0.18\linewidth,height=0.70in,keepaspectratio]{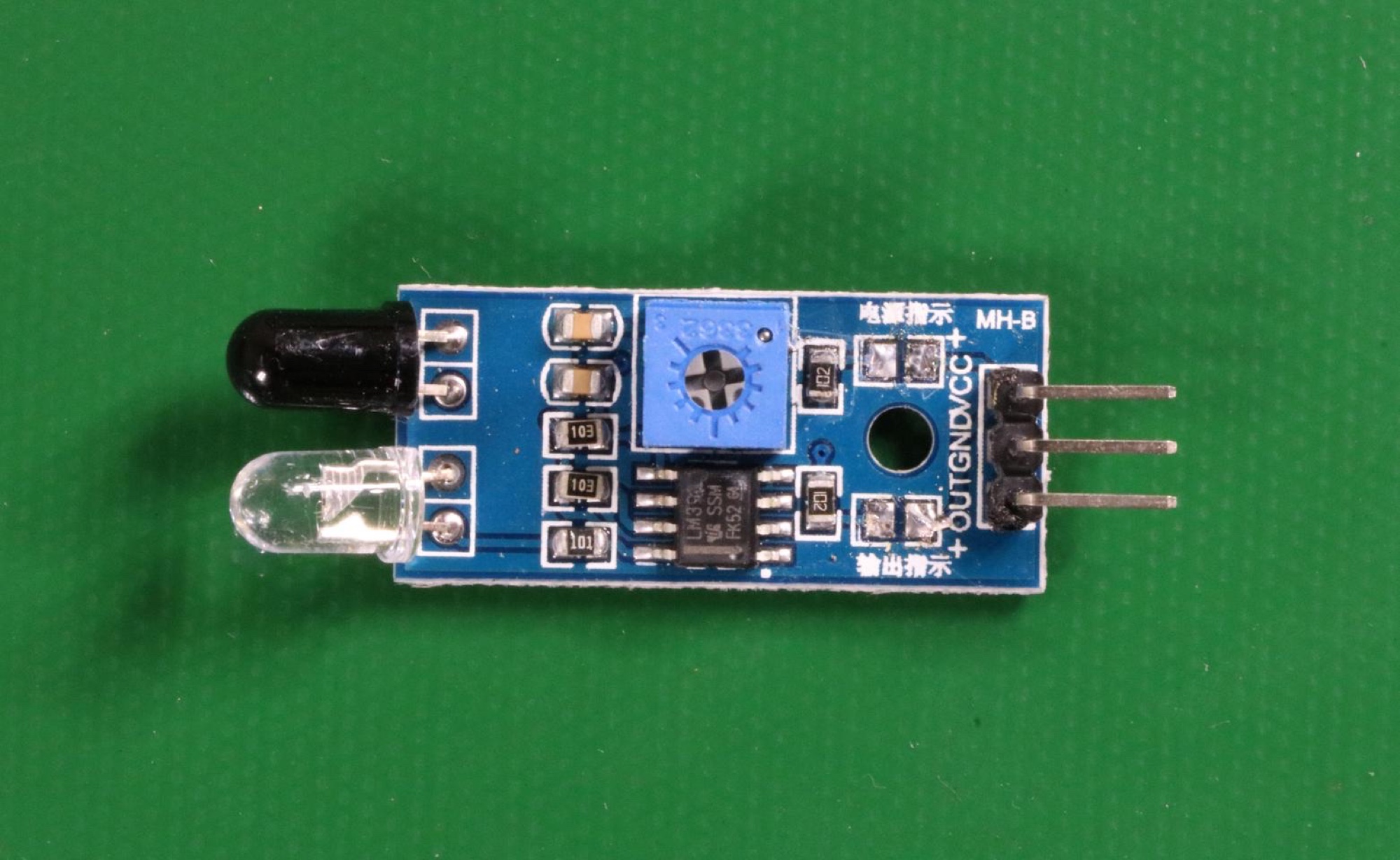} & \includegraphics[width=0.18\linewidth,height=0.70in,keepaspectratio]{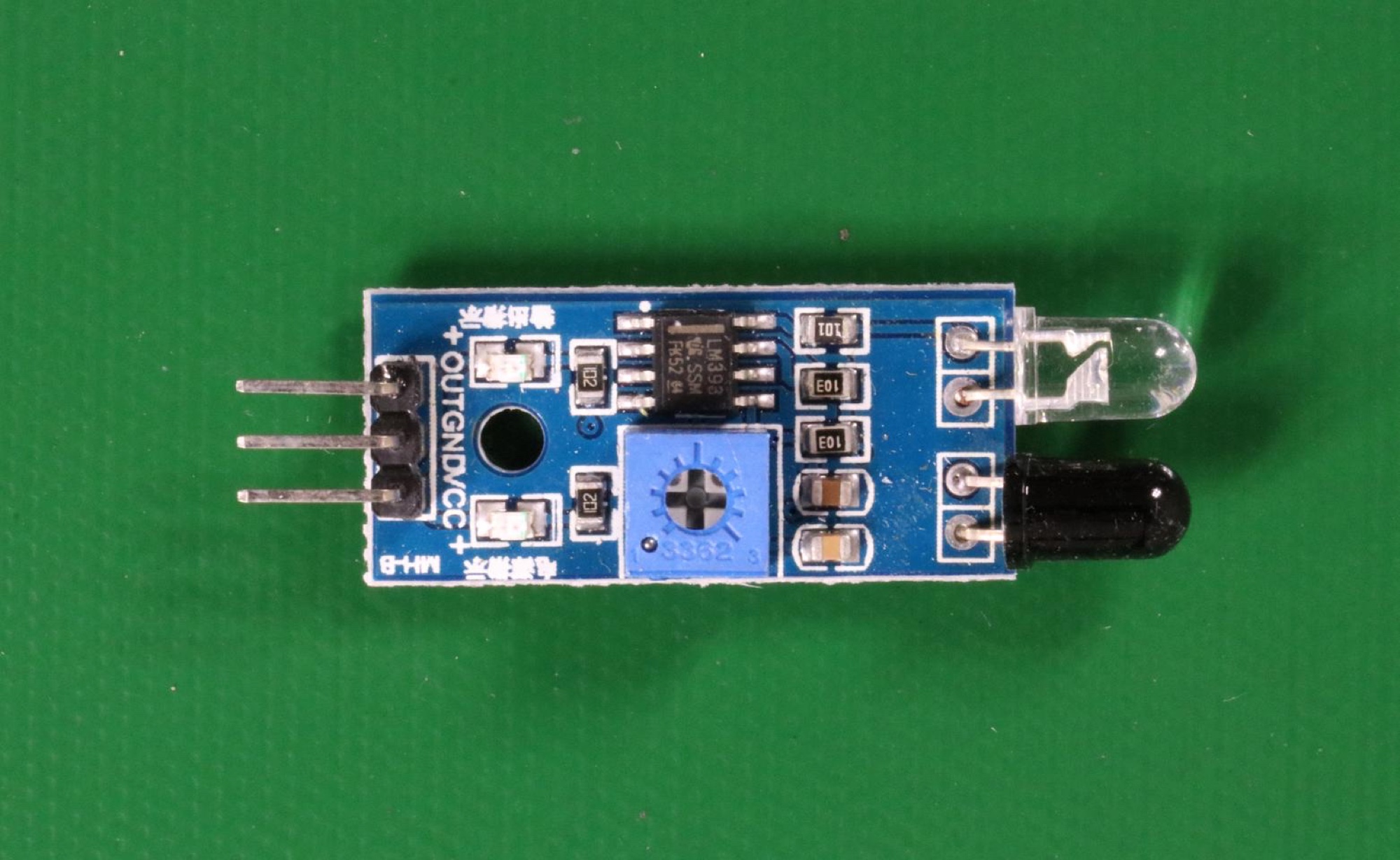} & \includegraphics[width=0.18\linewidth,height=0.70in,keepaspectratio]{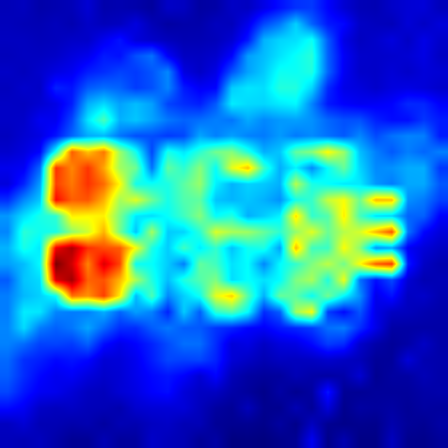} & \includegraphics[width=0.18\linewidth,height=0.70in,keepaspectratio]{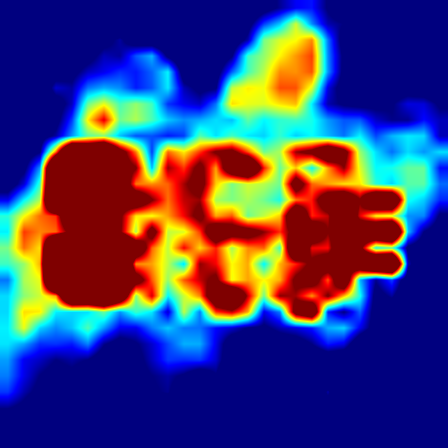} & \includegraphics[width=0.18\linewidth,height=0.70in,keepaspectratio]{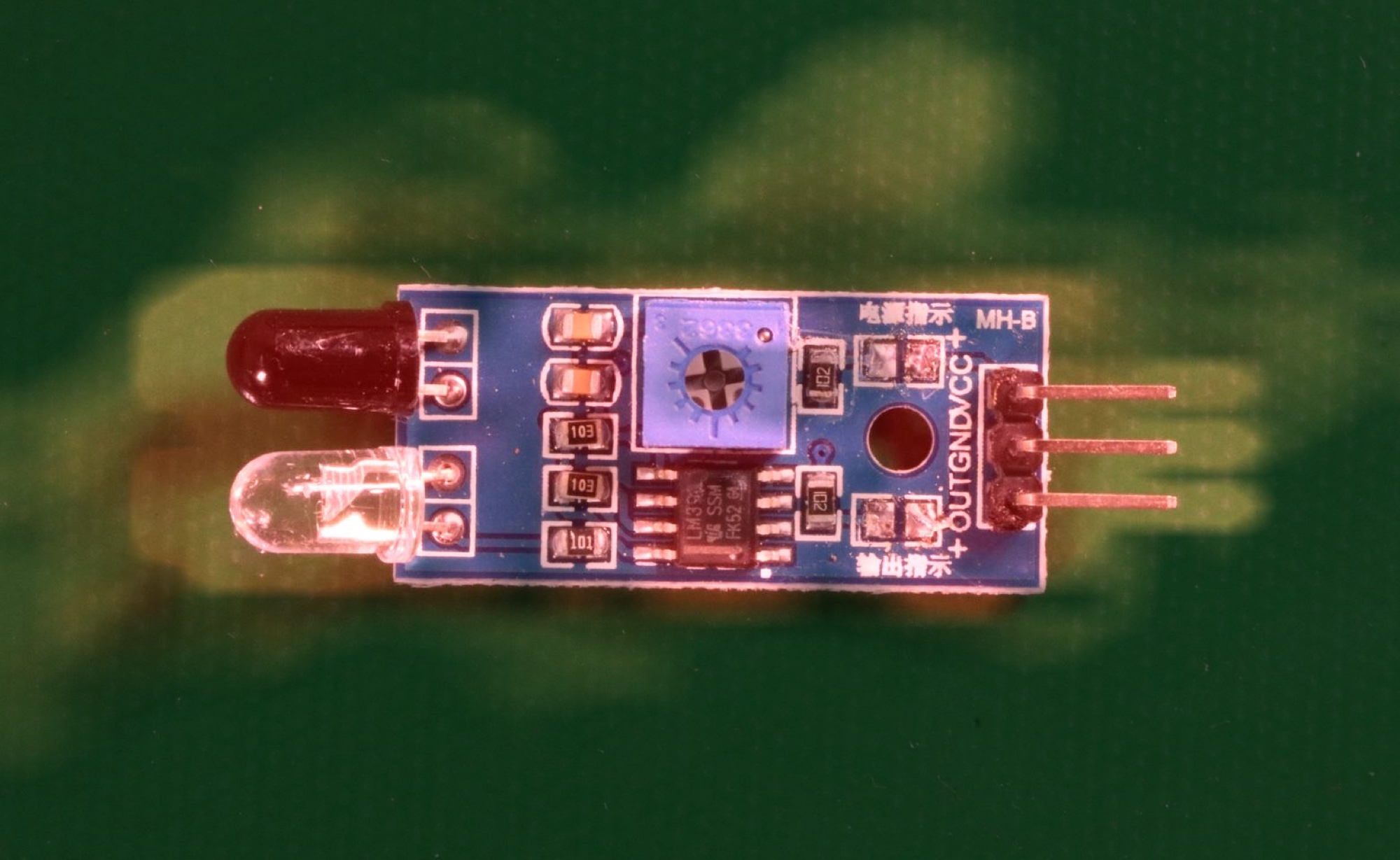} \\
\includegraphics[width=0.18\linewidth,height=0.70in,keepaspectratio]{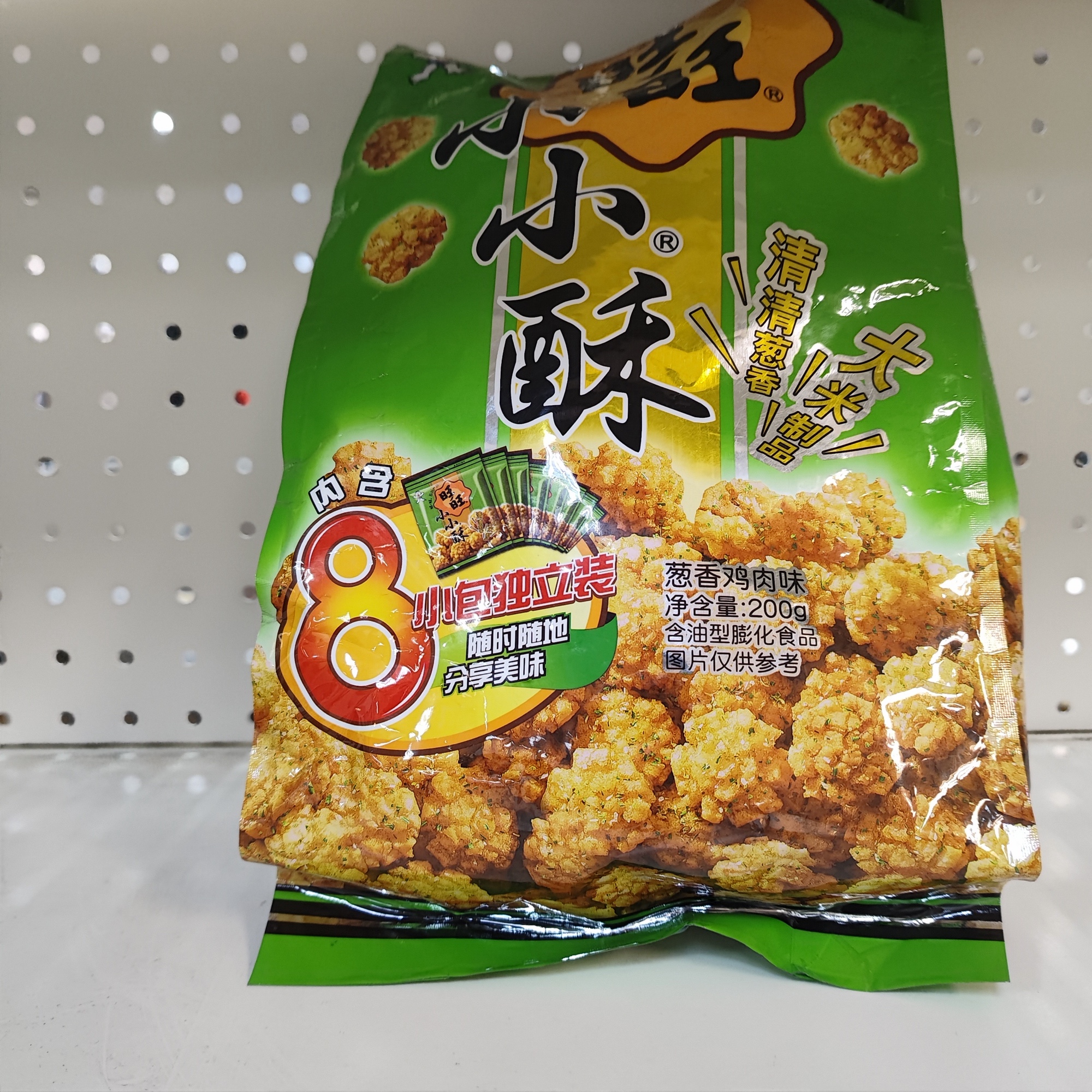} & \includegraphics[width=0.18\linewidth,height=0.70in,keepaspectratio]{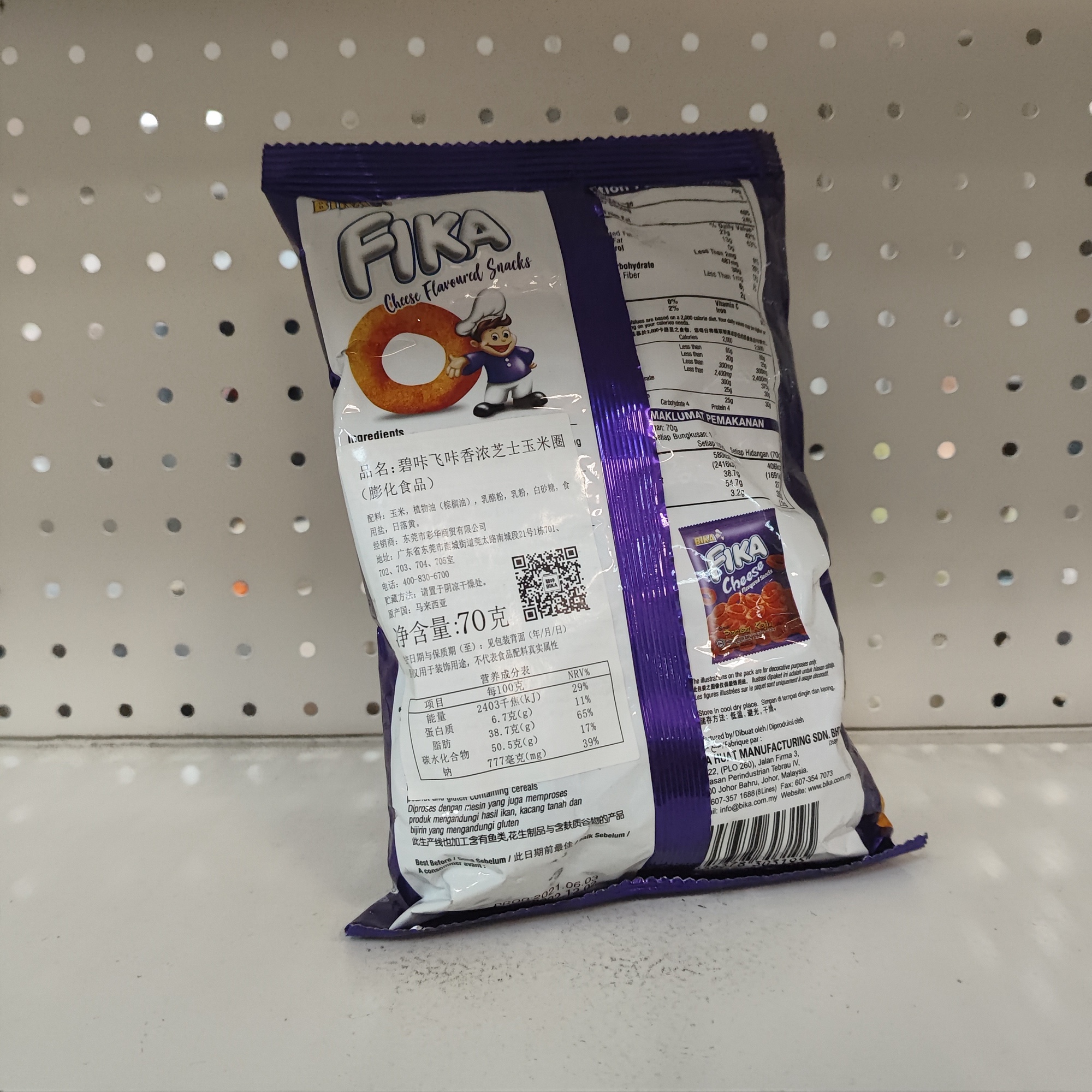} & \includegraphics[width=0.18\linewidth,height=0.70in,keepaspectratio]{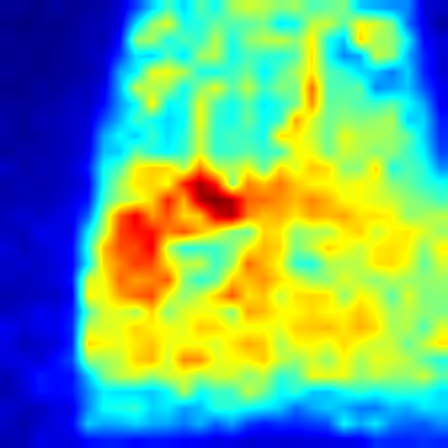} & \includegraphics[width=0.18\linewidth,height=0.70in,keepaspectratio]{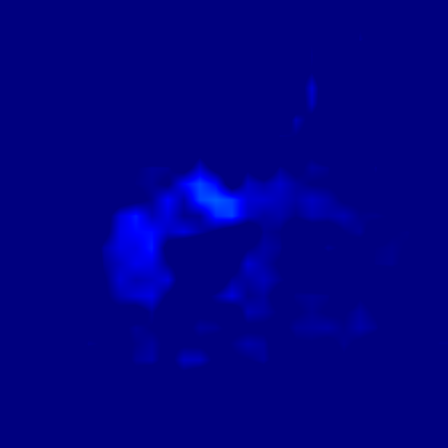} & \includegraphics[width=0.18\linewidth,height=0.70in,keepaspectratio]{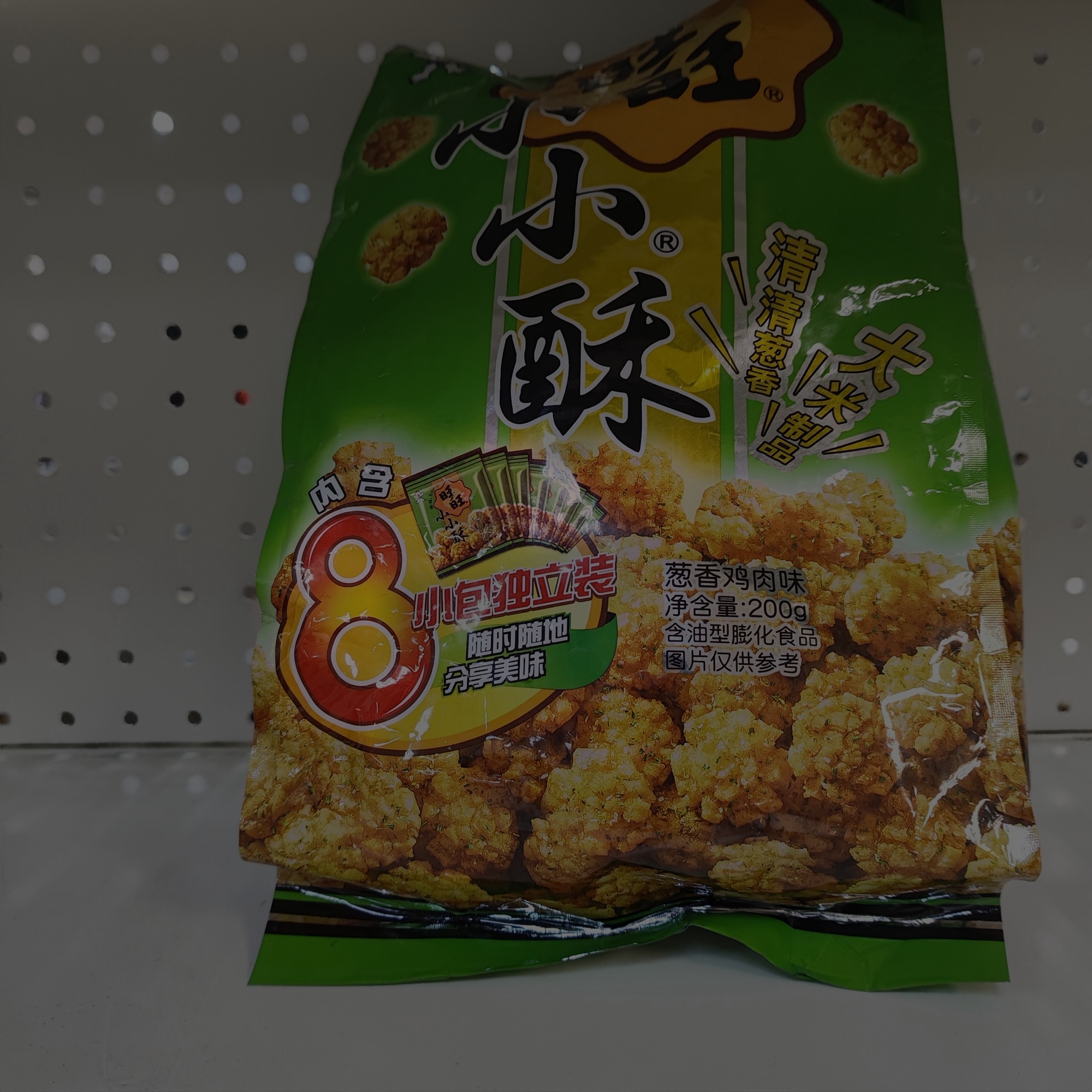} \\
\includegraphics[width=0.18\linewidth,height=0.70in,keepaspectratio]{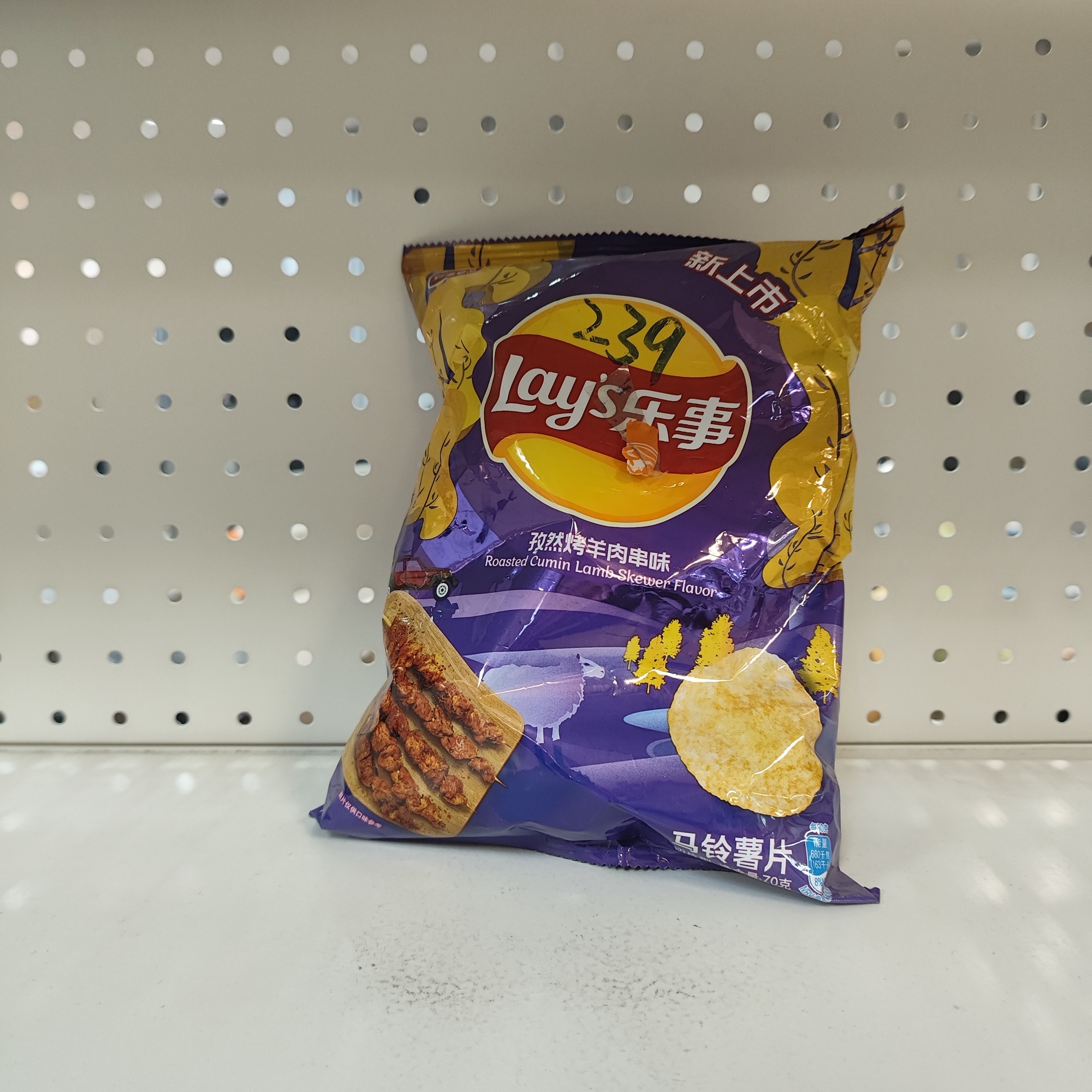} & \includegraphics[width=0.18\linewidth,height=0.70in,keepaspectratio]{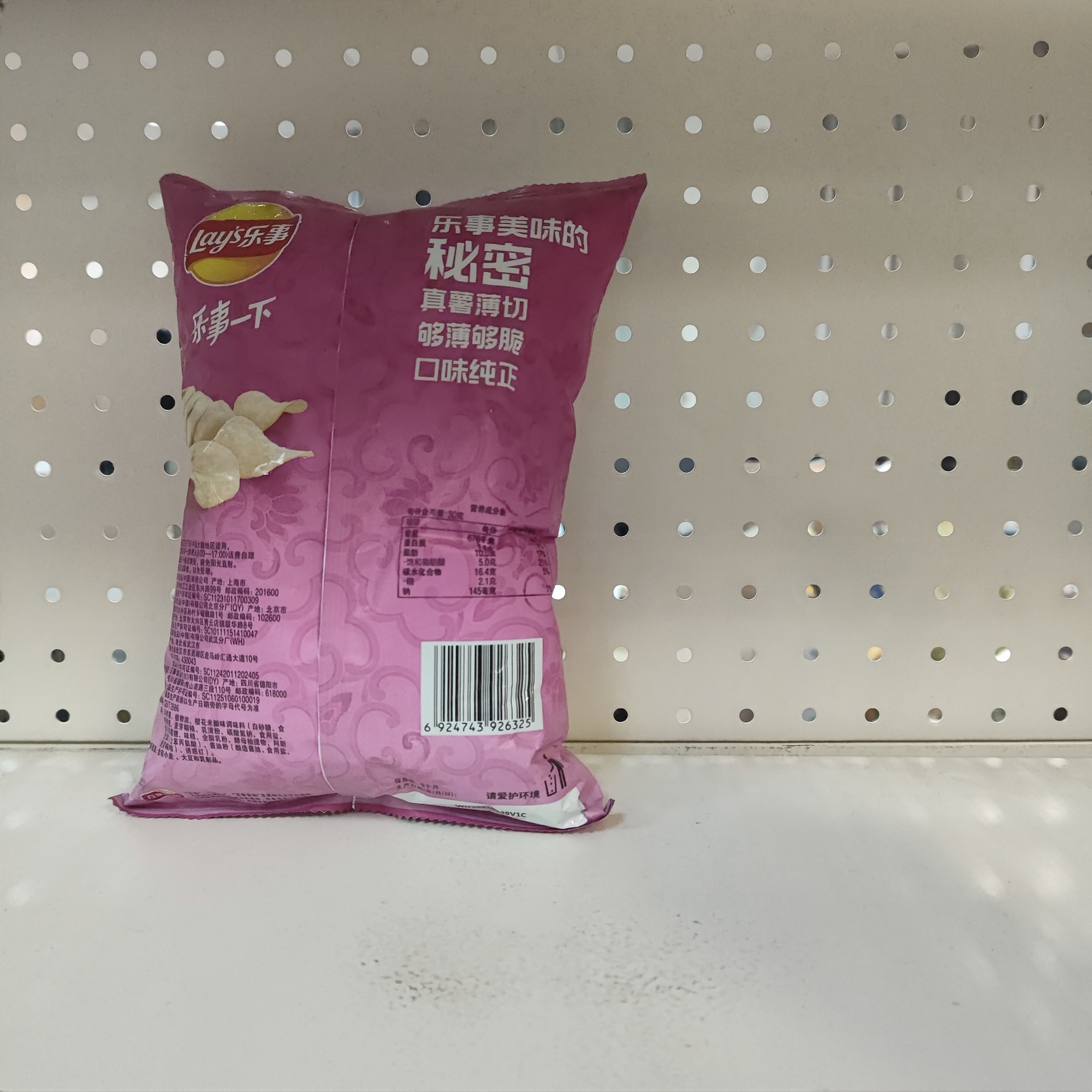} & \includegraphics[width=0.18\linewidth,height=0.70in,keepaspectratio]{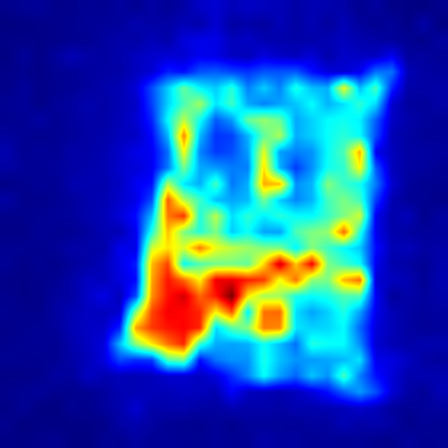} & \includegraphics[width=0.18\linewidth,height=0.70in,keepaspectratio]{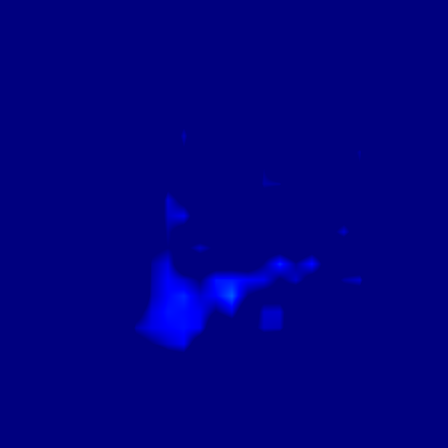} & \includegraphics[width=0.18\linewidth,height=0.70in,keepaspectratio]{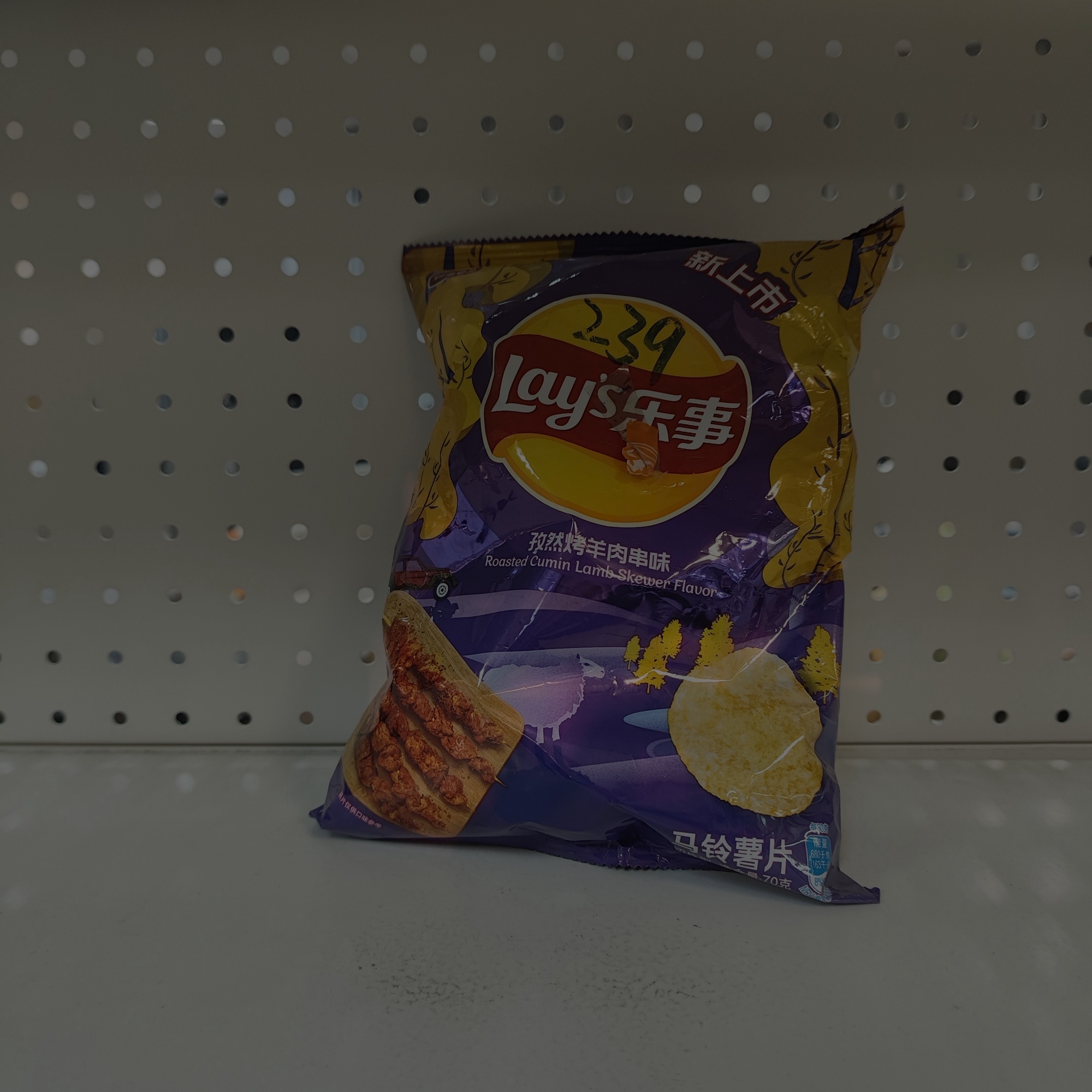} \\
\includegraphics[width=0.18\linewidth,height=0.70in,keepaspectratio]{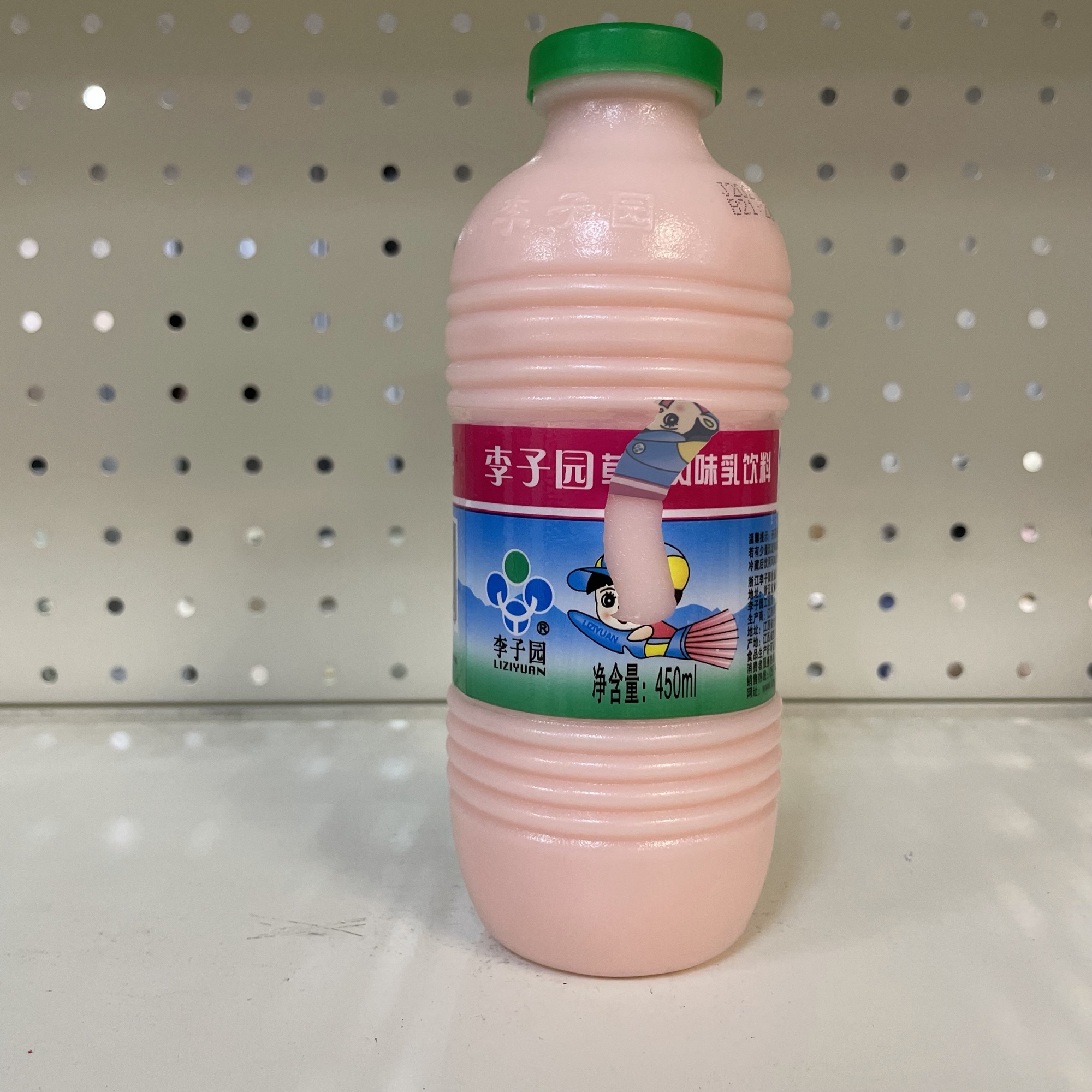} & \includegraphics[width=0.18\linewidth,height=0.70in,keepaspectratio]{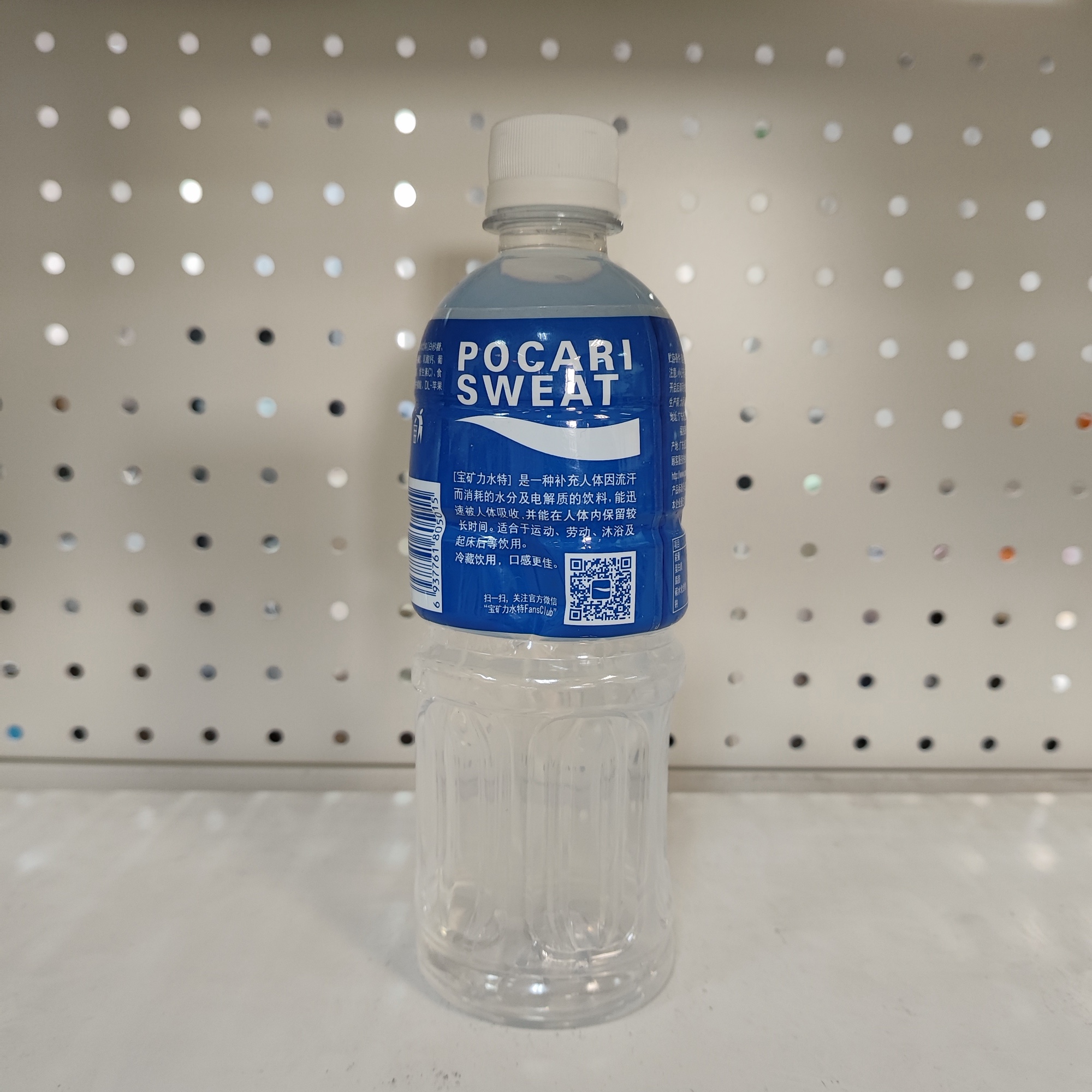} & \includegraphics[width=0.18\linewidth,height=0.70in,keepaspectratio]{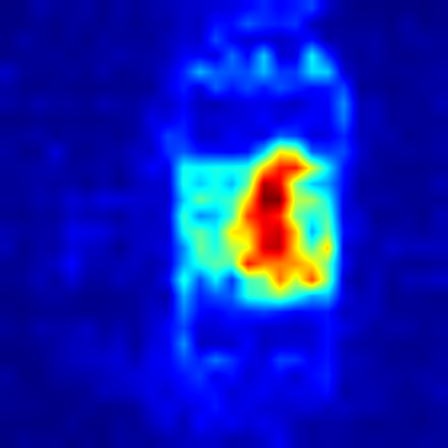} & \includegraphics[width=0.18\linewidth,height=0.70in,keepaspectratio]{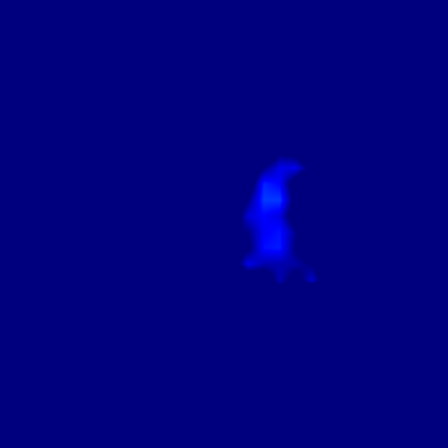} & \includegraphics[width=0.18\linewidth,height=0.70in,keepaspectratio]{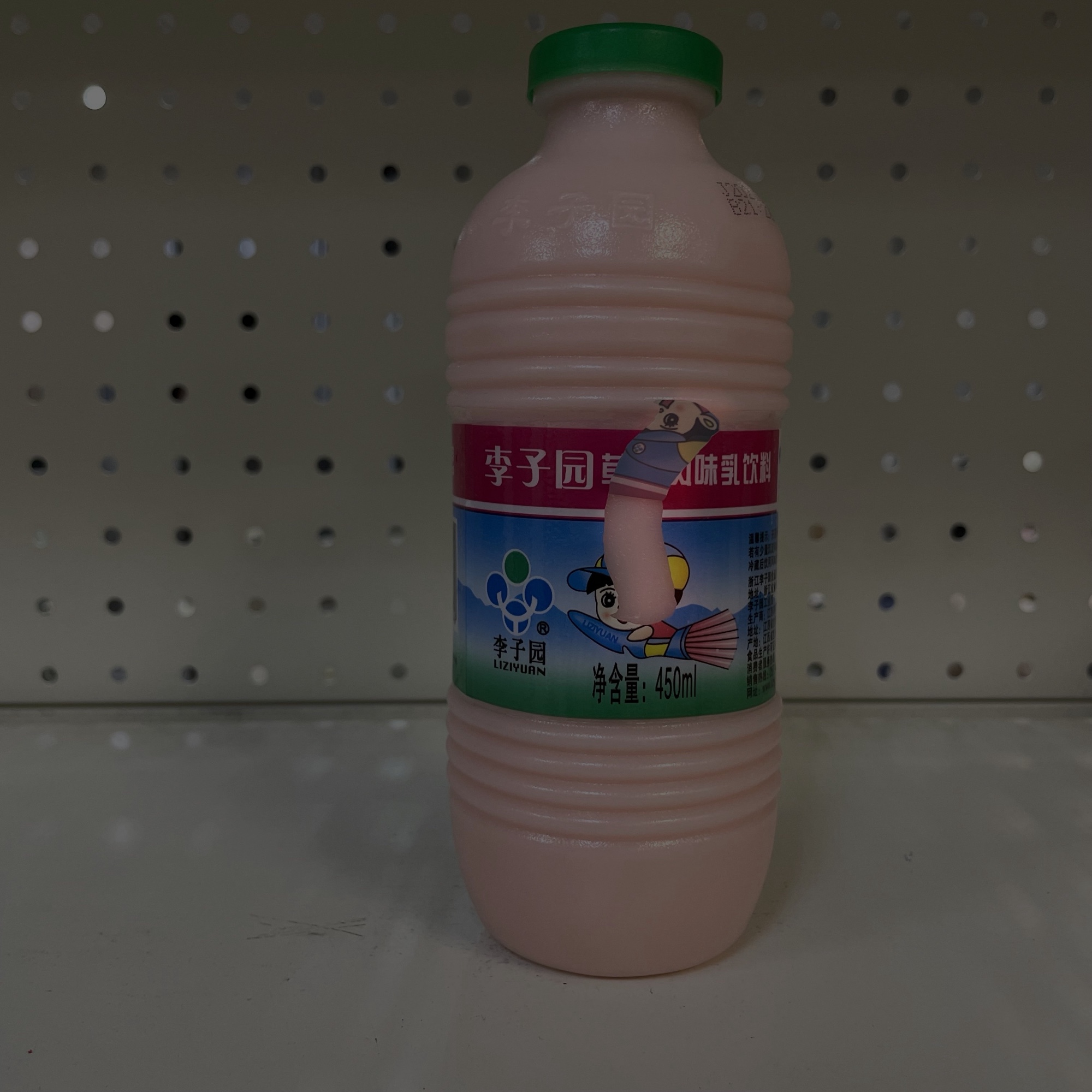} \\
\includegraphics[width=0.18\linewidth,height=0.70in,keepaspectratio]{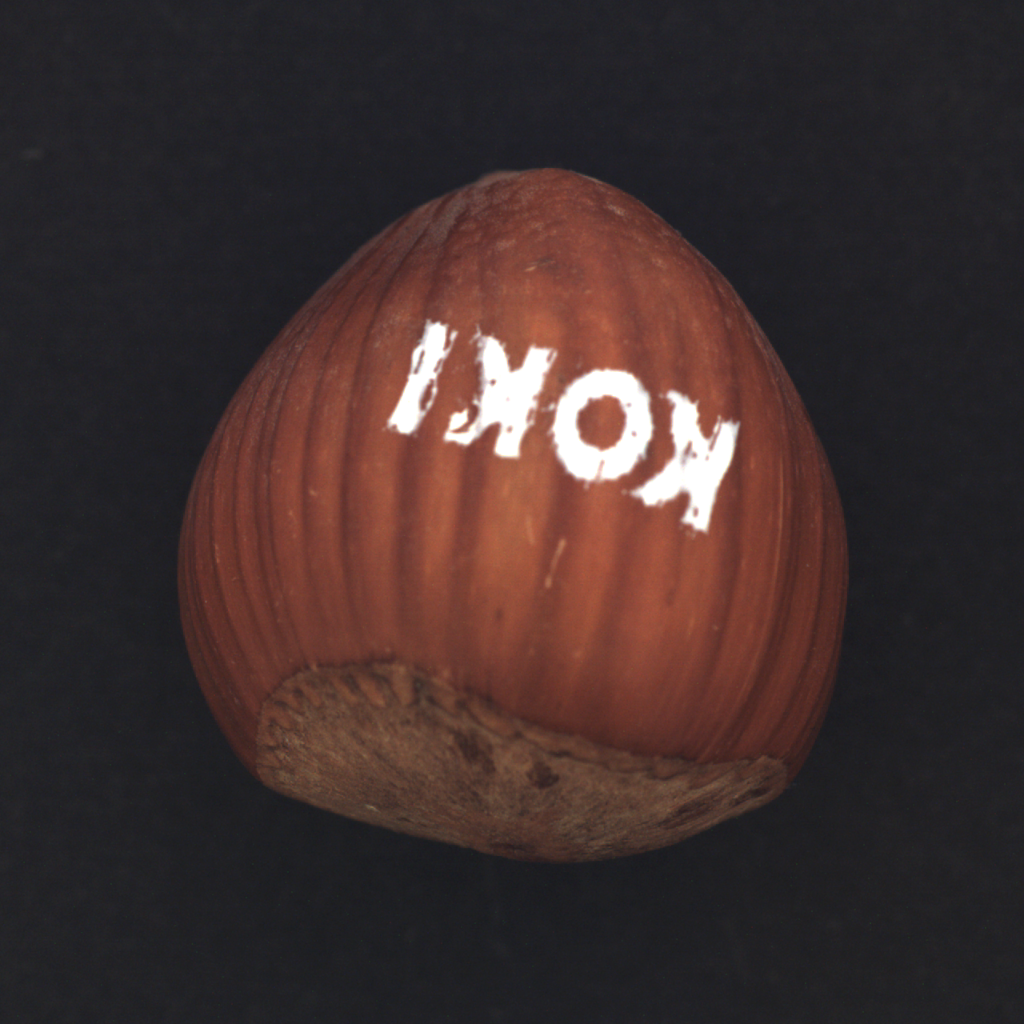} & \includegraphics[width=0.18\linewidth,height=0.70in,keepaspectratio]{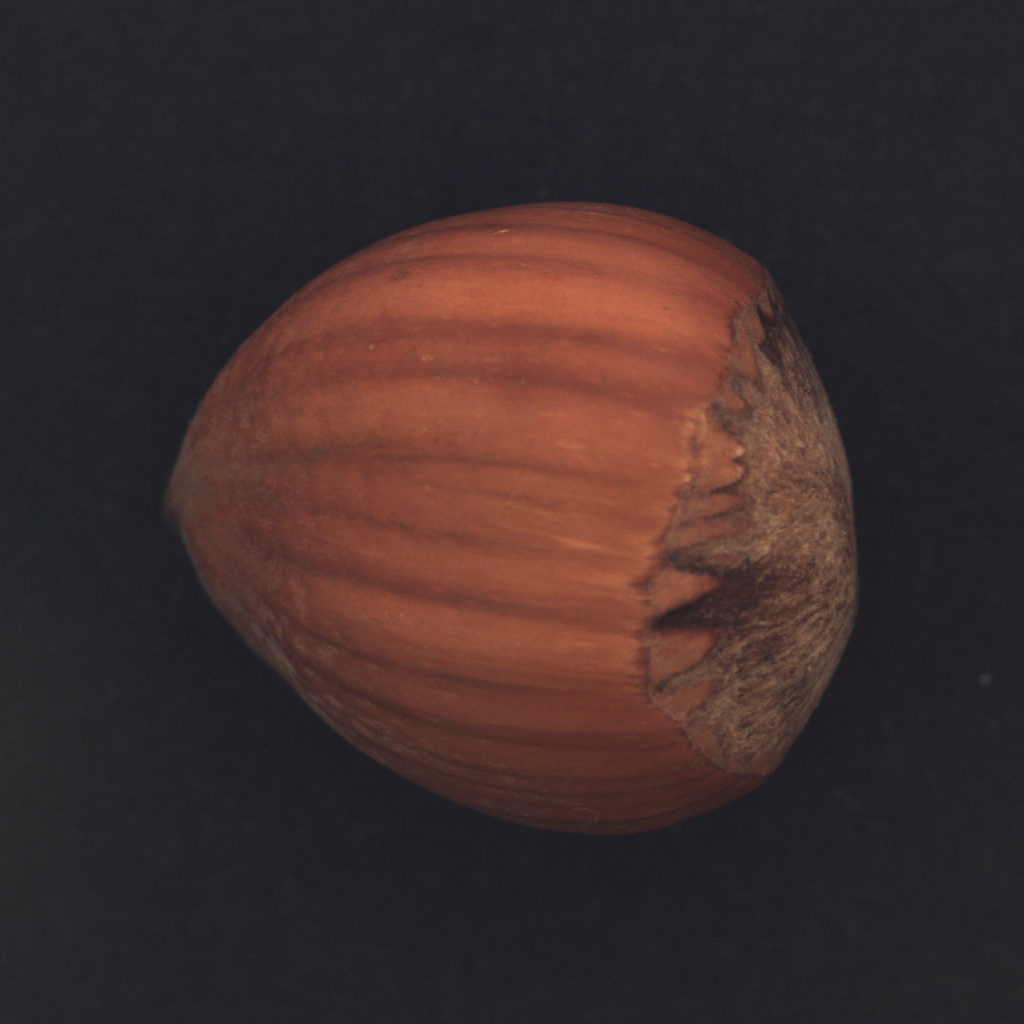} & \includegraphics[width=0.18\linewidth,height=0.70in,keepaspectratio]{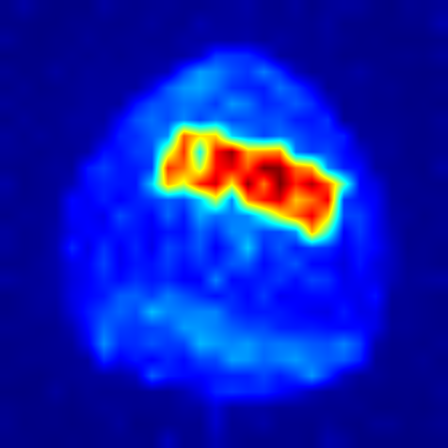} & \includegraphics[width=0.18\linewidth,height=0.70in,keepaspectratio]{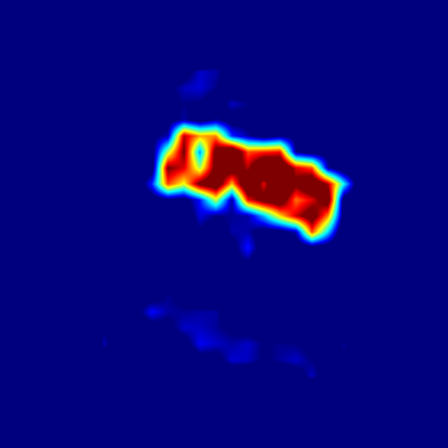} & \includegraphics[width=0.18\linewidth,height=0.70in,keepaspectratio]{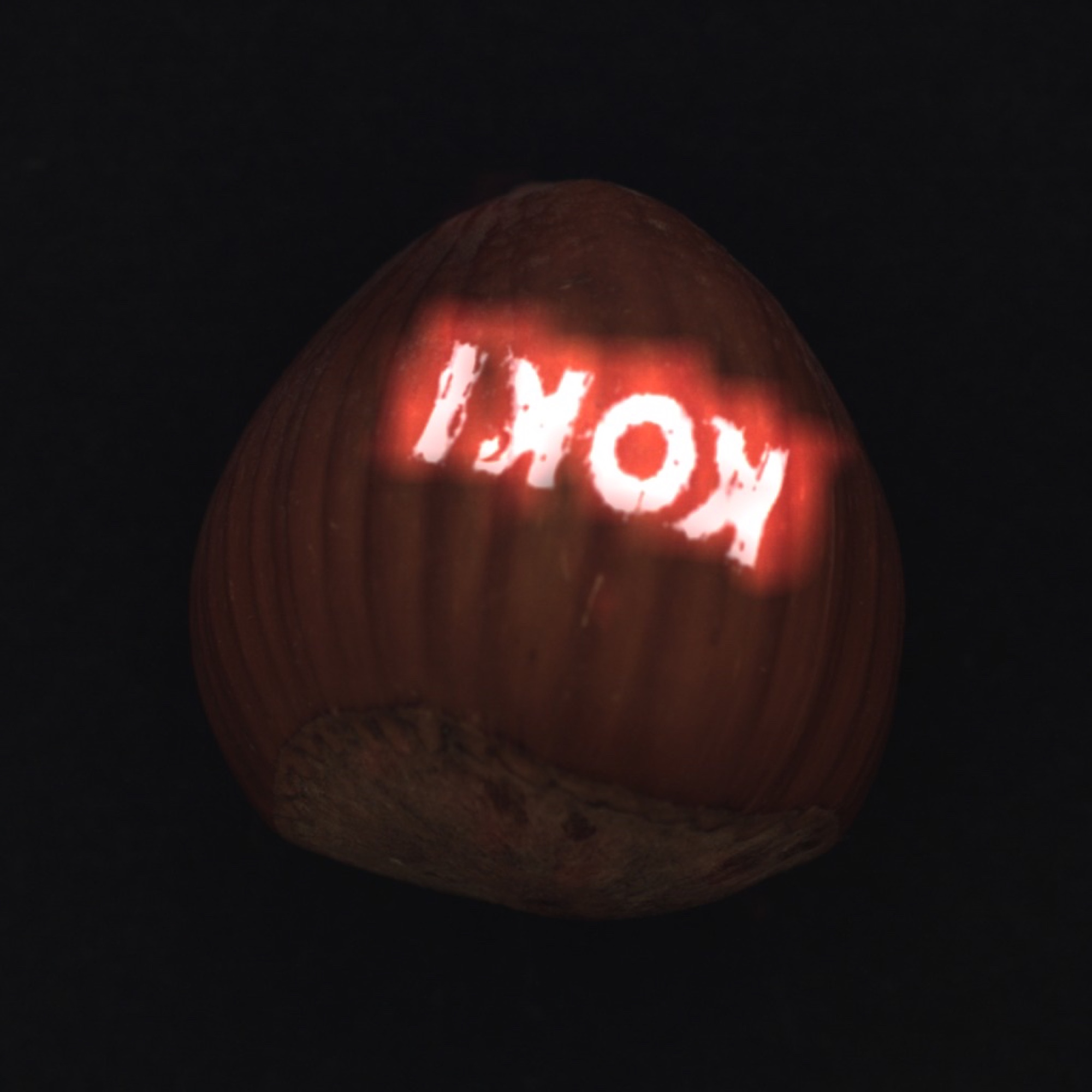} \\
\includegraphics[width=0.18\linewidth,height=0.70in,keepaspectratio]{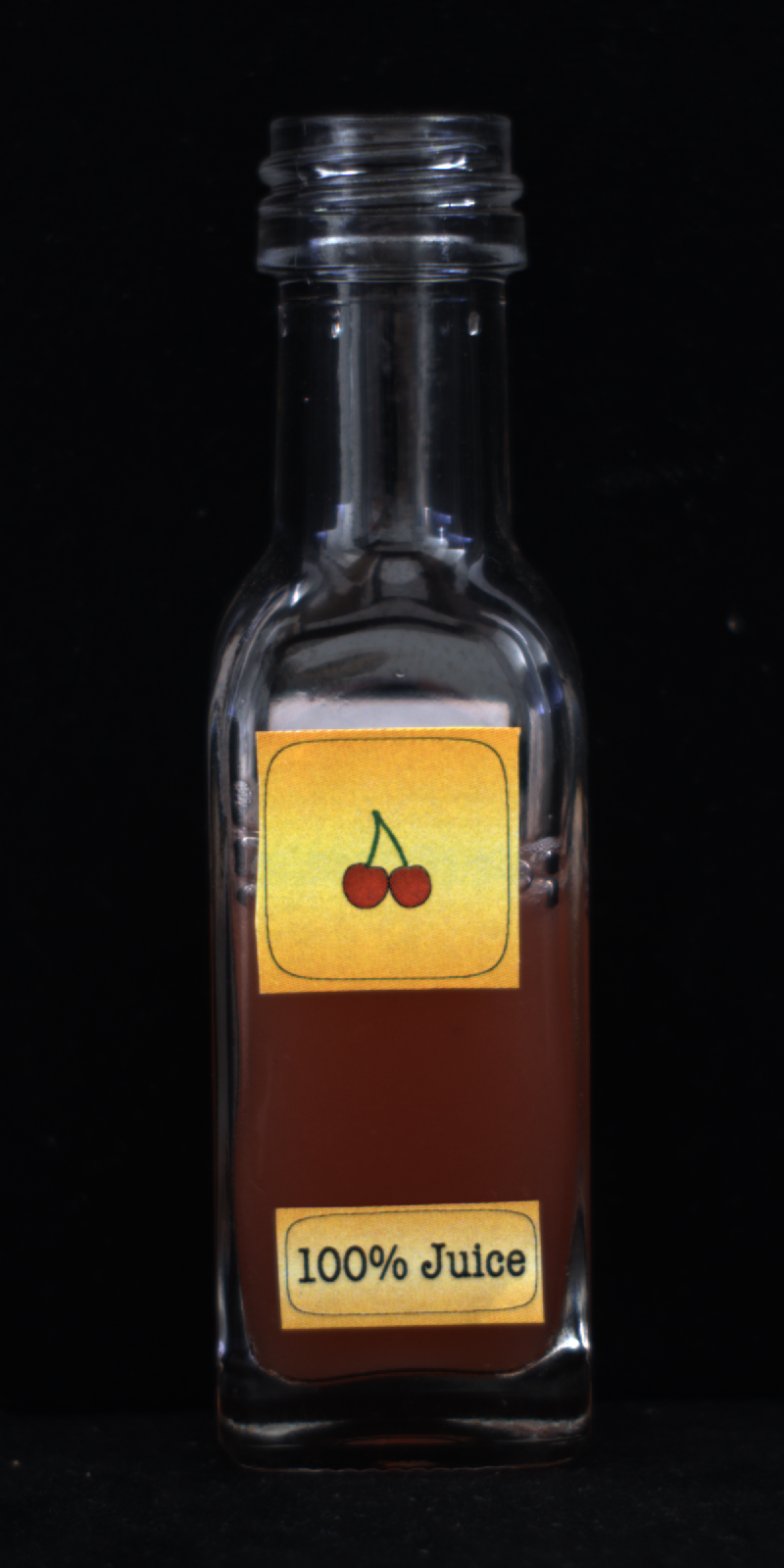} & \includegraphics[width=0.18\linewidth,height=0.70in,keepaspectratio]{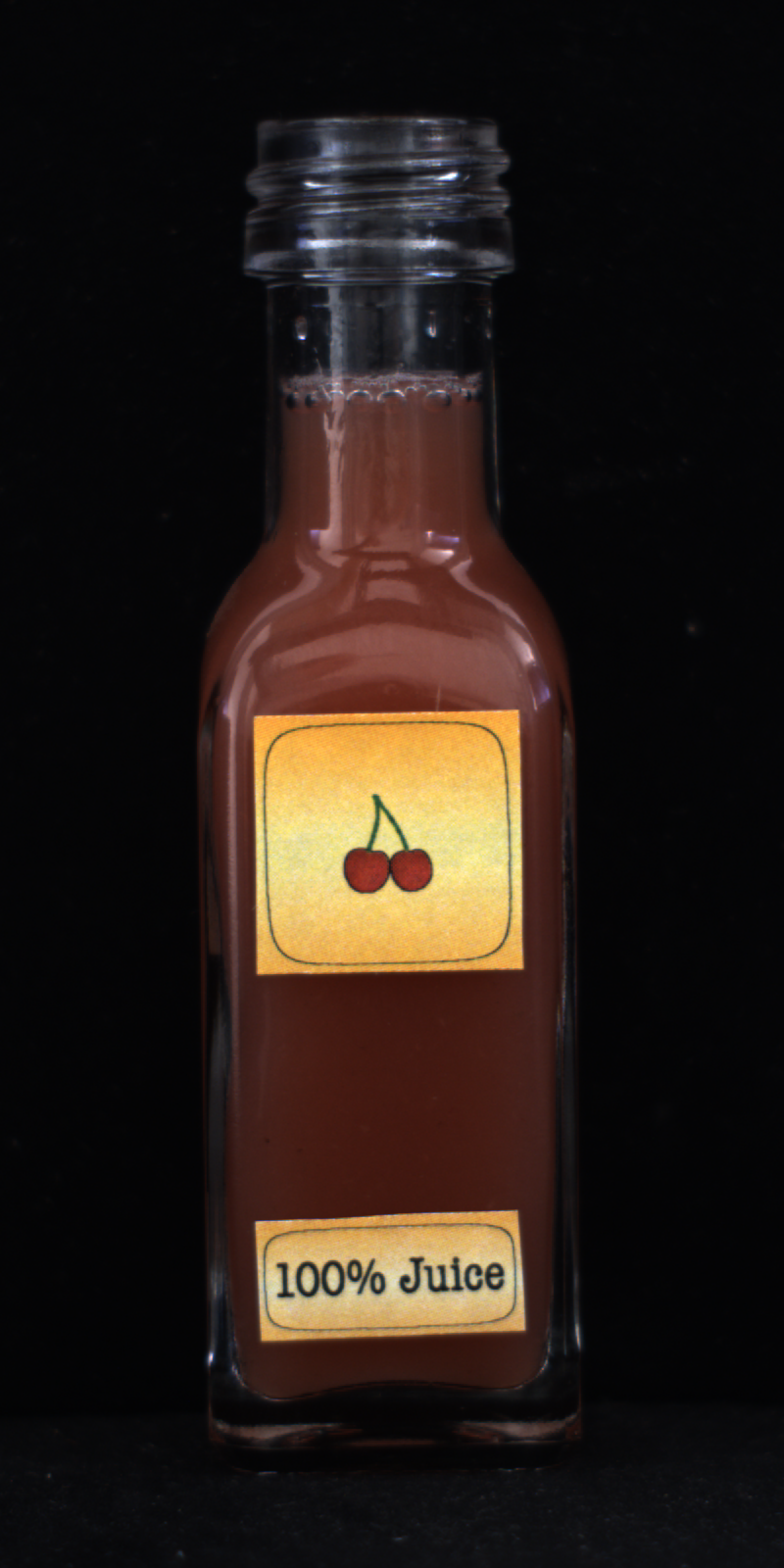} & \includegraphics[width=0.18\linewidth,height=0.70in,keepaspectratio]{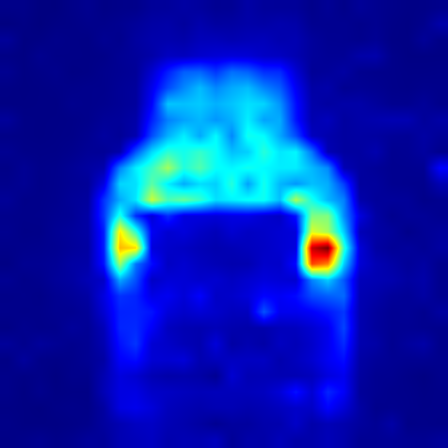} & \includegraphics[width=0.18\linewidth,height=0.70in,keepaspectratio]{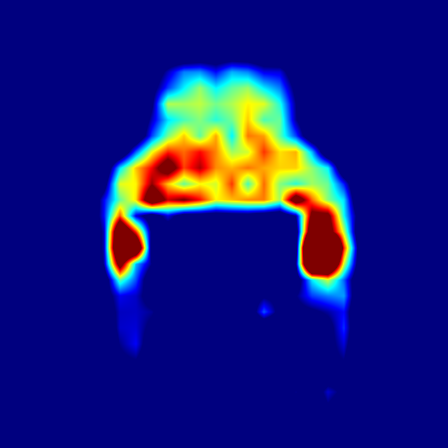} & \includegraphics[width=0.18\linewidth,height=0.70in,keepaspectratio]{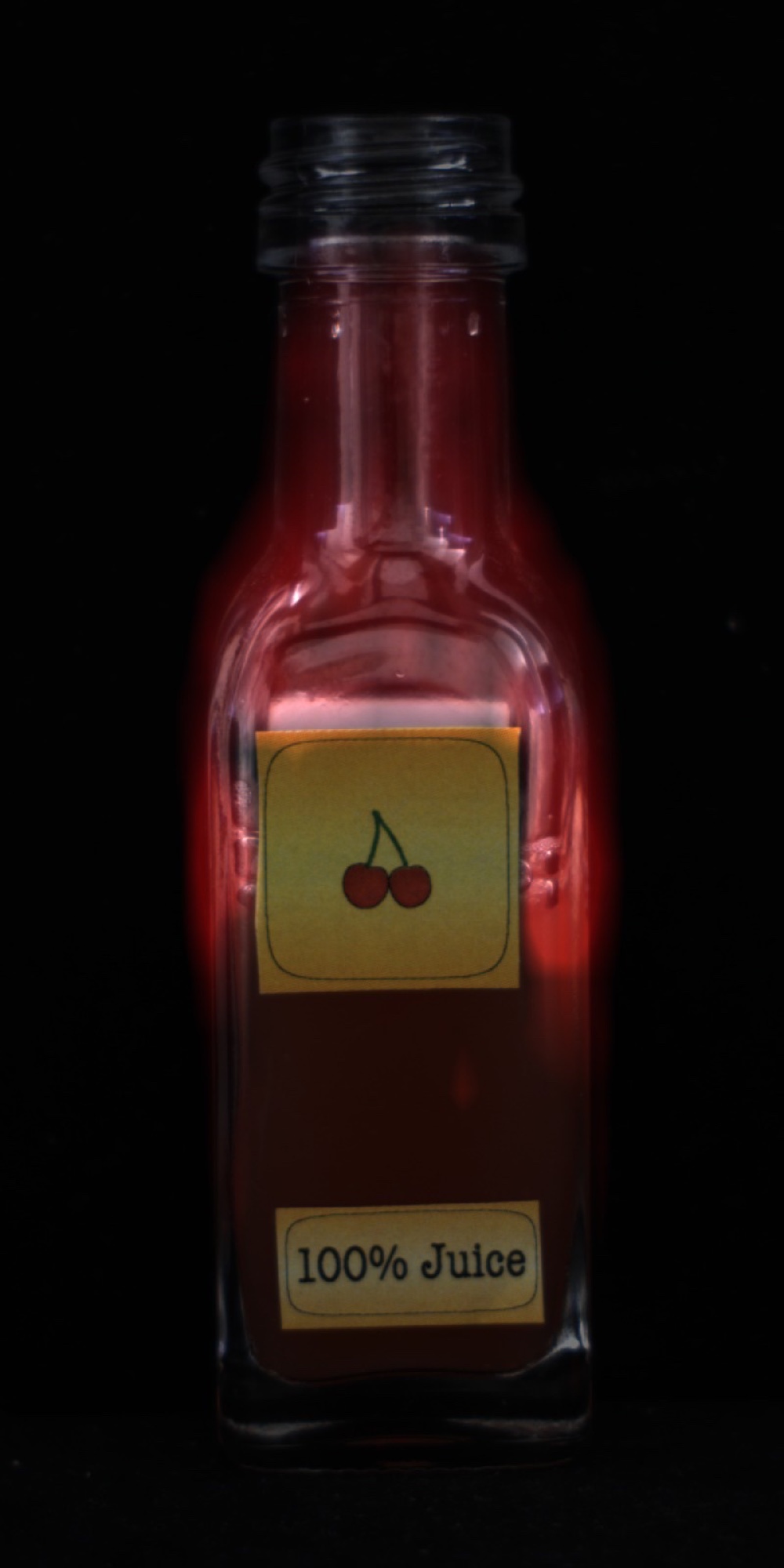} \\
\includegraphics[width=0.18\linewidth,height=0.70in,keepaspectratio]{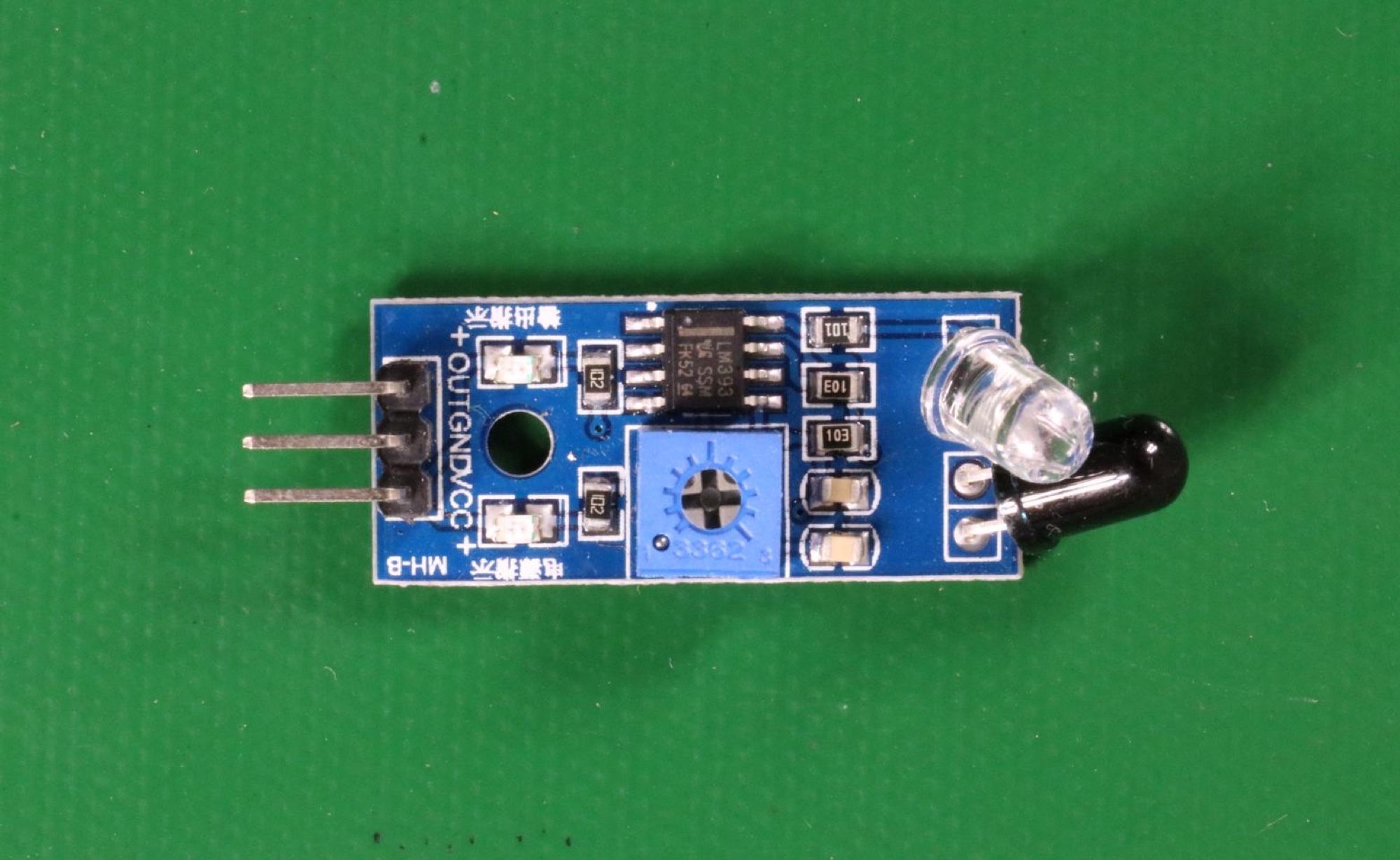} & \includegraphics[width=0.18\linewidth,height=0.70in,keepaspectratio]{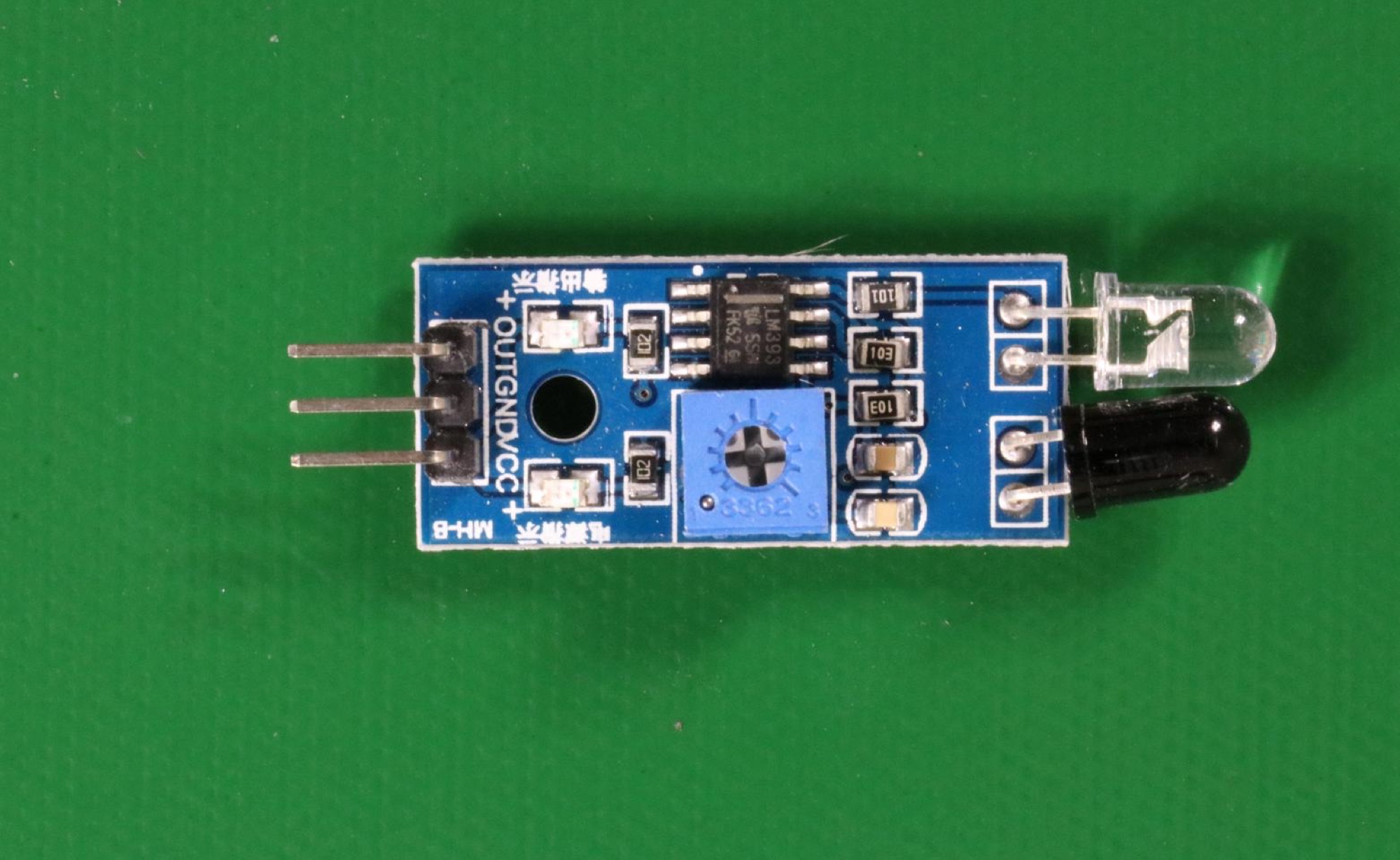} & \includegraphics[width=0.18\linewidth,height=0.70in,keepaspectratio]{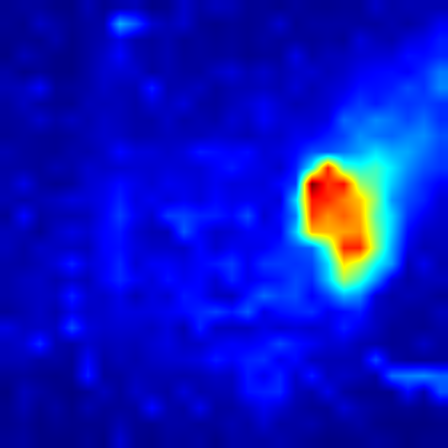} & \includegraphics[width=0.18\linewidth,height=0.70in,keepaspectratio]{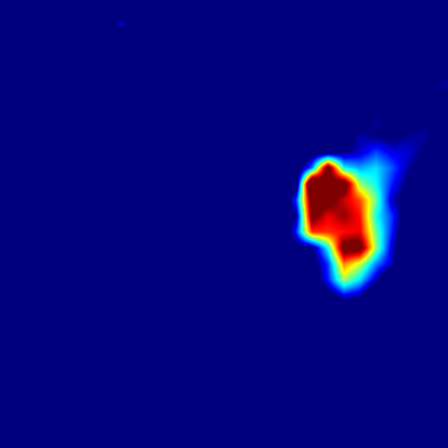} & \includegraphics[width=0.18\linewidth,height=0.70in,keepaspectratio]{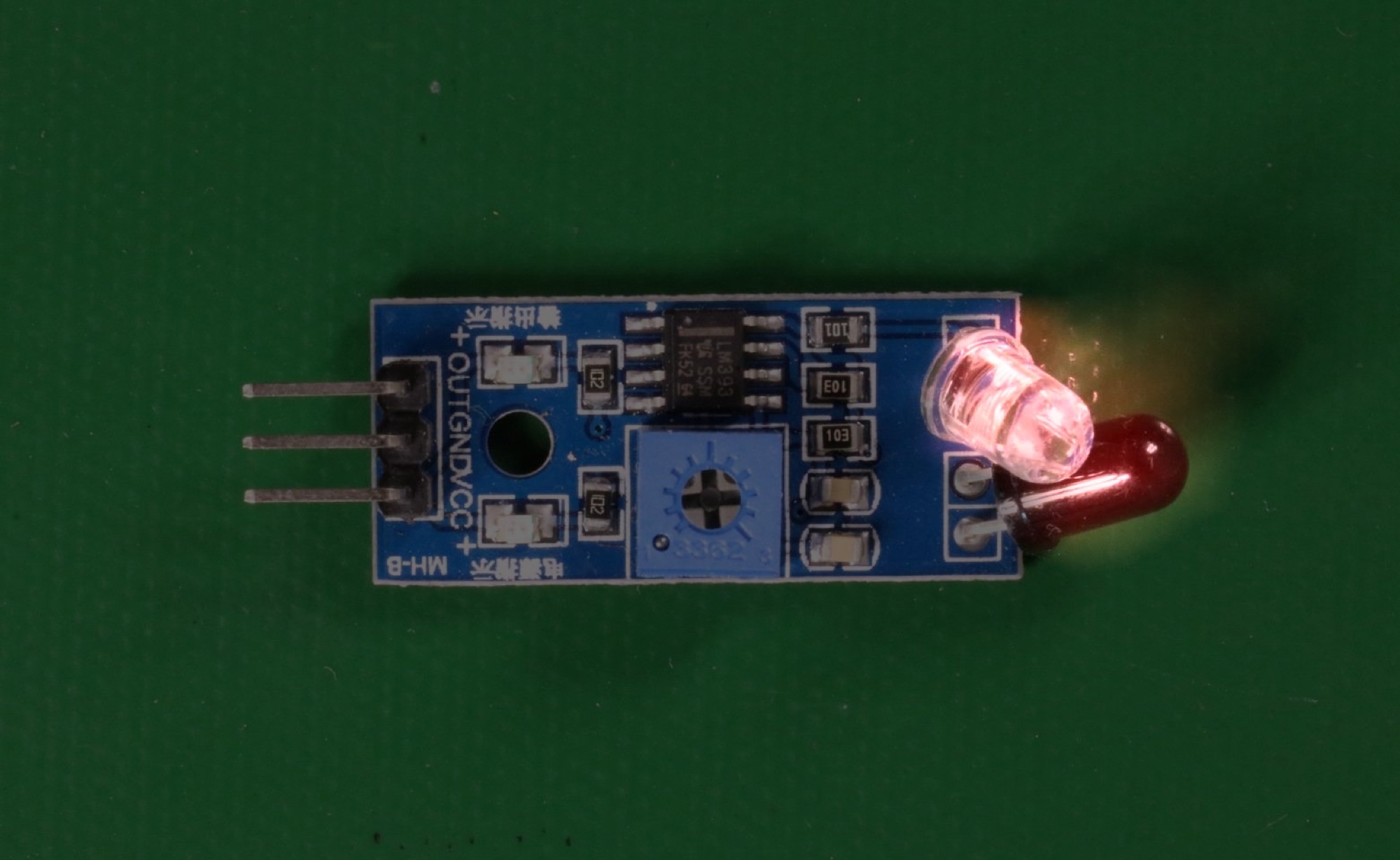} \\
\bottomrule
\end{tabular*}
\caption{Effect of Noise-Robust Calibration on defect-bearing queries. Raw
residual maps include query--reference differences beyond the defect; the
calibrated maps and final overlays concentrate the evidence supplied to expert
readers.}
\label{fig:calibration_focus_grid}
\end{figure}

\begin{figure}[H]
\centering
\small
\setlength{\tabcolsep}{0pt}
\renewcommand{\arraystretch}{1.12}
\begin{tabular*}{\linewidth}{@{\extracolsep{\fill}}ccccc@{}}
\toprule
Query & Normal reference & Raw map & Calibrated map & Overlay \\
\midrule
\includegraphics[width=0.18\linewidth,height=0.70in,keepaspectratio]{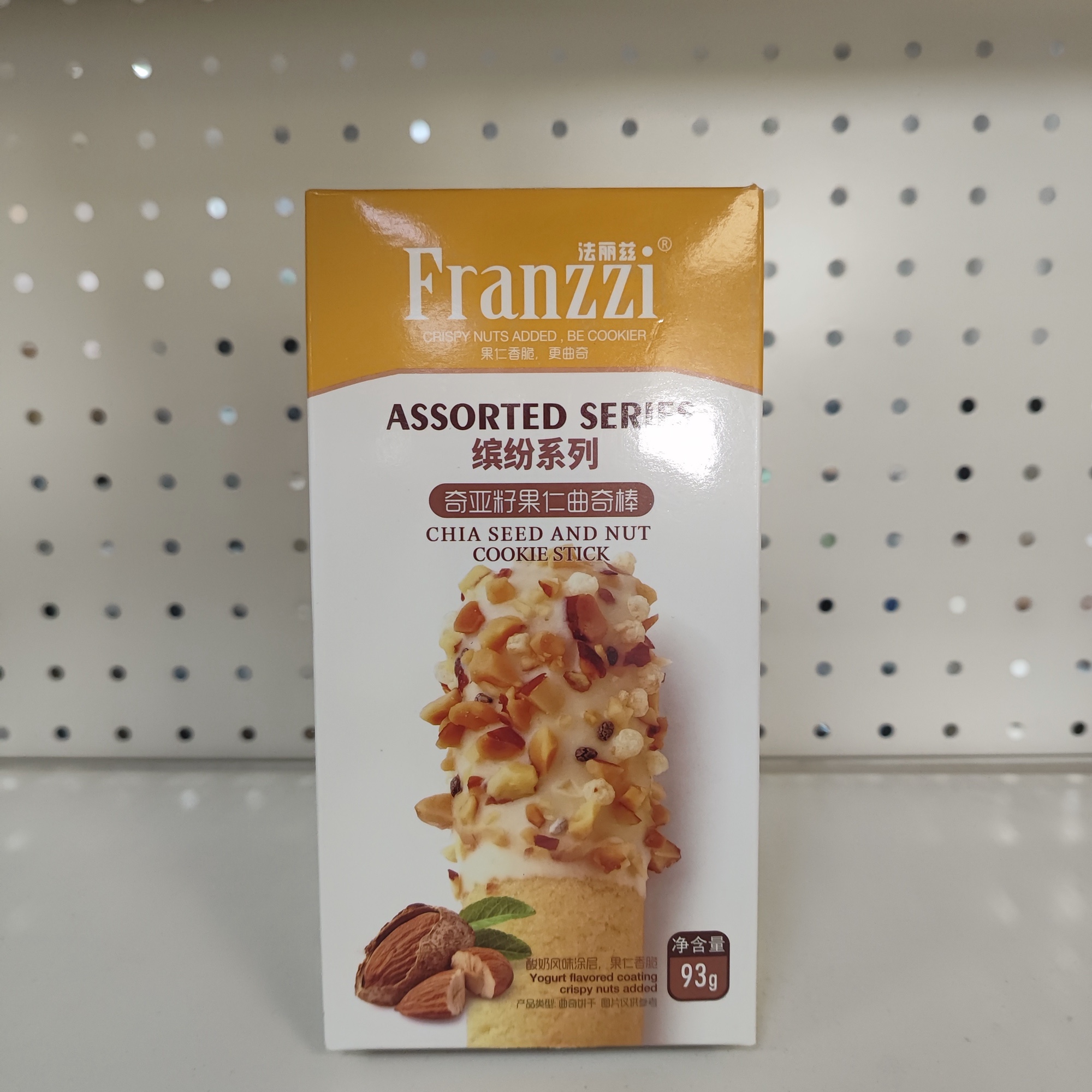} & \includegraphics[width=0.18\linewidth,height=0.70in,keepaspectratio]{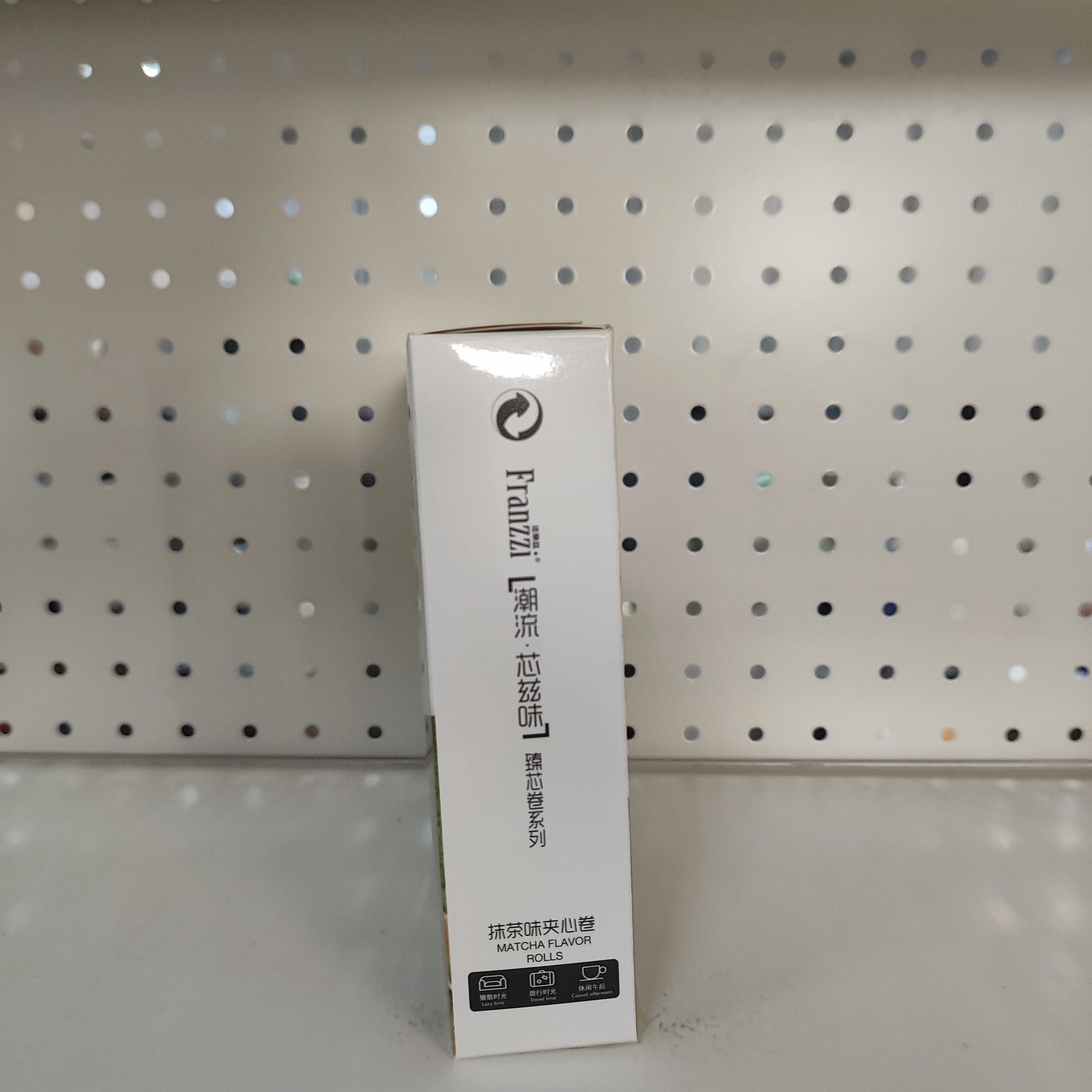} & \includegraphics[width=0.18\linewidth,height=0.70in,keepaspectratio]{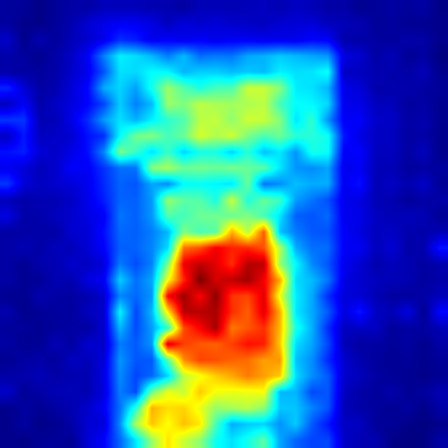} & \includegraphics[width=0.18\linewidth,height=0.70in,keepaspectratio]{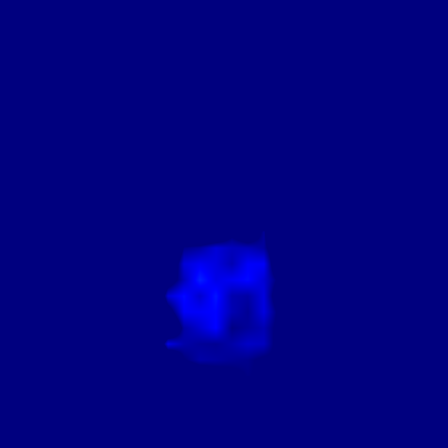} & \includegraphics[width=0.18\linewidth,height=0.70in,keepaspectratio]{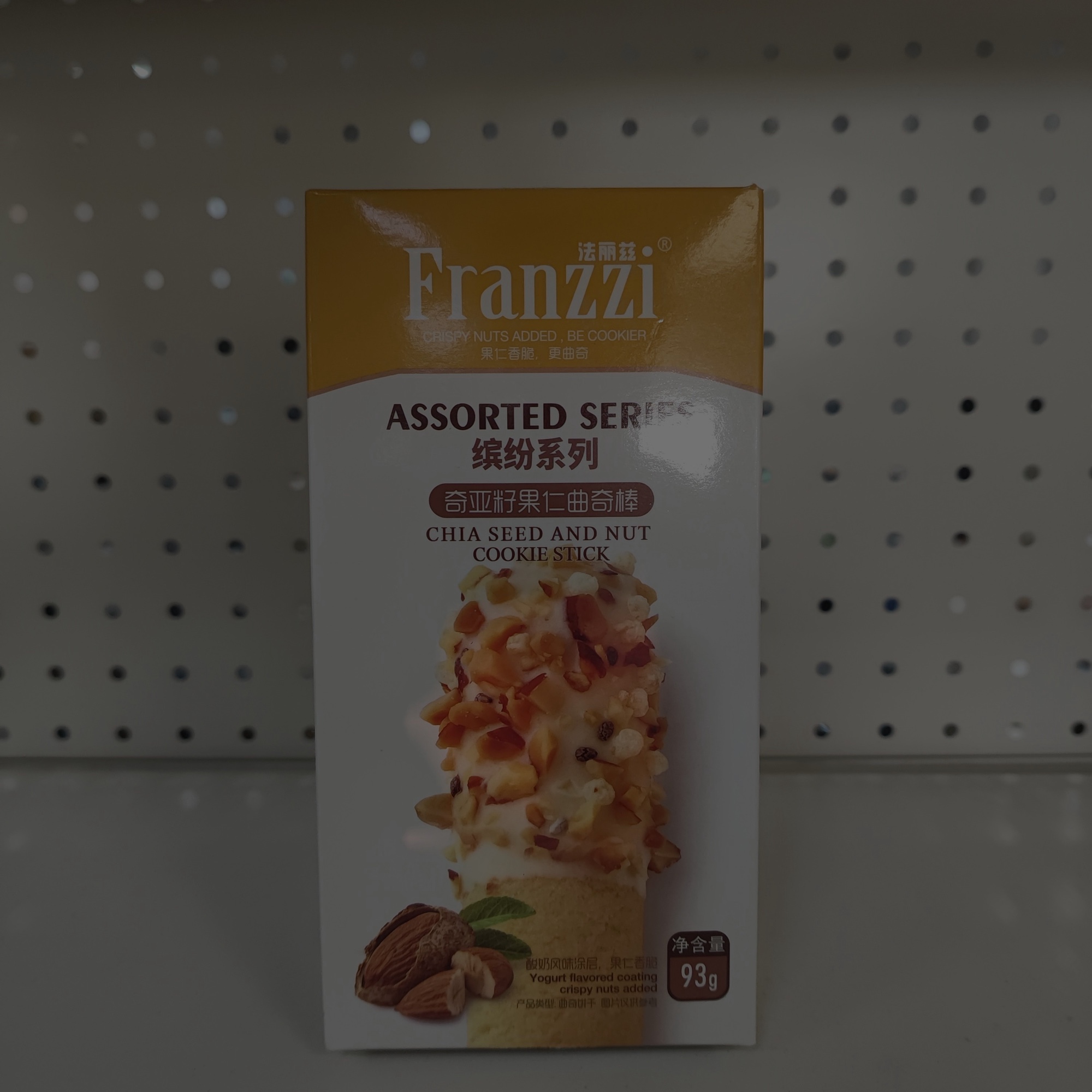} \\
\includegraphics[width=0.18\linewidth,height=0.70in,keepaspectratio]{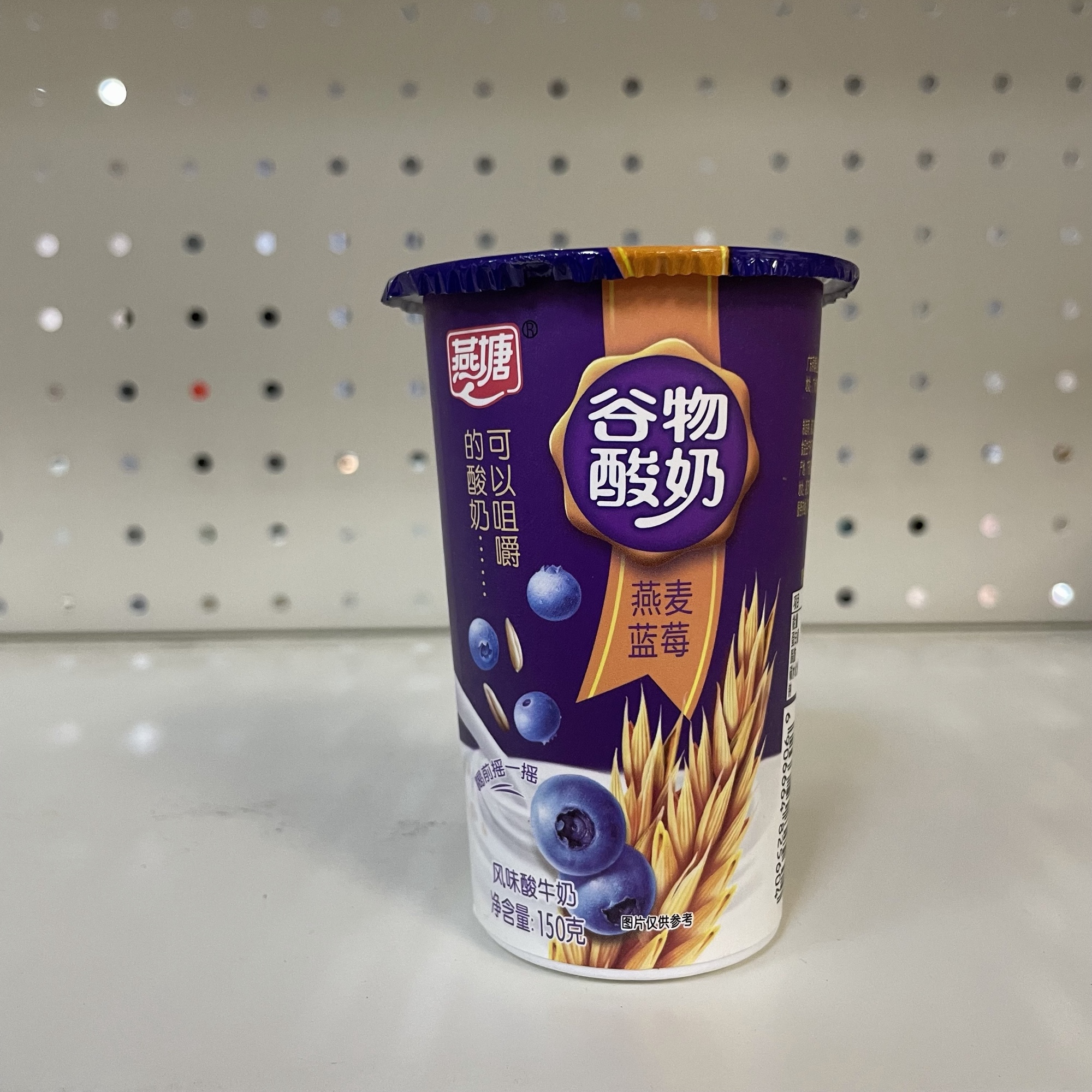} & \includegraphics[width=0.18\linewidth,height=0.70in,keepaspectratio]{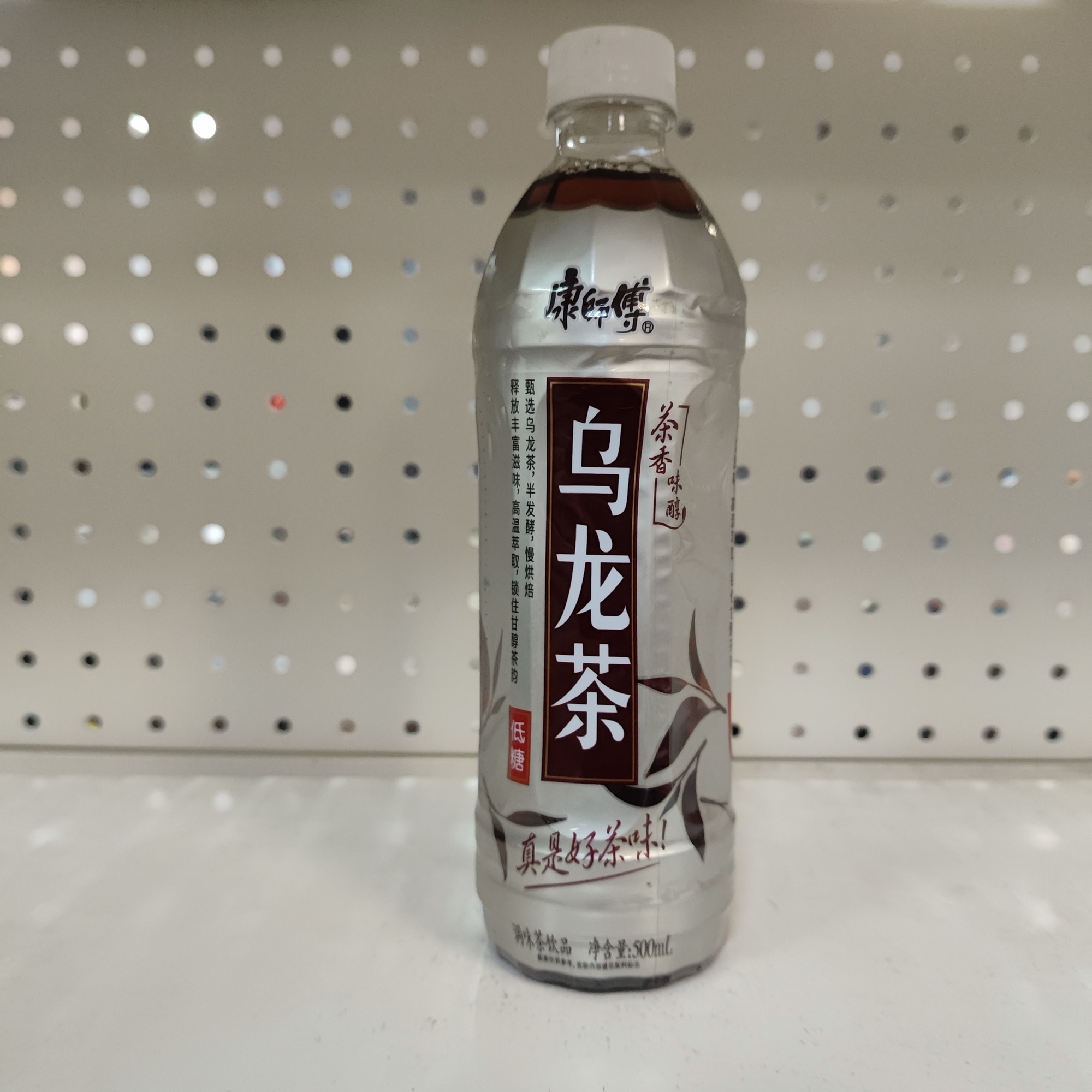} & \includegraphics[width=0.18\linewidth,height=0.70in,keepaspectratio]{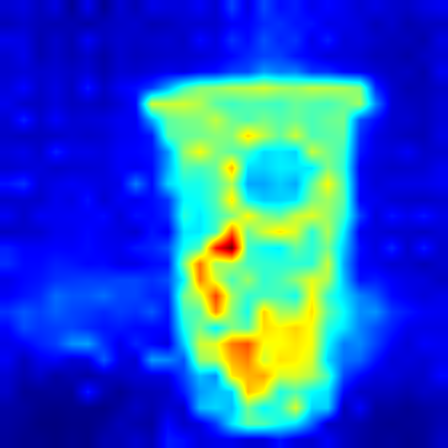} & \includegraphics[width=0.18\linewidth,height=0.70in,keepaspectratio]{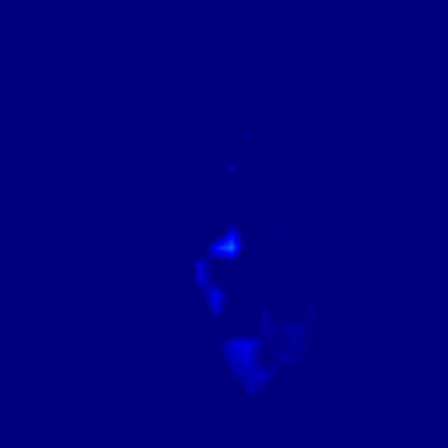} & \includegraphics[width=0.18\linewidth,height=0.70in,keepaspectratio]{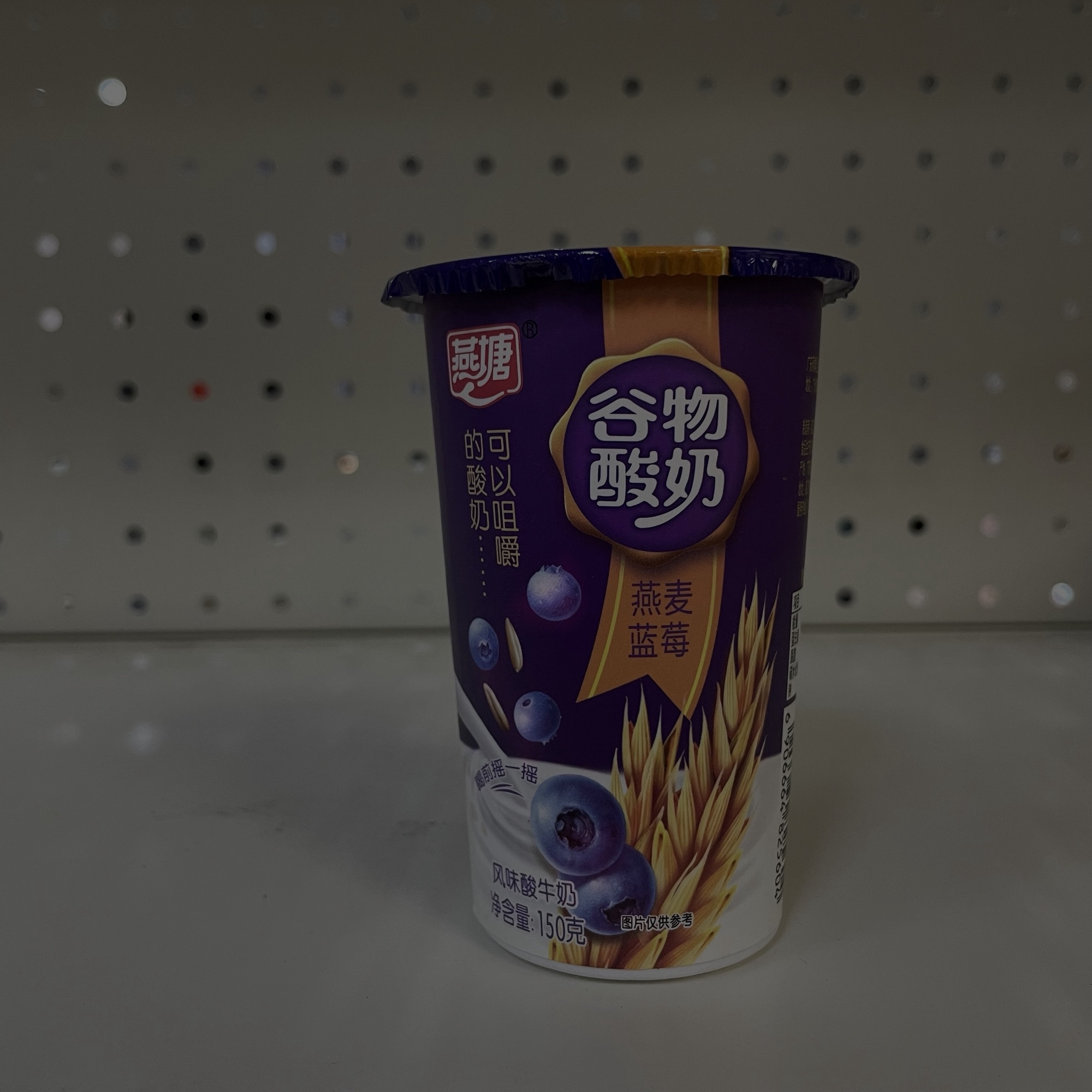} \\
\includegraphics[width=0.18\linewidth,height=0.70in,keepaspectratio]{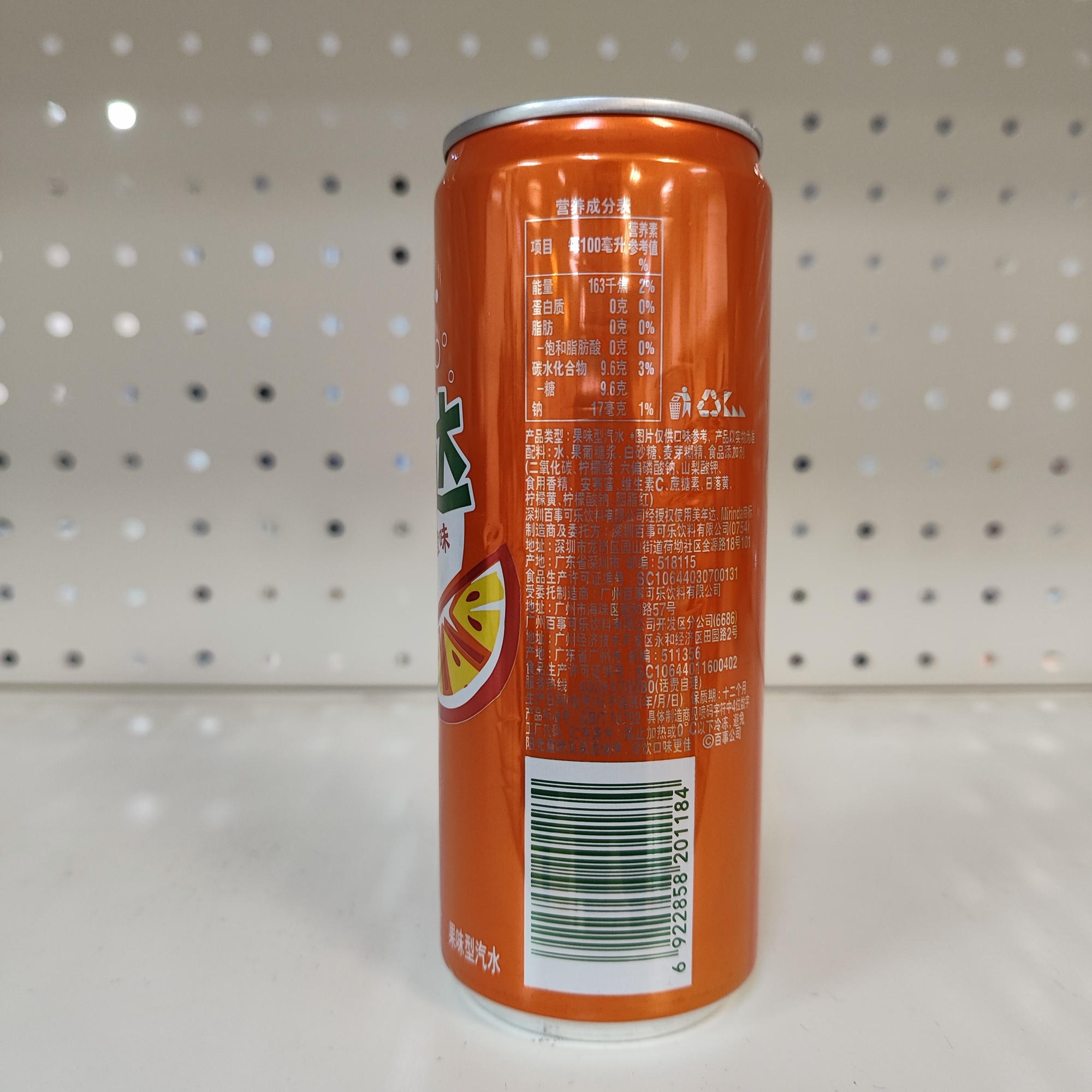} & \includegraphics[width=0.18\linewidth,height=0.70in,keepaspectratio]{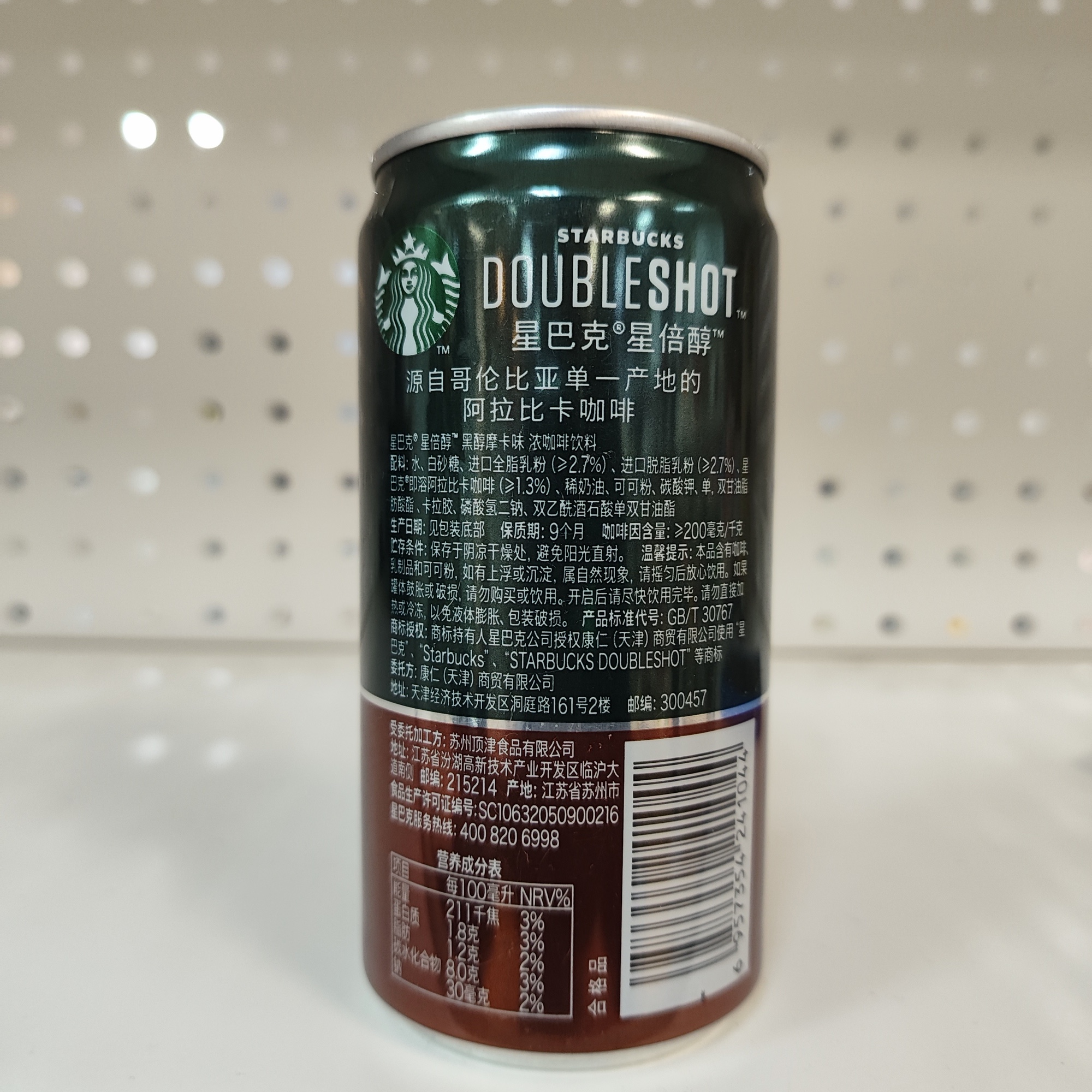} & \includegraphics[width=0.18\linewidth,height=0.70in,keepaspectratio]{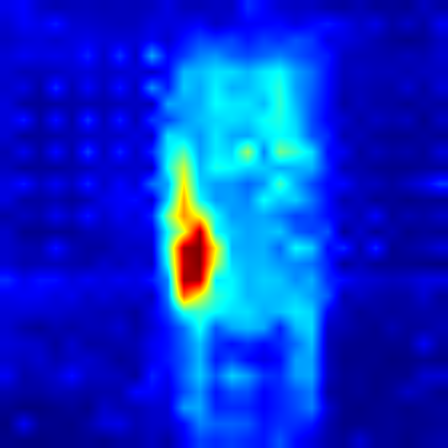} & \includegraphics[width=0.18\linewidth,height=0.70in,keepaspectratio]{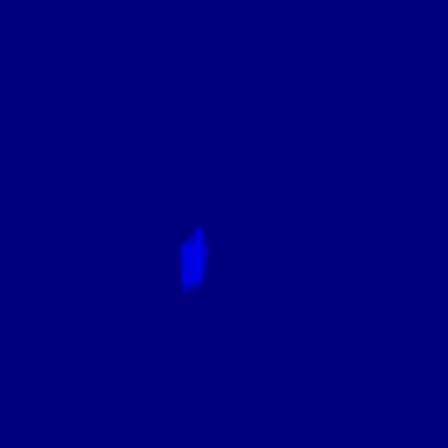} & \includegraphics[width=0.18\linewidth,height=0.70in,keepaspectratio]{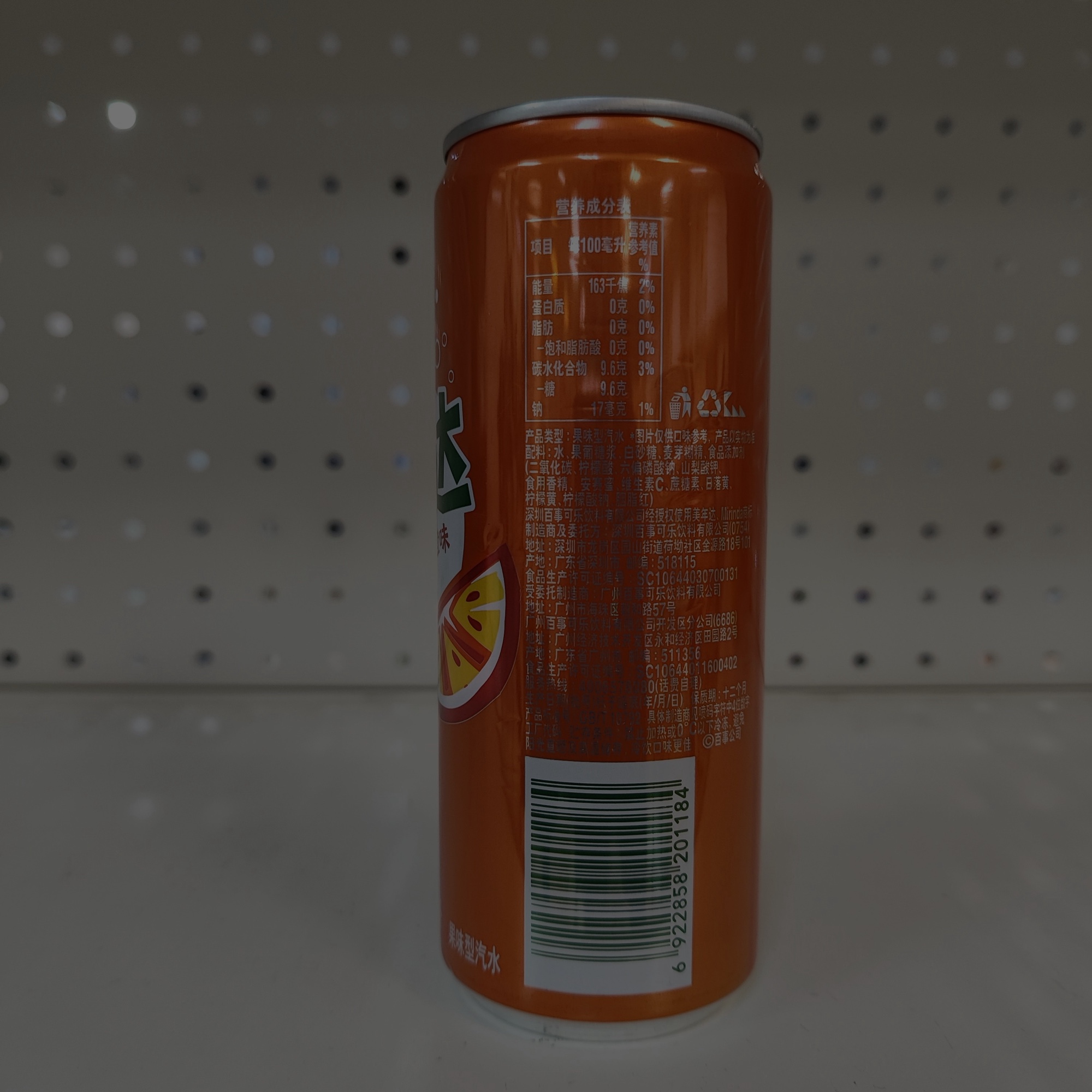} \\
\includegraphics[width=0.18\linewidth,height=0.70in,keepaspectratio]{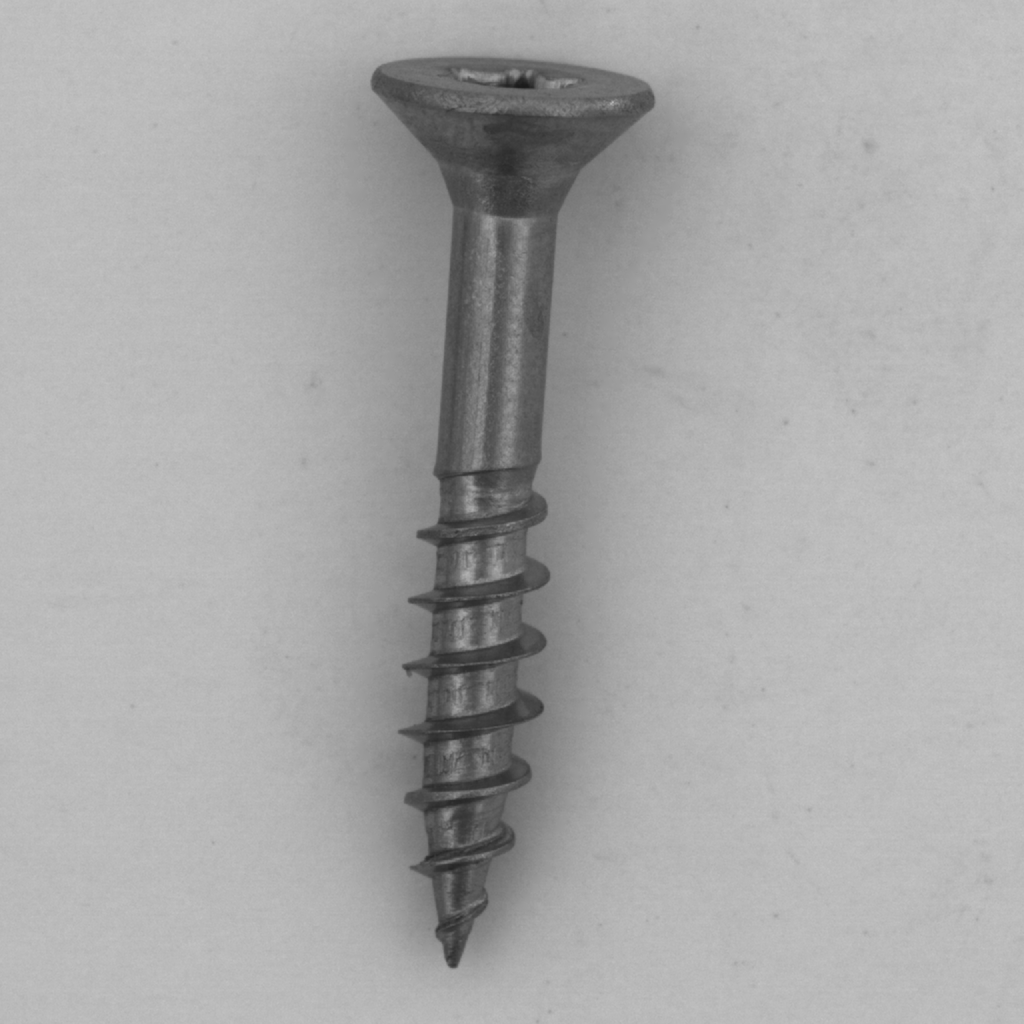} & \includegraphics[width=0.18\linewidth,height=0.70in,keepaspectratio]{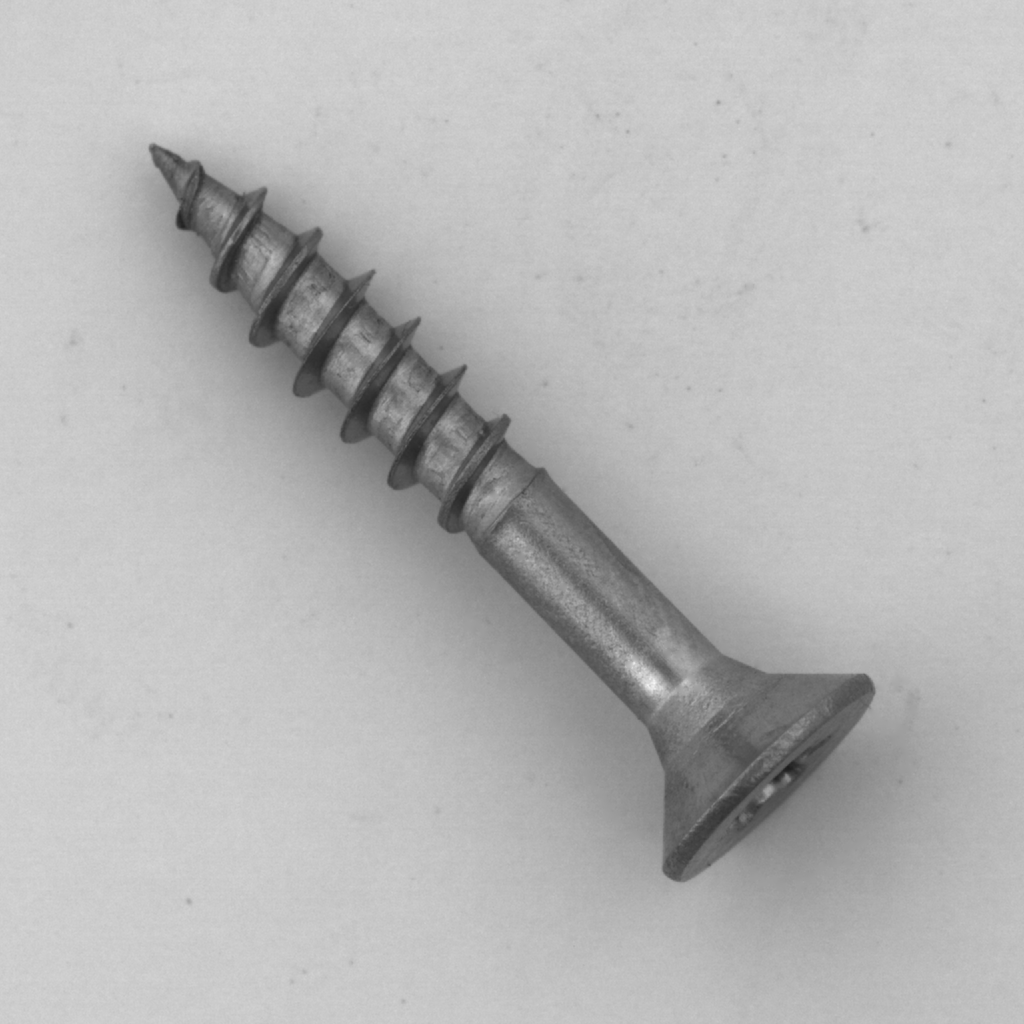} & \includegraphics[width=0.18\linewidth,height=0.70in,keepaspectratio]{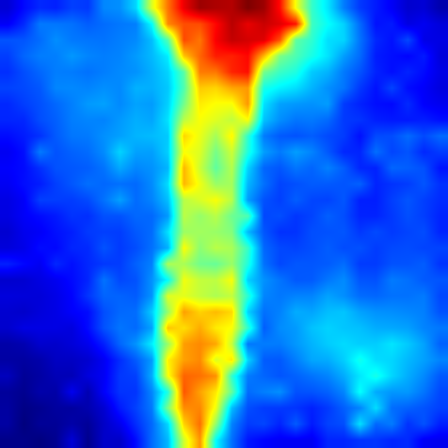} & \includegraphics[width=0.18\linewidth,height=0.70in,keepaspectratio]{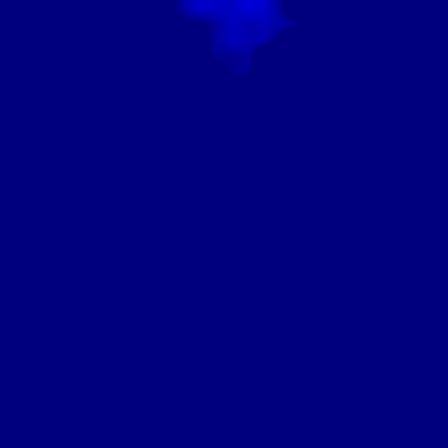} & \includegraphics[width=0.18\linewidth,height=0.70in,keepaspectratio]{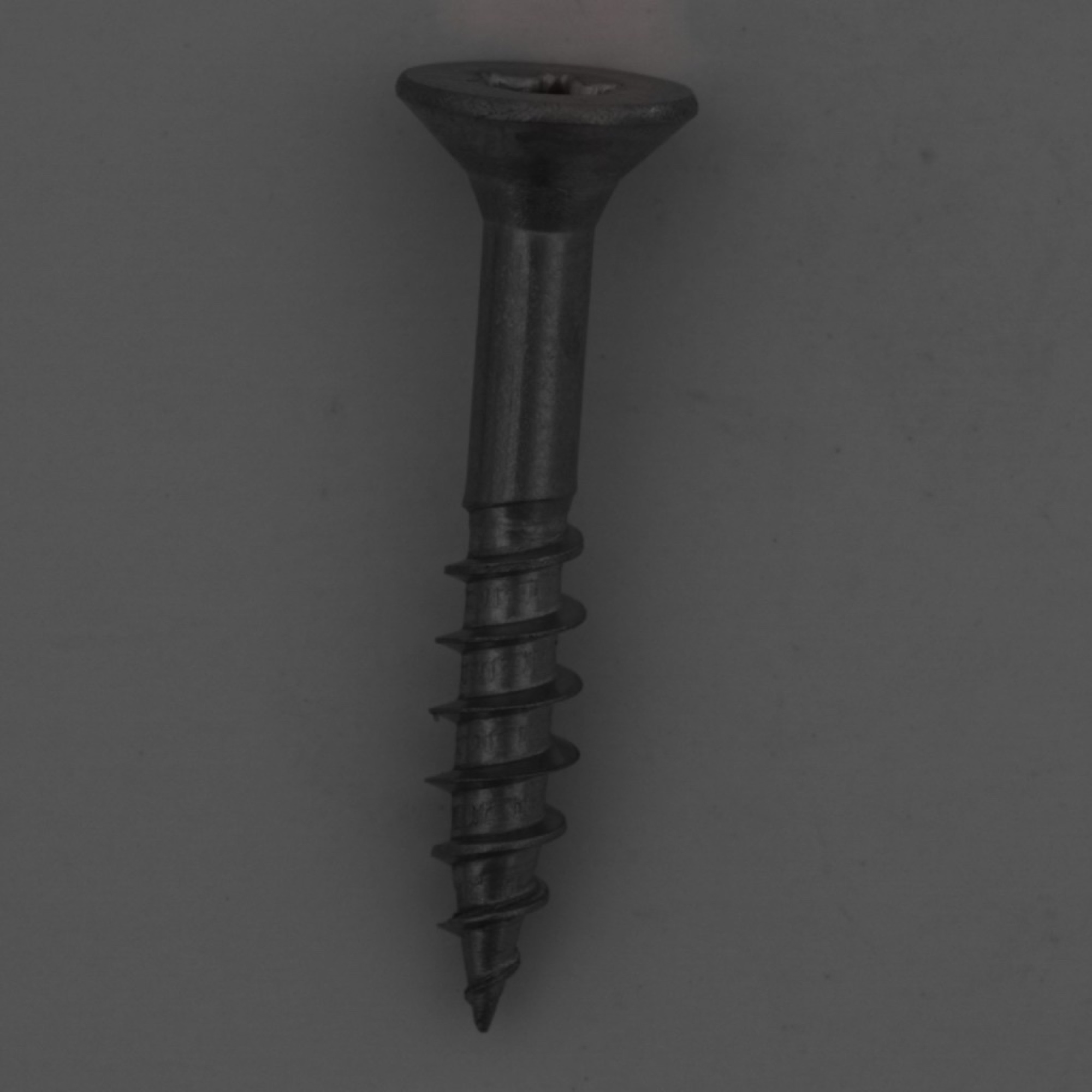} \\
\includegraphics[width=0.18\linewidth,height=0.70in,keepaspectratio]{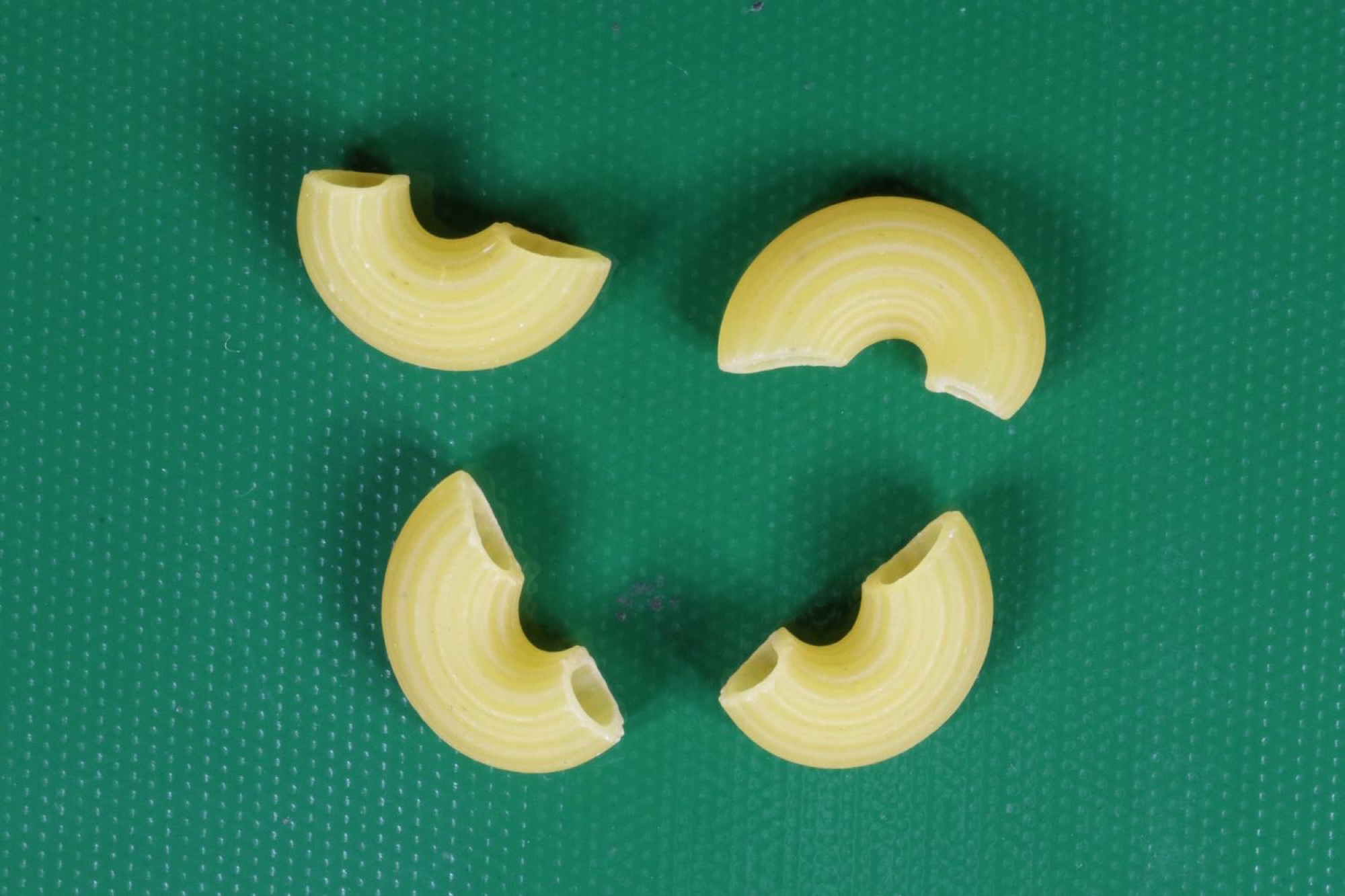} & \includegraphics[width=0.18\linewidth,height=0.70in,keepaspectratio]{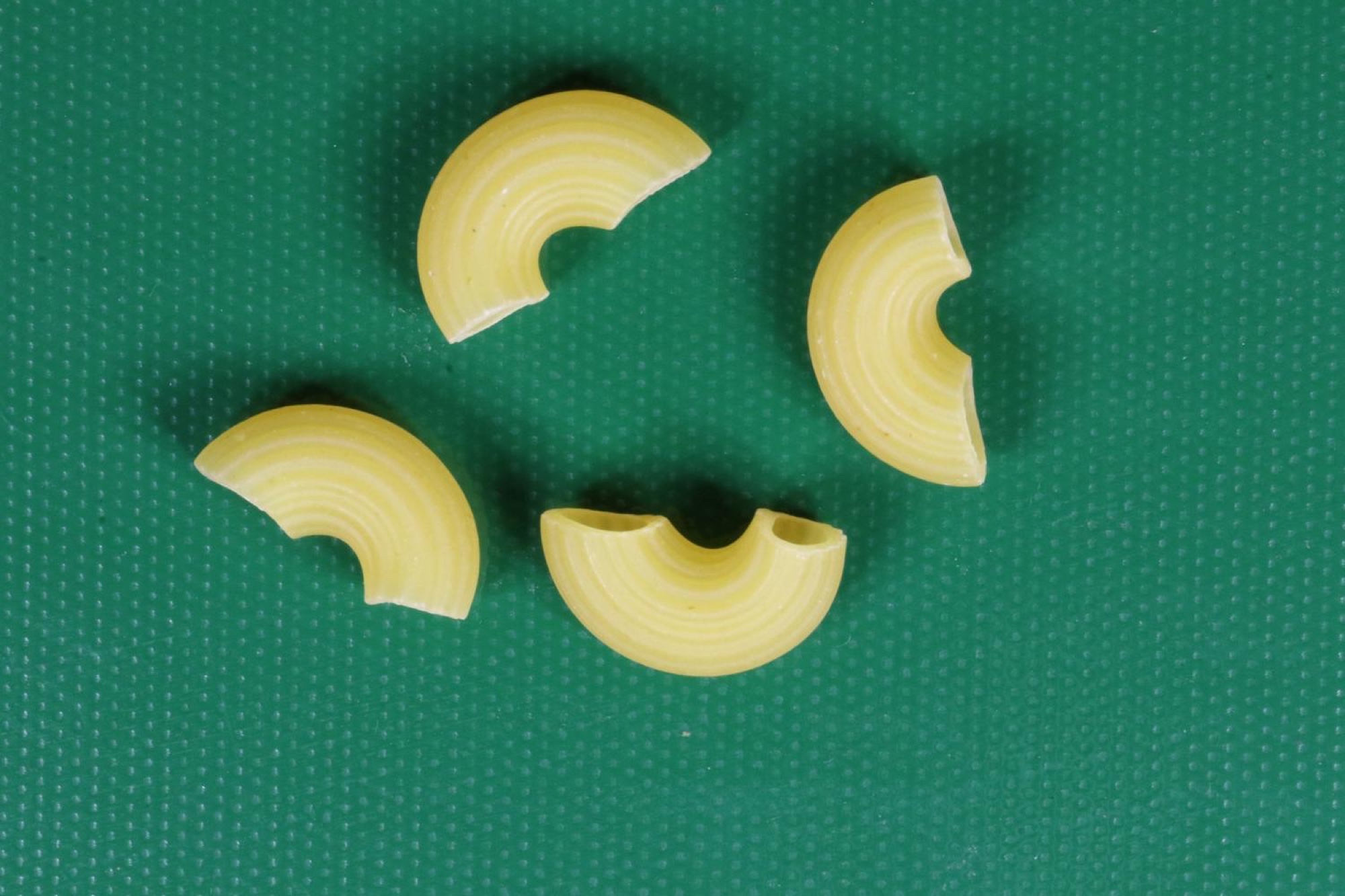} & \includegraphics[width=0.18\linewidth,height=0.70in,keepaspectratio]{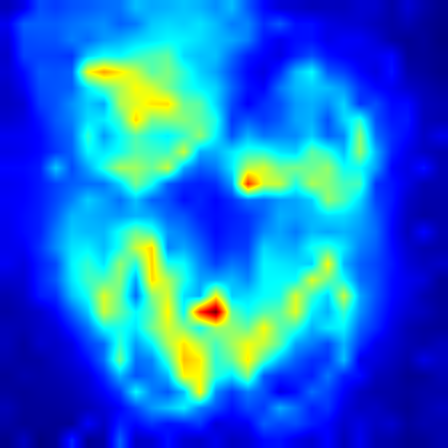} & \includegraphics[width=0.18\linewidth,height=0.70in,keepaspectratio]{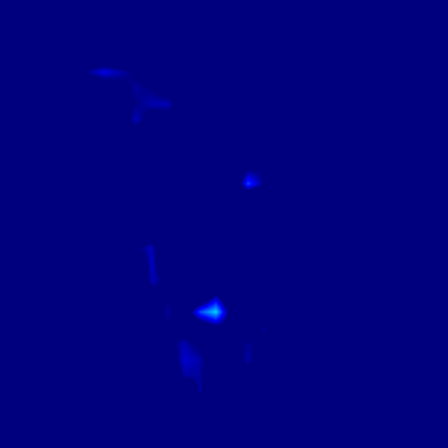} & \includegraphics[width=0.18\linewidth,height=0.70in,keepaspectratio]{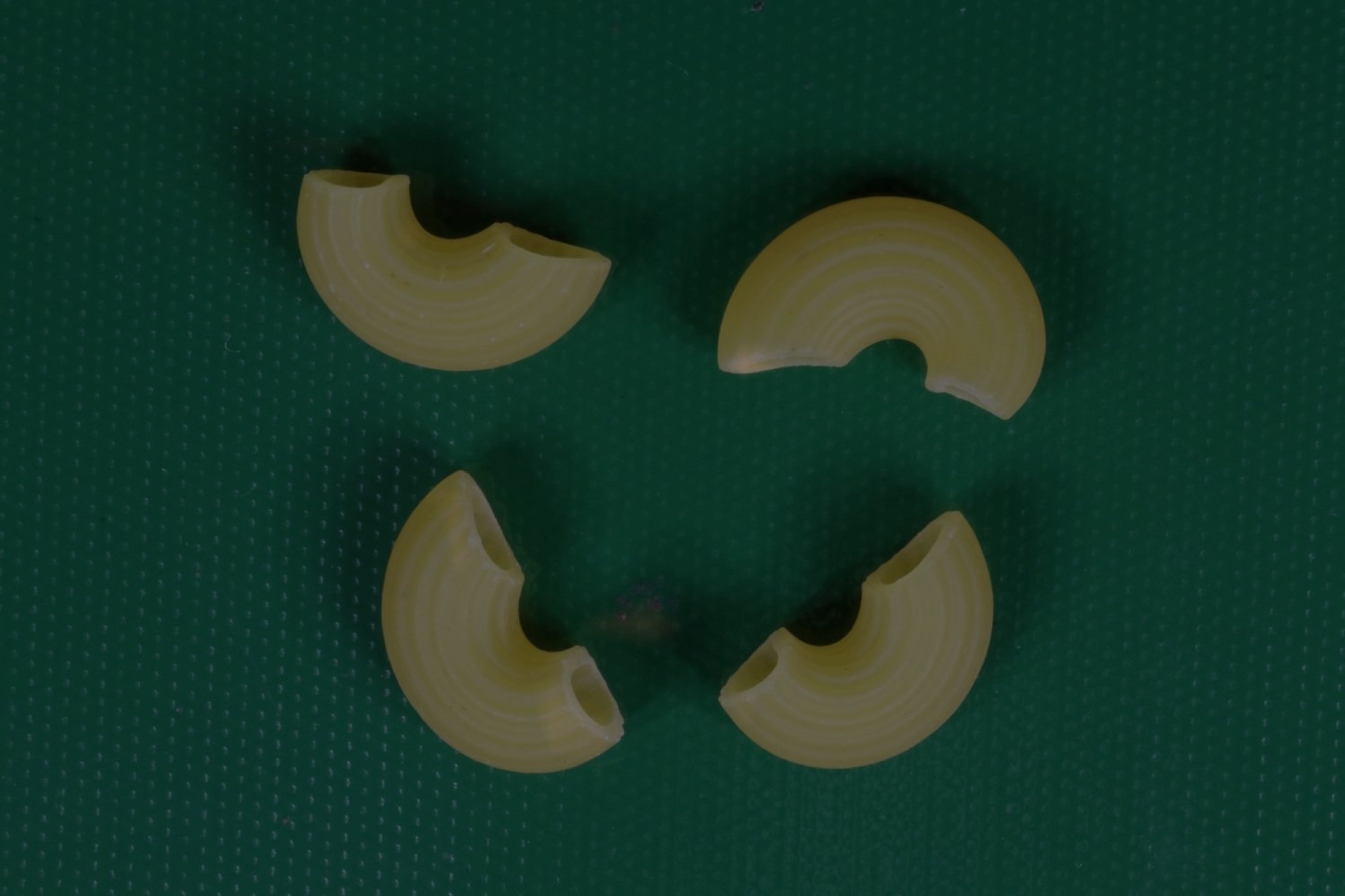} \\
\includegraphics[width=0.18\linewidth,height=0.70in,keepaspectratio]{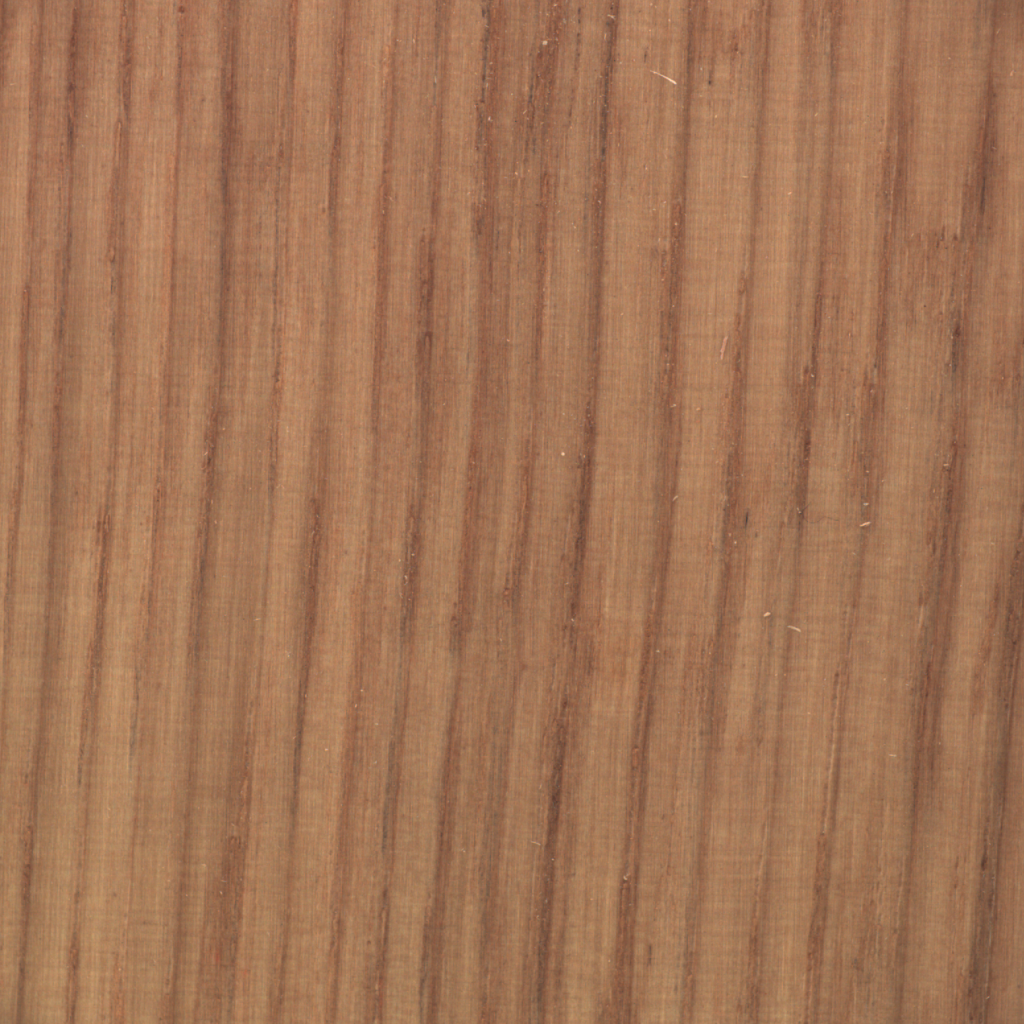} & \includegraphics[width=0.18\linewidth,height=0.70in,keepaspectratio]{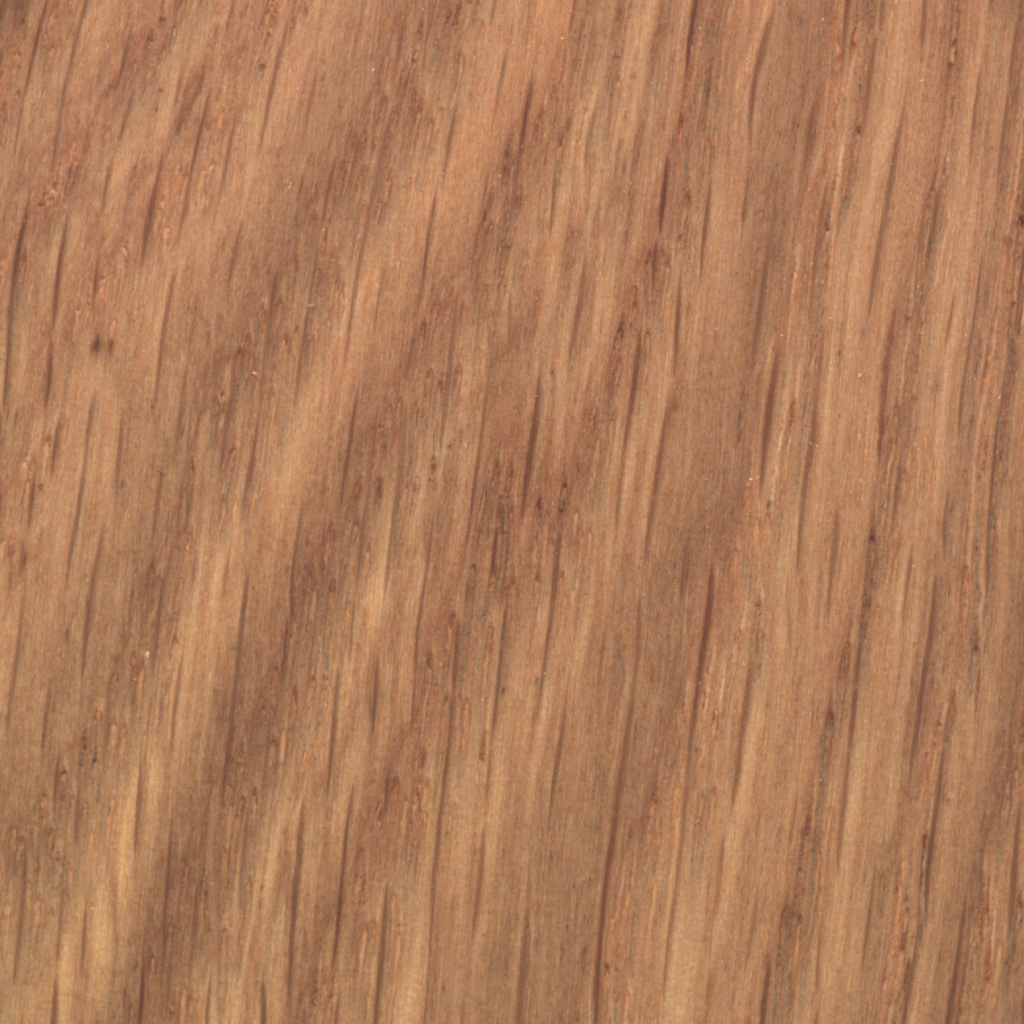} & \includegraphics[width=0.18\linewidth,height=0.70in,keepaspectratio]{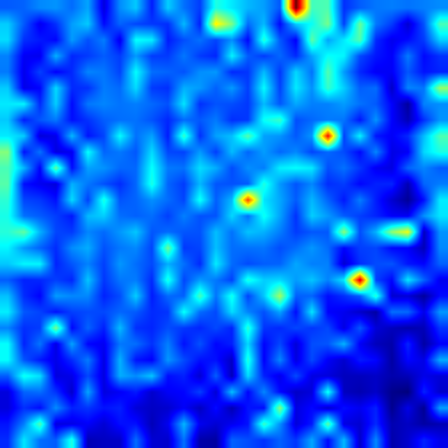} & \includegraphics[width=0.18\linewidth,height=0.70in,keepaspectratio]{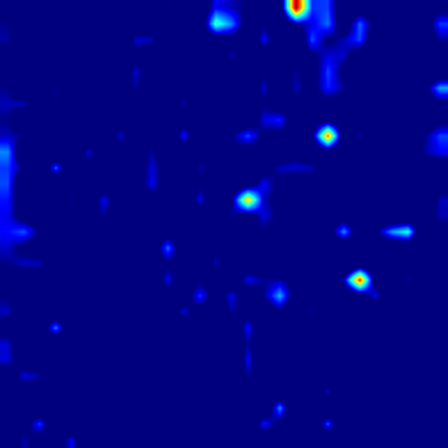} & \includegraphics[width=0.18\linewidth,height=0.70in,keepaspectratio]{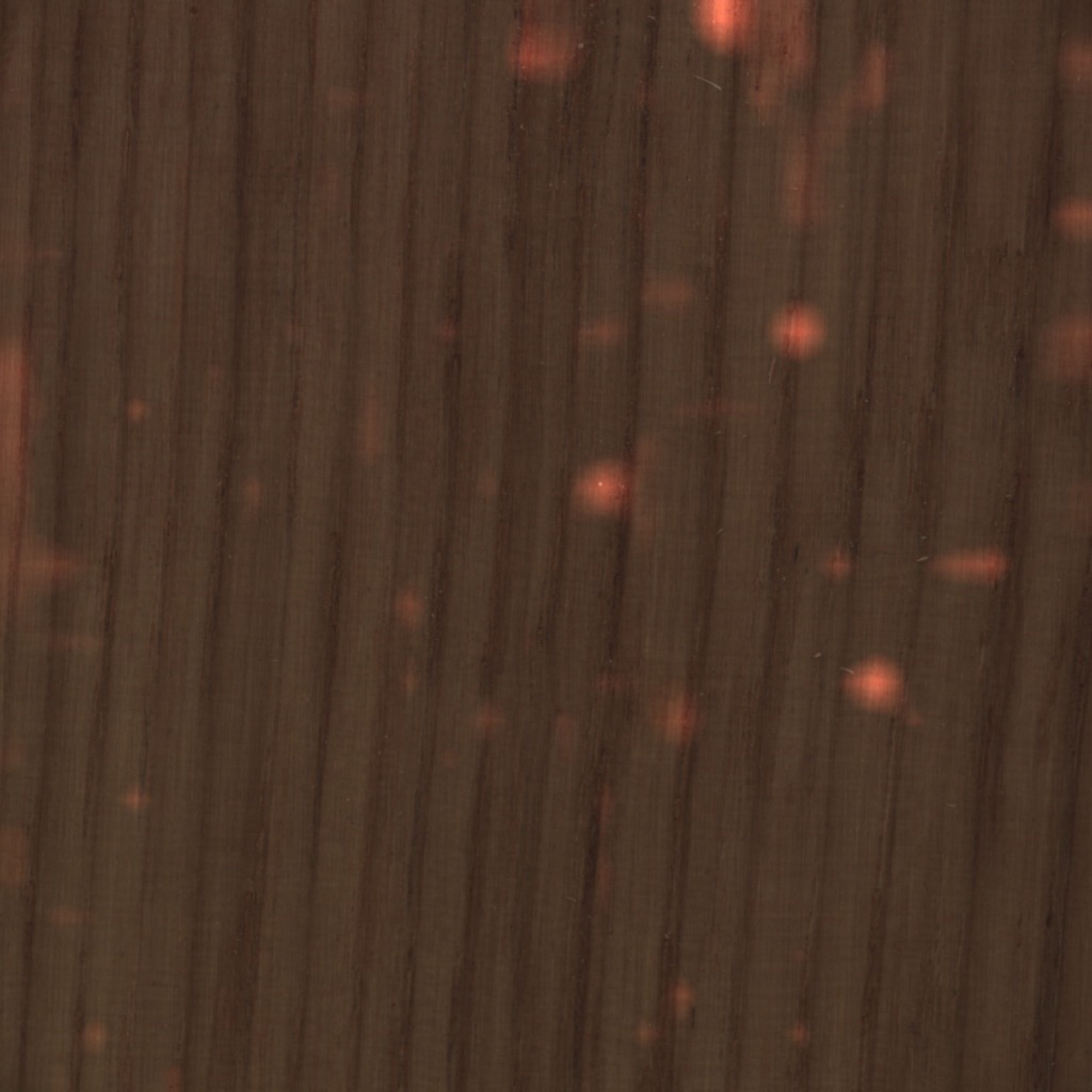} \\
\includegraphics[width=0.18\linewidth,height=0.70in,keepaspectratio]{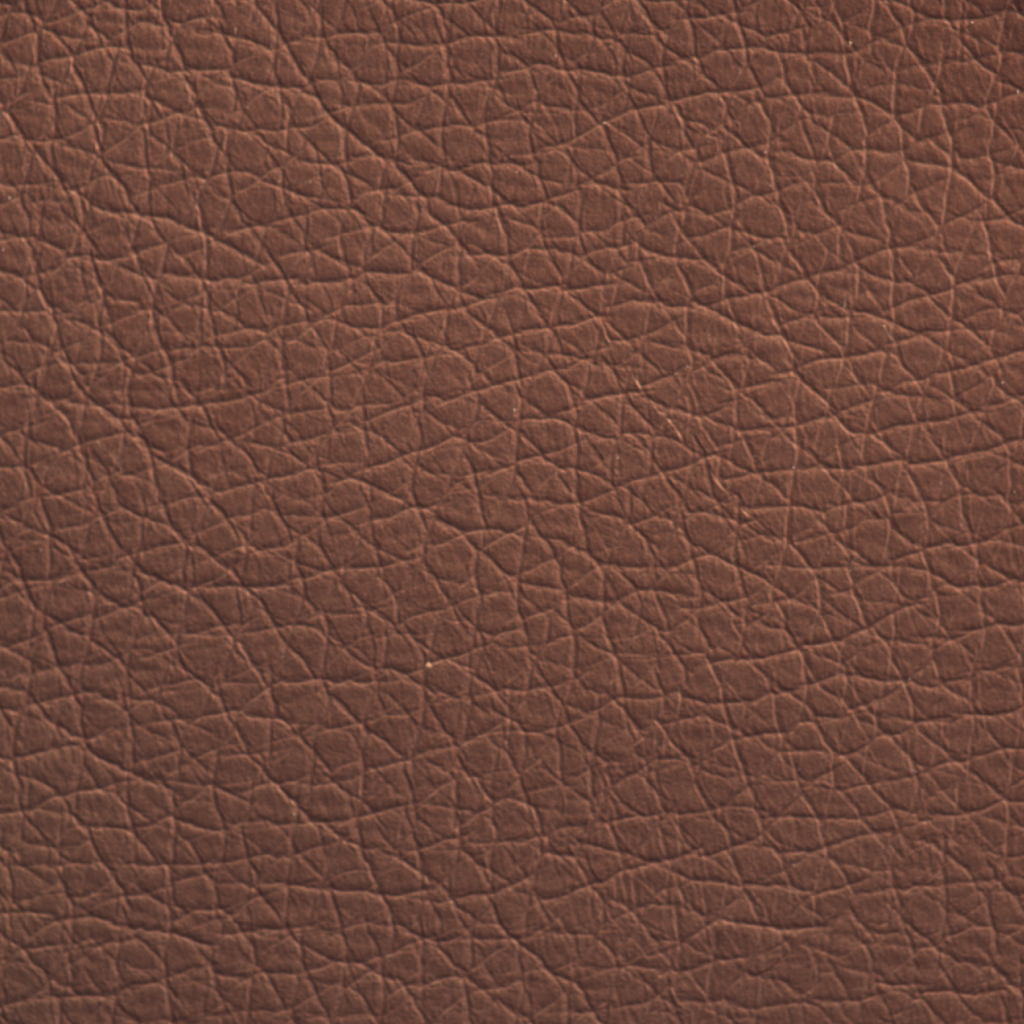} & \includegraphics[width=0.18\linewidth,height=0.70in,keepaspectratio]{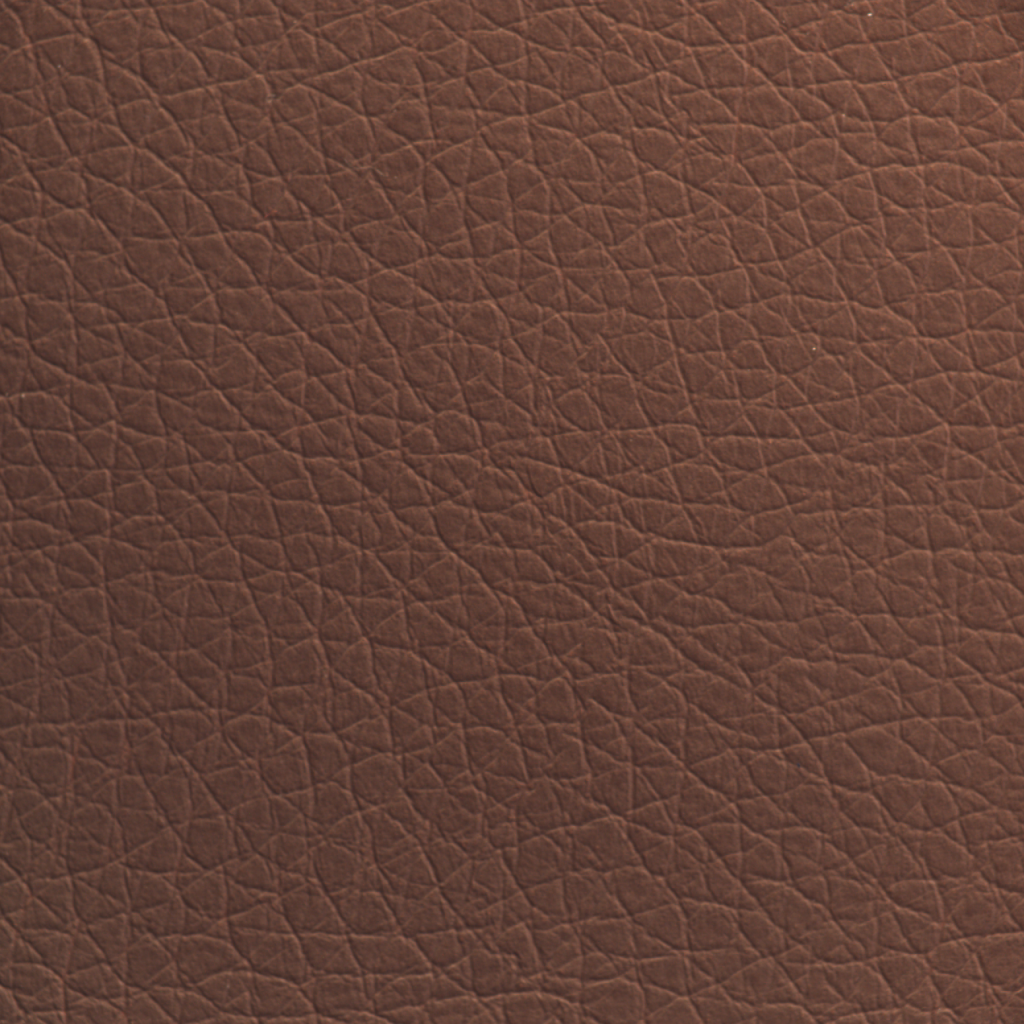} & \includegraphics[width=0.18\linewidth,height=0.70in,keepaspectratio]{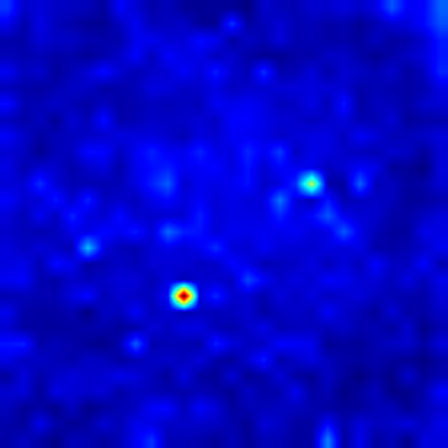} & \includegraphics[width=0.18\linewidth,height=0.70in,keepaspectratio]{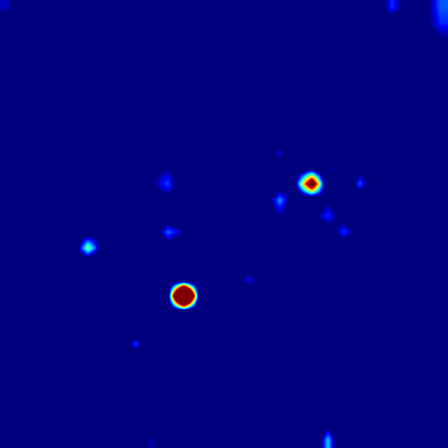} & \includegraphics[width=0.18\linewidth,height=0.70in,keepaspectratio]{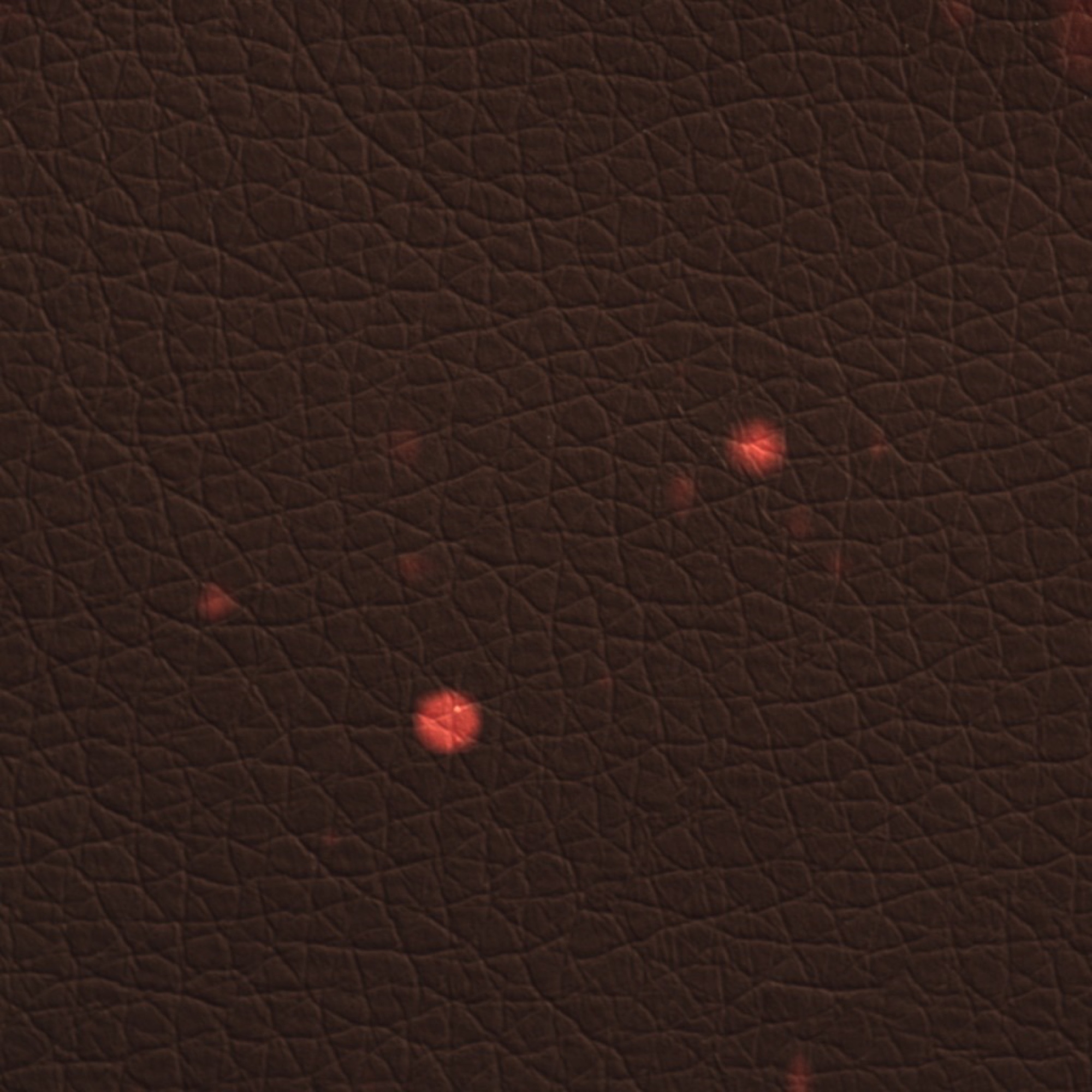} \\
\includegraphics[width=0.18\linewidth,height=0.70in,keepaspectratio]{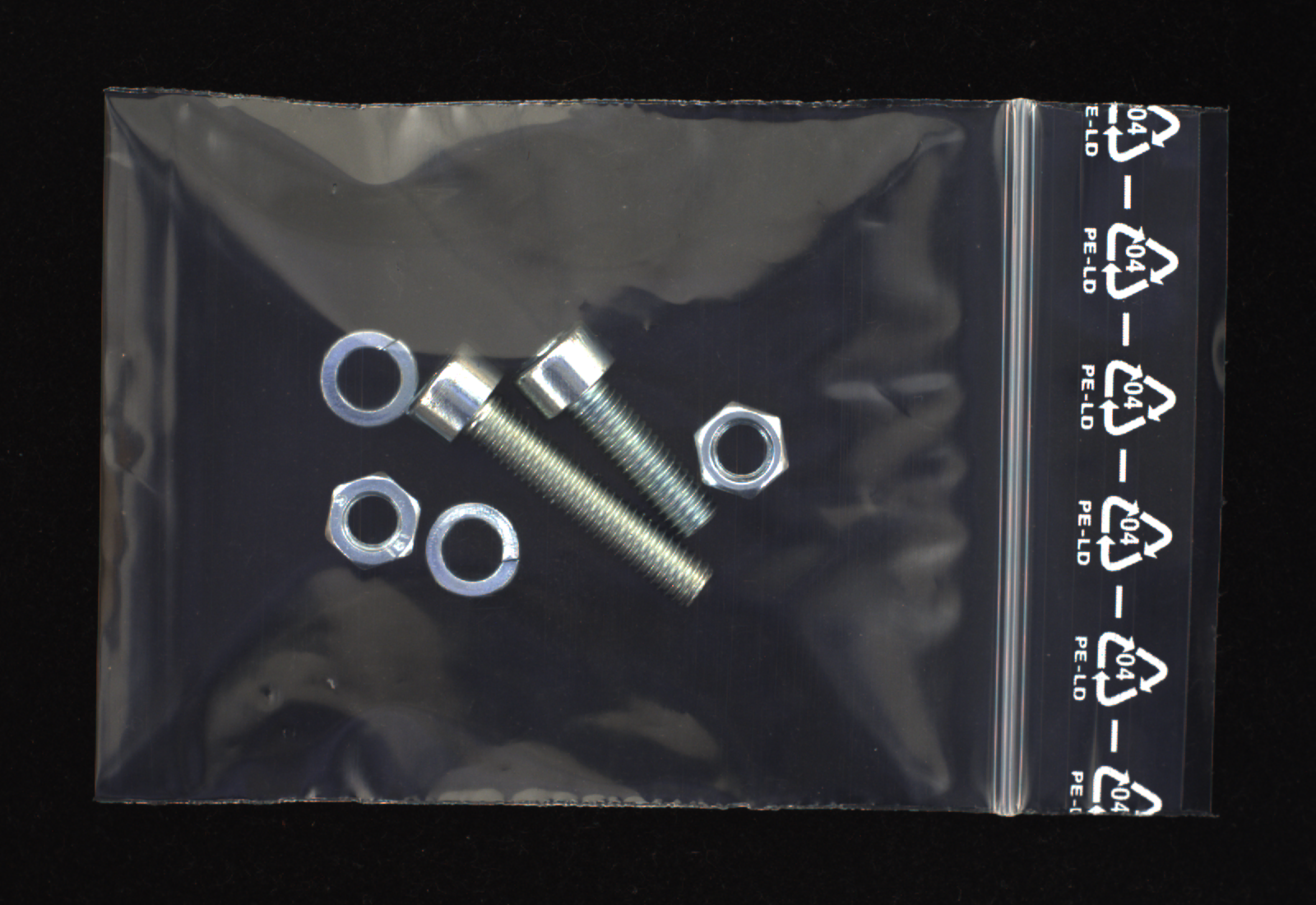} & \includegraphics[width=0.18\linewidth,height=0.70in,keepaspectratio]{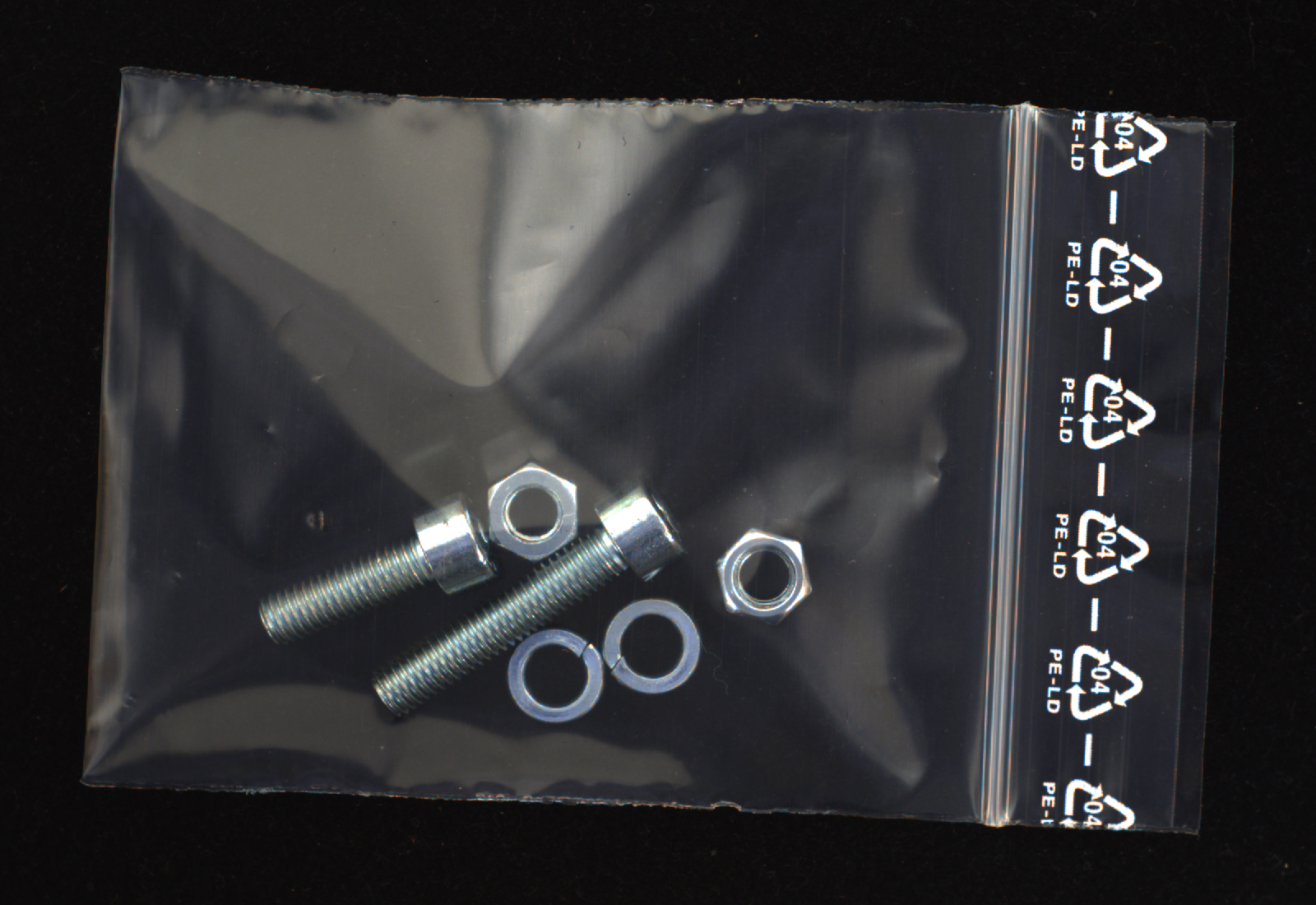} & \includegraphics[width=0.18\linewidth,height=0.70in,keepaspectratio]{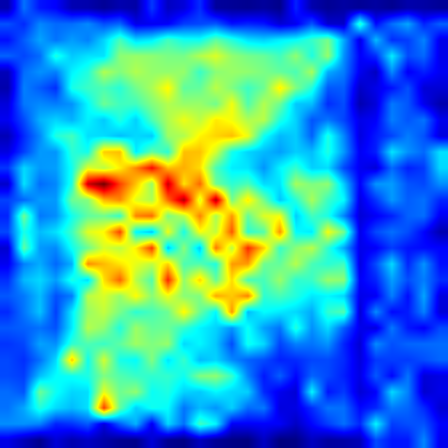} & \includegraphics[width=0.18\linewidth,height=0.70in,keepaspectratio]{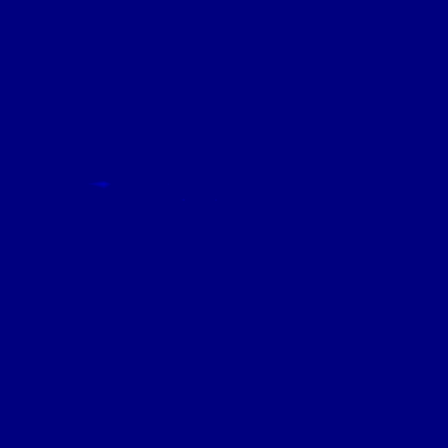} & \includegraphics[width=0.18\linewidth,height=0.70in,keepaspectratio]{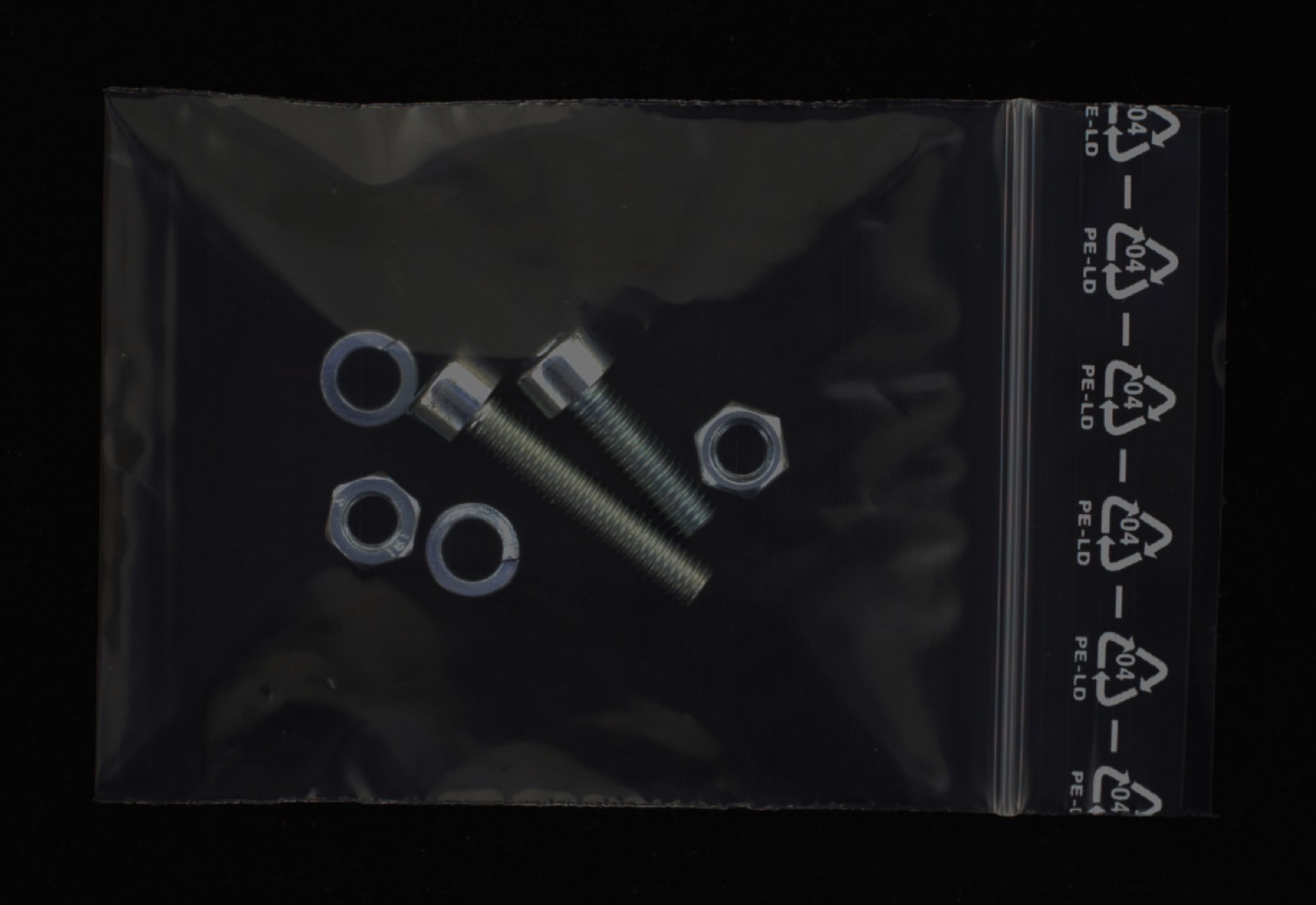} \\
\bottomrule
\end{tabular*}
\caption{Normal queries under raw per-image normalization and normal-bank
calibration. The calibrated maps attenuate responses caused by benign
query--reference variation before the expert-reader query gate.}
\label{fig:blank_on_normal_grid}
\end{figure}

This section complements the calibration analysis in Section~\ref{sec:exp-evidence}. \hyperref[fig:calibration_focus_grid]{Figures~\ref*{fig:calibration_focus_grid}} and~\hyperref[fig:blank_on_normal_grid]{\ref*{fig:blank_on_normal_grid}} visualize noise-robust map calibration on defective and normal queries, respectively; the task-dependent use of the resulting evidence follows Section~\ref{sec:controller}.

\paragraph{Defective queries.}
\autoref{fig:calibration_focus_grid} illustrates calibration on eight
defect-bearing queries. A category-matched normal reference can differ from
the query in pose, object placement, texture, or illumination, so raw
query--reference residuals need not coincide with the defect. Calibrating
against the category-specific normal bank suppresses responses typical of
normal variation and retains more localized evidence for the expert readers.
Each row shows the same query through the five stages of this comparison.

\FloatBarrier

\clearpage
\paragraph{Normal queries.}
\autoref{fig:blank_on_normal_grid} shows the complementary behavior on
eight normal queries. Per-image normalization can turn benign differences
between a query and its category-matched reference into conspicuous raw
responses. The Noise-Robust Calibration attenuates these responses before the
expert-reader route is considered. For anomaly-detection questions,
the spatial-coherence gate in Section~\ref{sec:controller} determines whether
expert readers are queried. The visual comparison
illustrates suppression of spurious evidence; it does not by itself establish
the final answer for each query.

\FloatBarrier

\end{document}